\documentclass[nohyperref]{article}

\usepackage{microtype}
\usepackage{graphicx}
\usepackage{subfigure}
\usepackage{booktabs} 

\usepackage{hyperref}

\usepackage[accepted, nohyperref]{styles/icml22/icml2022}

\usepackage{amsmath}
\usepackage{amssymb}
\usepackage{mathtools}
\usepackage{amsthm}

\usepackage[capitalize,noabbrev]{cleveref}

\theoremstyle{plain}

\theoremstyle{definition}

\theoremstyle{remark}

\usepackage[textsize=tiny]{todonotes}
\usepackage{algpseudocode}
\usepackage{float}
\usepackage{amsmath}
\usepackage{wrapfig} 
\usepackage{enumitem}
\usepackage{graphicx,subfigure}

\usepackage{amsmath,amsfonts,bm}

\def\eqref#1{equation~\ref{#1}}

\def\1{\bm{1}}

\DeclareMathAlphabet{\mathsfit}{\encodingdefault}{\sfdefault}{m}{sl}
\SetMathAlphabet{\mathsfit}{bold}{\encodingdefault}{\sfdefault}{bx}{n}

\DeclareMathOperator*{\argmax}{arg\,max}

\usepackage{multirow}
\usepackage{etoc}
\usepackage{pgfplots}
\usepackage{tikz}
\usetikzlibrary{arrows.meta}
\usetikzlibrary{shapes,positioning,decorations.pathmorphing}

\newenvironment{special_smallmatrix}
  {\Bigg[\begin{smallmatrix}}
  {\end{smallmatrix}\Bigg]}
 
\usepackage{titletoc}

\definecolor{highlightColor}{RGB}{51,133,255}
\definecolor{asparagus}{rgb}{0.53, 0.66, 0.42}
\definecolor{burntorange}{rgb}{0.8, 0.33, 0.0}

\newcommand{\highlight}[1]{\colorbox{highlightColor!20}{#1}}

\icmltitlerunning{Learning Neural Causal Models with Active Interventions}

\begin{document}

\twocolumn[
\icmltitle{Learning Neural Causal Models with Active Interventions}




\begin{icmlauthorlist}
\icmlauthor{Nino Scherrer}{eth}
\icmlauthor{Olexa Bilaniuk}{mila}
\icmlauthor{Yashas Annadani}{eth}
\icmlauthor{Anirudh Goyal}{mila}
\icmlauthor{Patrick Schwab}{gsk}
\icmlauthor{Bernhard Schölkopf}{mpi}
\icmlauthor{Michael C. Mozer}{brain}
\icmlauthor{Yoshua Bengio}{mila}
\icmlauthor{Stefan Bauer}{kth,gsk}
\icmlauthor{Nan Rosemary Ke}{deepmind}
\end{icmlauthorlist}

\icmlaffiliation{eth}{ETH Zurich}
\icmlaffiliation{mila}{Mila, Universite de Montréal}
\icmlaffiliation{gsk}{GlaxoSmithKline}
\icmlaffiliation{mpi}{Max Planck Institute for Intelligent Systems}
\icmlaffiliation{brain}{Google Research, Brain Team}
\icmlaffiliation{kth}{KTH Stockholm}
\icmlaffiliation{deepmind}{DeepMind}

\icmlcorrespondingauthor{Nino Scherrer}{\href{mailto:nino.scherrer@gmail.com}{nino.scherrer@gmail.com}}

\icmlkeywords{Machine Learning, ICML}

\vskip 0.3in
]



\printAffiliationsAndNotice{}  

\begin{abstract}
Discovering causal structures from data is a challenging inference problem of fundamental importance in all areas of science. The appealing properties of neural networks have recently led to a surge of interest in differentiable neural network-based methods for learning causal structures from data. So far, differentiable causal discovery has focused on static datasets of observational or fixed interventional origin. In this work, we introduce an active intervention targeting (AIT) method which enables a quick identification of the underlying causal structure of the data-generating process. Our method significantly reduces the required number of interactions compared with random intervention targeting and is applicable for both discrete and continuous optimization formulations of learning the underlying directed acyclic graph (DAG) from data. We examine the proposed method across multiple frameworks in a wide range of settings and demonstrate superior performance on multiple benchmarks from simulated to real-world data. 
\end{abstract}

\section{Introduction} \label{sec:intro}
Inferring causal structure from data is a challenging but important task that lies at the heart of scientific reasoning and accompanying progress \citep{lauritzen1988local,friedman2000using,robins2000marginal, sachs2005causal,korb2010bayesian,hill2016inferring,vandenbroucke2016causality,de2019causality}. Recently, there has been a surge in interest in differentiable causal structure learning with neural networks, also known as neural causal discovery \citep{ke2019learning, scholkopf2021toward, xia2021causal}. These methods propose to recast the discrete search over the combinatorial solution space by treating it as an optimization problem with smoothly differentiable parameters. The set of neural parameters embodies a neural causal model $\mathcal{N}$ that represents parameters of both structural and functional nature. Structural parameters express the belief about the graph structure through a distribution over graphs, for example with a soft-adjacency matrix. On the other hand, functional parameters characterize the conditional probability distributions of the factorized joint distribution of a directed graphical model. Overall, such models offer promising abilities with respect to generalization and fast adaptation \citep{bengio2019meta}.

\begin{figure}[t!]
    \centering
    \vspace{-1mm}
    \begin{tikzpicture}[scale=0.63]
        \scriptsize
        \node[cloud, draw, align=left, fill=yellow!10, opacity=1, cloud puffs=20, cloud puff arc=120, aspect=3, inner sep=0mm] at (-6.1, 1.7)   (cloud_obs) {\scriptsize Obs. Data};
        
        \node[cloud, draw, align=left, fill=yellow!80, opacity=1, cloud puffs=20, cloud puff arc=120, aspect=3, inner sep=0mm] at (-3.9, 1.7)   (cloud_obs) {\scriptsize Int. Data};
         
        \draw[black, dashed, align=center,] (-7.4,0.95) rectangle (-2.7,2.4)
                node[above left] {\scriptsize Fused Data};
         
         \draw[black,align=center,fill=lightgray!10] (-7,-2) rectangle (-3,-0.5)
                node[pos=.5] {Neural \\ Causal Discovery};
                
        \draw[black,align=center,fill=asparagus!20, fill opacity=1.0, text opacity=1] (-5,0.75) rectangle (-3,  0.1)
                node[pos=.5] {AIT};
        
        \draw [-{Stealth[slant=0]}](-2.7,-1.2) -- (-1.7,-1.2);
        
        \draw [black, dashed](-3.8,-0.4) -- (-3.8,0);
        \draw [-{Stealth[slant=0]}, black, dashed](-3.8,0.75) -- (-3.8,1.2);
        \draw [black, dashed](-4.2,1.2) -- (-4.2,0.75);
        \draw [-{Stealth[slant=0]}, black, dashed](-4.2,0) -- (-4.2,-0.4);
        
        \draw [-{Stealth[slant=0]}](-6,1.1) -- (-6,-0.4);
        
        \draw[black,rounded corners=10] (-1.5,-2) rectangle (5,2.4)
                node[above left] {\scriptsize Neural Causal Model $\mathcal{N}$};
                
        
        \draw[black,rounded corners=10,dotted] (-1.3,-1.8) rectangle (1.2,1.7)
                node[above left] {\scriptsize Structural};
        
        \draw[black,rounded corners=10,dotted] (1.3,-1.8) rectangle (4.7,1.7)
                node[above left] {\scriptsize Functional};
        
        
         \node[shape=circle,draw=black, scale=0.4] (A-0) at (0,1.2) {$X_0$};
         \node[shape=circle,draw=black, scale=0.5] (B-0) at (-0.7,0.4) {$X_1$};
         \node[shape=circle,draw=black, scale=0.4] (C-0) at (0.7,0.4) {$X_2$};
         
         \path [-{Stealth[slant=0]}] (A-0) edge node[left] {} (B-0);
         \path [-{Stealth[slant=0]}] (A-0) edge node[left] {} (C-0);
         \path [-{Stealth[slant=0]}] (B-0) edge node[left] {} (C-0);
         
         \draw [double, Implies-Implies] (0.0, -0.2) -- (0.0, 0.25);
         
        \draw[black, draw=none, align=center, rounded corners=0, fill opacity=0, text opacity= 1.0] (-0.2, -1.9) rectangle (0.2, 0.3) node[pos=.5] { $\begin{special_smallmatrix} 0.0 & 0.1 & 0.1\\ 0.9 & 0.0 & 0.1\\ 0.9 & 0.9 & 0.0  \end{special_smallmatrix} $ };
        
        \node [align=center] at (3, 0.7) (function1)   {\scriptsize $P_\theta(X_0 | X_{pa(0)} )$};
        
        \node[align=center] at (3, 0.0)  (function_i)    {\scriptsize $P_\theta(X_1 | X_{pa(1)} )$};
        
        \node[align=center] at (3, -0.7)  (function_N)   {\scriptsize $P_\theta(X_2 | X_{pa(2)} )$};
        
    \end{tikzpicture}
    \vspace{-2mm}
    \caption{AIT is an active intervention targeting technique which is applicable to all neural causal discovery frameworks of fused data. Based on the state of a learned neural causal model $\mathcal{N}$ up to a given timepoint, AIT selects the next informative intervention target for the causal discovery. }
    \vspace{-1.5\baselineskip}
    \label{fig:AIT_page1}
\end{figure}
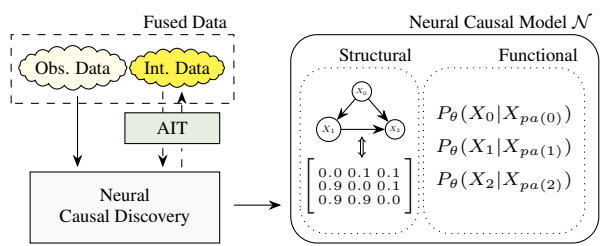

\begin{figure*}[t!]
    \centering
    \vspace{-1mm}
    \begin{tikzpicture}[scale=0.75]
        \scriptsize
        
        \draw[black,rounded corners=10] (0,3.5) rectangle (10,6.5)
                node[above left] {\scriptsize Neural Causal Model $\mathcal{N}$};
            
        \draw[black,rounded corners=10,dotted] (0.4, 3.75) rectangle (4.75,6)
                node[above left] {\scriptsize Structural Parameters $\gamma$};    
                
         \draw[black, draw=none, align=center, rounded corners=0] (0.75, 4.25) rectangle (4.5, 5.5) node[pos=.5] {$ \sigma(\gamma) = \begin{bmatrix}
0.0 & 0.1 & 0.0\\
0.9 & 0.0 & 0.5\\
0.0 & 0.5 & 0.0
\end{bmatrix} $ } ;

\draw [-{Stealth[slant=0]}](1.2,3.75) -- (1.2,2.75);
\draw [-{Stealth[slant=0]}](7.35,3.75) -- (7.35,3.2) -- (12.25, 3.2) -- (12.25,2.75);

        \draw[black,rounded corners=10,dotted] (5.25, 3.75) rectangle (9.6, 6)
                node[above left] {\scriptsize Functional Parameters $\theta$};  
                
         \draw[black, very thin, align=center, rounded corners=0] (5.55, 3.9) rectangle (9.2,4.4) node[pos=.5] {MLP$_2(\theta) \leftarrow p_\theta(X_2 | ...)$};
         \draw[black, very thin, align=center, rounded corners=0] (5.55 , 4.6) rectangle (9.2,5.1) node[pos=.5] {MLP$_1(\theta) \leftarrow p_\theta(X_2 | ...)$};
         \draw[black, very thin, align=center, rounded corners=0] (5.55, 5.3) rectangle (9.2,5.8) node[pos=.5] {MLP$_0(\theta) \leftarrow p_\theta(X_0 | ...)$};
                
        \draw[black, dashed, rounded corners=0, fill=asparagus!20,fill opacity=0.2, text opacity=1.0] (0,-1) rectangle (20,3)
                node[above left] {\scriptsize AIT};
                
         \draw[black, solid, align=center, rounded corners=0, fill=asparagus!15] (0.5,-0.75) rectangle (2,2.75)
                node[pos=.5] {Two-Phase\\DAG\\Sampling};
         
         \node[draw=none,fill=none] at (2.5, 1.4) {$\mathcal{G}_0$:};       
         \node[shape=circle,draw=black, scale=0.6] (A-0) at (3.1,1.4) {0};
         \node[shape=circle,draw=black, scale=0.6] (B-0) at (4.0,1.4) {1};
         \node[shape=circle,draw=black, scale=0.6] (C-0) at (4.9,1.4) {2};
        \path [-{Stealth[slant=0]}] (A-0) edge node[left] {} (B-0);
        \path [-{Stealth[slant=0]}] (B-0) edge node[left] {} (C-0);
         
         \node[draw=none,fill=none] at (2.5, 0.7) {$\mathcal{G}_1$:};  
         \node[shape=circle,draw=black, scale=0.6] (A-1) at (3.1,0.7) {0};
         \node[shape=circle,draw=black, scale=0.6] (B-1) at (4.0,0.7) {1};
         \node[shape=circle,draw=black, scale=0.6] (C-1) at (4.9,0.7) {2};
         \path [-{Stealth[slant=0]}] (A-1) edge node[left] {} (B-1);
         \path [-{Stealth[slant=0]}] (C-1) edge node[left] {} (B-1);
                
        \draw[black, solid, align=center, rounded corners=0, fill=asparagus!15] (5.5,-0.75) rectangle (7,2.75)
                node[pos=.5] {Apply\\Inter.};
             
        
        \node[draw=none,fill=none] at (8.0, 2.4) {$\mathcal{G}_{0,\textcolor{burntorange}{0}}$:};  
         \node[shape=circle,draw=orange!60, fill=orange!20, scale=0.6] (A-0-0) at (9.0,2.4) {0};
         \node[shape=circle,draw=black, scale=0.6] (B-0-0) at (9.9,2.4) {1};
         \node[shape=circle,draw=black, scale=0.6] (C-0-0) at (10.8,2.4) {2};
        \path [-{Stealth[slant=0]}] (A-0-0) edge node[left] {} (B-0-0);
        \path [-{Stealth[slant=0]}] (B-0-0) edge node[left] {} (C-0-0);
        
        \node[draw=none,fill=none] at (8.0, 1.9) {$\mathcal{G}_{1,\textcolor{burntorange}{0}}$:}; 
        \node[shape=circle,draw=orange!60, fill=orange!20, scale=0.6] (A-1-0) at (9.0,1.9) {0};
        \node[shape=circle,draw=black, scale=0.6] (B-1-0) at (9.9,1.9) {1};
        \node[shape=circle,draw=black, scale=0.6] (C-1-0) at (10.8,1.9) {2};
        \path [-{Stealth[slant=0]}] (A-1-0) edge node[left] {} (B-1-0);
        \path [-{Stealth[slant=0]}] (C-1-0) edge node[left] {} (B-1-0);
        
        \draw [dashed, opacity=0.4] (7.2,1.55) -- (11.3,1.55);
        
        
        \node[draw=none,fill=none] at (8.0, 1.2) {$\mathcal{G}_{0,\textcolor{burntorange}{1}}$:}; 
        \node[shape=circle,draw=black, scale=0.6] (A-0-1) at (9.0,1.2) {0};
         \node[shape=circle,draw=orange!60, fill=orange!20, scale=0.6] (B-0-1) at (9.9,1.2) {1};
         \node[shape=circle,draw=black, scale=0.6] (C-0-1) at (10.8,1.2) {2};
        \path [-{Stealth[slant=0]}] (B-0-1) edge node[left] {} (C-0-1);
        
        \node[draw=none,fill=none] at (8.0, 0.7) {$\mathcal{G}_{1,\textcolor{burntorange}{1}}$:}; 
        \node[shape=circle,draw=black, scale=0.6] (A-1-1) at (9.0,0.7) {0};
        \node[shape=circle,draw=orange!60, fill=orange!20, scale=0.6] (B-1-1) at (9.9,0.7) {1};
        \node[shape=circle,draw=black, scale=0.6] (C-1-1) at (10.8,0.7) {2};
        \path [-{Stealth[slant=0]}] (C-1-1) edge node[left] {} (B-1-1);
        
        \draw [dashed, opacity=0.4] (7.2,0.35) -- (11.3,0.35);
        
        
        \node[draw=none,fill=none] at (8.0, 0.0) {$\mathcal{G}_{0,\textcolor{burntorange}{2}}$:};
        \node[shape=circle,draw=black, scale=0.6] (A-0-2) at (9.0,0.0) {0};
         \node[shape=circle,draw=black, scale=0.6] (B-0-2) at (9.9,0.0) {1};
         \node[shape=circle,draw=orange!60, fill=orange!20, scale=0.6] (C-0-2) at (10.8,0.0) {2};
        \path [-{Stealth[slant=0]}] (A-0-2) edge node[left] {} (B-0-2);
        
        \node[draw=none,fill=none] at (8.0, -0.5) {$\mathcal{G}_{1,\textcolor{burntorange}{2}}$:};
        \node[shape=circle,draw=black, scale=0.6] (A-1-2) at (9.0,-0.5) {0};
        \node[shape=circle,draw=black, scale=0.6] (B-1-2) at (9.9,-0.5) {1};
        \node[shape=circle,draw=orange!60, fill=orange!20, scale=0.6] (C-1-2) at (10.8,-0.5) {2};
        \path [-{Stealth[slant=0]}] (A-1-2) edge node[left] {} (B-1-2);
        \path [-{Stealth[slant=0]}] (C-1-2) edge node[left] {} (B-1-2);
                
        \draw[black, solid, align=center, rounded corners=0, fill=asparagus!15] (11.5,-0.75) rectangle (13,2.75)
                node[pos=.5] {Ancestral\\Sampling};
                
        \node[draw=none,fill=none] at (14, 2.4) {$S_{0,0}$};
        \node[draw=none,fill=none] at (14, 1.9) {$S_{1,0}$};
            
        \draw [dashed, opacity=0.4] (13.2,1.55) -- (14.85,1.55);
        
        \node[draw=none,fill=none] at (14, 1.2) {$S_{0,1}$};
        \node[draw=none,fill=none] at (14, 0.7) {$S_{1,1}$};
        
        \draw [dashed, opacity=0.4] (13.2,0.35) -- (14.85,0.35);
        
        \node[draw=none,fill=none] at (14, 0.0) {$S_{0,1}$};
        \node[draw=none,fill=none] at (14, -0.5) {$S_{1,1}$};

        \draw[black, solid, align=center, rounded corners=0, fill=asparagus!15] (15.0,-0.75) rectangle (16.5,2.75)
                node[pos=.5] {Score\\Comp.};
                
        \node[draw=none,fill=none] at (17.2, 2.15) {$D_{0}$};
        \draw [dashed, opacity=0.4] (16.7,1.55) -- (17.85,1.55);    
        \node[draw=none,fill=none] at (17.2, 0.95) {$D_{1}$};
        \draw [dashed, opacity=0.4] (16.7,0.35) -- (17.85,0.35);
        \node[draw=none,fill=none] at (17.2, -0.25) {$D_{2}$};
                
         \draw[black, solid, align=center, rounded corners=0, fill=asparagus!15] (18,-0.75) rectangle (19.5,2.75)
                node[pos=.5] {$\underset{k}{\argmax}$ \\\quad\vspace{-0mm}\\ $D_k$};
                
        
        \node[cloud, draw, align=left, fill=yellow!10, opacity=1, cloud puffs=20, cloud puff arc=120, aspect=2, inner sep=0mm] at (13, 5.8)   (cloud_obs) {\scriptsize Obs. Data};
        
        \draw [-{Stealth[slant=0]}](12,5.8) -- (10.2,5.8);
        \draw [-{Stealth[slant=0]}, dashed, opacity=0.5](15.5,5.8) -- (14.1,5.8);
        
         \node[cloud, draw, align=left, fill=yellow!80, opacity=1, cloud puffs=20, cloud puff arc=120, aspect=2, inner sep=0mm] at (13, 4.5)   (cloud_obs) {\scriptsize Int. Data};
         
         \draw [-{Stealth[slant=0]}](12,4.5) -- (10.2,4.5);
         \draw [-{Stealth[slant=0]}, dashed, opacity=0.5](15.5,4.5) -- (14.1,4.5);
         
         \draw [-{Stealth[slant=0]}, dashed, opacity=0.5] (18.75, 2.75) -- (18.75,4.5) -- (18.4,4.5);
         
                node[pos=.5] {Perform Intervention};
         
         \node[cloud, draw, align=left, fill=gray!20, text opacity=1, fill opacity=0.2, cloud puffs=20, cloud puff arc=120, aspect=1.3, inner sep=-1mm] at (17, 5.3)   (cloud_obs) {\scriptsize \\\vspace{20mm}Unobserved Nature};    
         
         \node[shape=circle,draw=black, scale=0.6] (A-true) at (16.1,5.8) {0};
         \node[shape=circle,draw=black, scale=0.6] (B-true) at (17.0,5.8) {1};
         \node[shape=circle,draw=black, scale=0.6] (C-true) at (17.9,5.8) {2};
         \path [-{Stealth[slant=0]}] (A-true) edge node[left] {} (B-true);
         \path [-{Stealth[slant=0]}] (B-true) edge node[left] {} (C-true);
         
         \node[shape=circle,draw=black, scale=0.6] (A-perturb) at (16.1,4.5) {0};
         \node[shape=circle,draw=black, fill=yellow!60,, scale=0.6] (B-perturb) at (17.0,4.5) {1};
         \node[shape=circle,draw=black, scale=0.6] (C-perturb) at (17.9,4.5) {2};
         \path [-{Stealth[slant=0]}] (A-perturb) edge node[left] {} (B-perturb);
         \path [-{Stealth[slant=0]}] (B-perturb) edge node[left] {} (C-perturb);

    \end{tikzpicture}
    \vspace{-1mm}
    \caption{AIT decides where to intervene by computing a score for all possible intervention targets. Given a neural causal model $\mathcal{N}$, AIT starts by sampling a set of hypothesis DAGs $\mathcal{G}$ from the structural parameters. It proceeds by applying an intervention $I_k$ to all $\mathcal{G}_i \in \mathcal{G}$. Based on the topological orderings of the post-interventional DAGs $\mathcal{G}_{i,k}$ and the functional parameters, it generates a set of post-interventional samples $S_{i,k}$. AIT proceeds by comparing statistics of the post-interventional sample distributions across the hypothesis graphs to compute the discrepancy score $D_k$. 
    }
    \label{fig:AIT_in_detail}
    \vspace{-1\baselineskip}
\end{figure*}
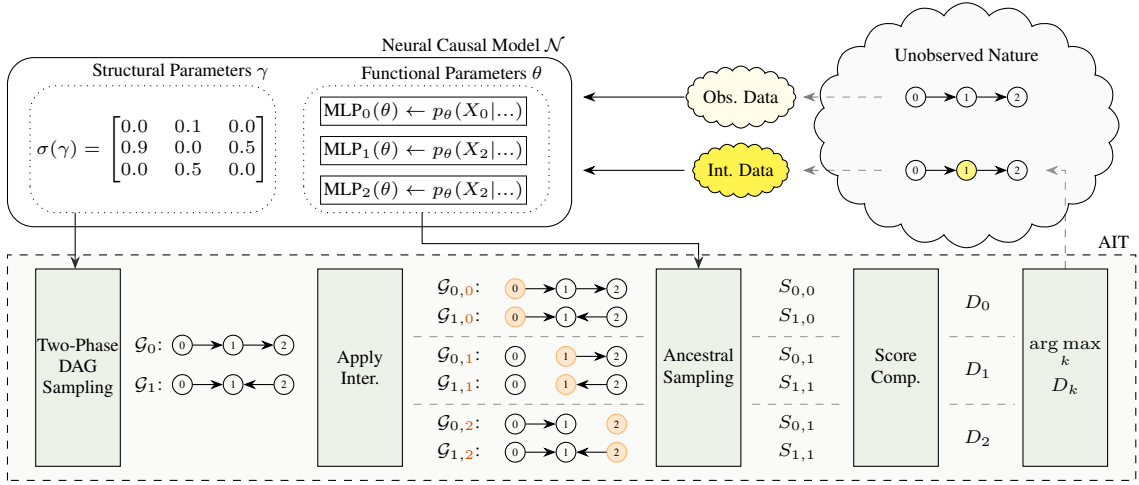

Existing neural causal discovery methods focus on fixed datasets of either observational \citep{zheng2018dags, yu2019dag, zheng2020learning, bengio2019meta, lorch2021dibs, annadani2021variational, cundy2021bcd} or fused (observational and interventional) nature \citep{ke2019learning, brouillard2020differentiable, lippe2021efficient}. While having access to interventional data can significantly improve the identification of the underlying causal structure, the improvement critically depends on the nature of the experiments and the number of interventional samples available to the learner \cite{heckerman1995learning, eberhardt2012number}. However, interventions tend to be costly and can be technically impossible or even unethical \cite{Peters2011b}. Hence it is desirable for an agent to conduct active interventions to recover the underlying causal structure in an adaptive and efficient manner.  While a large body of work has addressed this need based on non-differentiable frameworks \citep{he2008active, eberhardt2012almost, hyttinen2013experiment, hauser2014two, shanmugam2015learning, kocaoglu2017experimental, kocaoglu2017cost, lindgren2018experimental, ghassami2018budgeted, ghassami2019interventional, greenewald2019sample, squires2020active, murphy2001active, tong2001active, masegosa2013interactive, cho2016reconstructing, ness2017bayesian, agrawal2019abcd, zemplenyi2021bayesian, gamella2020active}, existing work in neural causal discovery has not yet focused on incorporating active interventions.

In this work, we propose to augment neural causal discovery methods with the ability to actively intervene. Therefore, we introduce Active Intervention Targeting (AIT), an adaptive intervention design technique for the batch-wise acquisition of interventional samples. AIT can be easily incorporated into any neural causal discovery method which provides access to structural and functional parameters (see \Cref{fig:AIT_page1}). In AIT, we decide where to intervene by computing a score for all possible intervention targets (over a single or multiple variables). This score provides us with an estimate how informative an intervention at that target would be with respect to the current evidence. For a set of hypothesis graphs sampled from the structural belief and a \emph{fixed} intervention target, we apply the intervention on all hypothesis graphs and generate hypothetical samples through an ancestral sampling process based on the functional parameters. This allows us to compare statistics of the post-interventional sample distributions across the hypothesis graphs (see \Cref{fig:AIT_in_detail}). We conjecture (and empirically show) that interventions that do not agree across different hypothesis graphs contain more information about the causal structure and hence enable more efficient learning. 

\textbf{Summary of Empirical Results.} We propose an 
intervention design method (single and multi-target) which identifies the underlying graph efficiently and can be used for any differentiable causal discovery method. We examine the proposed intervention-targeting method across multiple differentiable causal discovery frameworks in a wide range of settings and demonstrate superior performance against established competitive baselines on multiple benchmarks from simulated to real-world data. We provide empirical insights on the distribution of selected intervention targets and its connection to the topological order of the variables in the underlying data-generating distribution.

\vspace{-3mm}
\section{Preliminaries}

\textbf{Structural Causal Model (SCM).} An SCM \citep{peters2017elements} is defined over a set of random variables $X_1, \ldots, X_M$ or just $X$ for short, associated with a directed acyclic graph (DAG) $G=(V,E)$ over  variable nodes $V=\{1, \ldots M\}$. The random variables are connected by edges in $E$ via functions $f_i$ and jointly independent noise variables $U_i$ through
$X_i \:= f_i(X_{pa(i)}, U_i)$
where $X_{pa(i)}$ are $X_i$'s parents in $G$, and directed edges in the graph represent direct causation. The conditionals $P(X_i|X_{pa(i)})$ define the conditional distribution of $X_i$ given its parents. This characterization entails a factorization of the joint observational distribution:
\vspace{-2mm}
\[
    \footnotesize
    P(X_1, \ldots, X_N) = \prod_{i=1}^{N} P(X_i|X_{pa(i)}) 
\]
\vspace{-5mm}

\textbf{Interventions.}
Interventions on $X_i$ change the conditional distribution of $P(X_i|X_{pa(i)})$ to a different distribution, hence affecting the outcome of $X_i$. Interventions can be perfect (hard) or imperfect (soft). Hard interventions entirely remove the dependencies of a variable $X_i$ on its parents $X_{pa(i)}$, hence defining the conditional probability distribution of $X_i$ by some $\tilde{P}(X_i)$ rather than $P(X_i|X_{pa(i)})$. A more general form of intervention is the soft intervention, where the intervention changes the effect of the parents of $X_i$ on itself by modifying the conditional distribution from $P_i(X_i|X_{pa(i)})$ to an alternative, denoted $\tilde{P_i}(X_i|X_{pa(i)})$.

\textbf{Neural Causal Discovery from Fused Data.}
\label{sec:SDI}
Neural causal discovery from fused data aims at fitting fused data with a neural causal model $\mathcal{N}$, an SCM with smoothly differentiable parameters of functional and structural nature, using a score-based objective. Structural parameters $\gamma$ encode our belief in the underlying graph structure $G$, usually in form of a learned soft-adjacency matrix representing a distribution over graphs. Functional parameters $\theta$ encode the conditional probability distributions (CPDs) $P(X_i|X_{pa(i})$ through neural networks that either learn parameters of a distributional family (e.g. Gaussians or normalizing flows \cite{rezende2015variational}) or approximate the function itself. This is usually realized by a stack of MLPs, i.e. one MLP per variable, to represent its conditional distribution.

We evaluate our proposed intervention design method under two frameworks that can handle fused data. While we focus on \emph{Structure Discovery from Interventions (SDI)} \citep{ke2019learning} in the main text, 
we provide futher context and demonstrate AIT's ability based on \emph{Differentiable Causal Discovery from Interventional Data (DCDI)} \citep{brouillard2020differentiable} in $\S$\ref{sec:appendix_dcdi}.

\begin{figure}[t!]
 \begin{center}
    \includegraphics[width=0.42\textwidth]{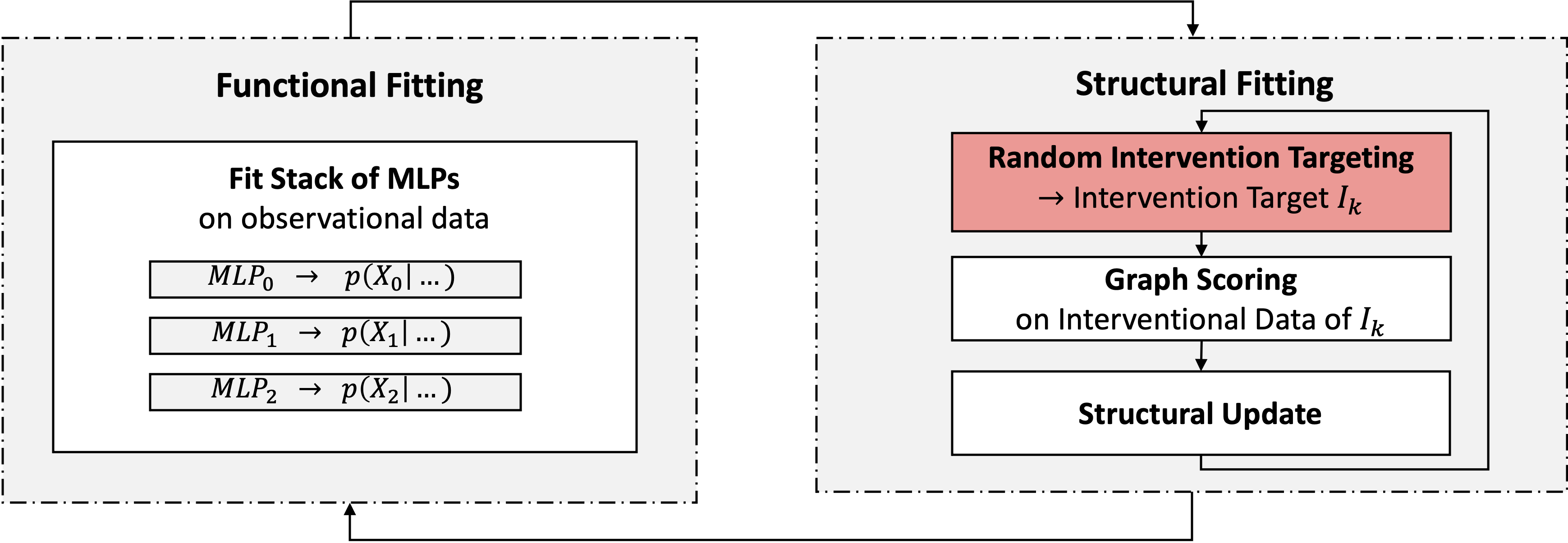}
 \end{center}
 \vspace{-1\baselineskip}
 \caption{SDI learns a neural causal model from fused data using alternating phases of functional and structural fitting.}
 \label{fig:SDI}
 \vspace{-2\baselineskip}
\end{figure}

\textbf{Structure Discovery from Interventions (SDI).} The SDI approach reformulates the problem of causal discovery from \emph{discrete} data as a discrete optimization problem using neural networks. The framework proposes to learn the parameters of a neural causal model using a two-stage training procedure with alternating phases of optimization (see \Cref{fig:SDI}). Under a fixed structural belief, the \emph{functional fitting} stage fits the functional parameters $\theta$ (representing the observational CPDs) to observational data. In order to account for the stochastic nature of the structural belief, the method samples different hypothesized graphs in this stage and uses them in a dropout-like fashion to mask out all variables except the direct causal parents according to the graph while fitting the functional parameters. This enforces the CPDs to be trained on different sets of parents and will converge to the set of true parents as the structure converges. On the other hand, the \emph{structural fitting} freezes the functional parameters and evaluates the fit to interventional data of different hypothesized graphs. The adaptation scores are then used to update the belief in the graph structure by propagating them to update the structural parameters. The method performs competitively to many other methods. However, it processes all interventions in a random and independent manner, a strategy that scales poorly to larger graphs.


\vspace{-2mm}
\section{Active Intervention Targeting (AIT)}
We present a score-based, adaptive intervention design strategy, called AIT, which is applicable to any neural causal discovery method which provides access to structural and functional parameters.  In addition, we present a scalable two-stage DAG sampling technique for the efficient generation of hypothesis DAGs based on a soft-adjacency matrix, which is a common parametrization of the structural belief. Finally, we show how our proposed method can be easily plugged into recent differentiable causal discovery frameworks for guided exploration using interventional data.

\textbf{Assumptions.} 
The proposed method does not have to assume causal sufficiency per se. However, it inherits the assumptions of the selected base framework, and this may include causal sufficiency depending on the base algorithm of choice. In case the underlying framework can handle unobserved variables and offers a generative method for interventional samples, then our method is also applicable

\vspace{-2mm}
\subsection{A score for intervention targeting}
\label{sec:method_score}
Given a structural belief state $\gamma$ with its corresponding functional parameters $\theta$, and a possible set of intervention targets $I$ (single and multi-node intervention targets), we wish to select the most \textit{informative} intervention target(s) $I_{k^*} \in I$ to identify as quickly as possible the underlying structure. In AIT, we decide where to intervene by computing a score for all possible intervention targets. This score provides us with an estimate how informative an intervention at that target would be with respect to the current evidence. We claim that such informative interventions would yield relatively high discrepancies between post-interventional samples drawn under different hypothesis graphs, making it possible to discriminate better among these candidate graphs and indicating larger uncertainty about the intervention target's relation to its parents and/or children. 

We thus construct an F-test-inspired score to seek the target $I_{k^*}$ exhibiting the highest discrepancies between post-interventional sample distributions generated by likely graph structures under fixed functional parameters $\theta$. In order to compare sample distributions over different graphs, we distinguish between two sources of variation: variance \textit{between graphs} (\texttt{VBG}) and variance \textit{within graphs} (\texttt{VWG}). While \texttt{VBG} characterizes the variance of sample means over multiple graphs, \texttt{VWG} accounts for the sample variance when a specific graph is fixed. 
We mask the contribution of the intervened variables $I_k$ to \texttt{VBG} and \texttt{VWG}, and construct our discrepancy score $D$ as a ratio  $D = \frac{\texttt{VBG}}{\texttt{VWG}}$.

This discrepancy score attains high values for intervention targets of particular interest. While \texttt{VBG} itself indicates for which intervention targets the model is unsettled about, an extension to the proposed variance ratio enables more control over the region of interest. Given a fixed set of graphs $\mathcal{G}$ and a fixed interventional sample size across all graphs, let us assume a scenario where multiple intervention targets attain high \texttt{VBG}. Assessing \texttt{VWG} allows us to distinguish between two extreme cases: (a) targets with sample populations that exhibit large \texttt{VWG} (b) targets with sample populations that exhibit low \texttt{VWG}. While high \texttt{VBG} in (a) might be induced by an insufficient sample size due to high variance in the interventional distribution itself, (b) clearly indicates high discrepancy between graphs and should be preferentially studied. 

\vspace{-2mm}
\paragraph{Computational Details.} We begin by sampling a set of graphs $\mathcal{G}=\{\mathcal{G}_i\}, \, i = 1,2,3,\ldots$ from our structural parameters $\gamma$. This $\mathcal{G}$ will remain fixed for all considered interventions for the current experimental round. Then, we fix an intervention target $I_k$ and apply the corresponding intervention to $\theta$, resulting in partially altered functional parameters $\theta_k$ where some conditionals have been temporarily changed to be overriden by the intervention. Next, we draw interventional samples $\smash{S_{i,k}}$ from $\theta_k$ on the post-interventional graphs $\mathcal{G}_{i,k} $ (i.e. intervention on target $I_k$ applied to graph $\mathcal{G}_i$). In the variance calculation, we set the variables of the intervention targets $I_k$ to zero to mask off their contribution to the variance. Having collected all samples over the considered graphs for the specific intervention target $I_k$, we compute $\smash{\texttt{VBG}_k}$ and $\smash{\texttt{VWG}_k}$ as follows:
\vspace{-1mm}
\[
\footnotesize
\begin{split}
   \texttt{VBG}_k &= \sum_i < \big(\mu_{i,k} - \bar{\mu}_{k}\big), \big(\mu_{i,k} - \bar{\mu}_{k}\big) > \\ 
   \texttt{VWG}_k &= \sum_i \sum_j <\big(\big[S_{i,k}\big]_{j} - \mu_{i,k}\big), \big(\big[S_{i,k}\big]_{j} - \mu_{i,k}\big)>
\end{split}
\]
\vspace{-1mm}
where $\bar{\mu}_k$ is a vector of the same dimension as any sample in $S$ and denotes the overall sample-mean over all graphs in the interventional setting $I_k$. Further, $\mu_{i,k}$ denotes the mean of samples drawn from graph $\mathcal{G}_{i,k}$ and $\big[S_{i,k}\big]_{j}$ is the $j$-th sample of the $i$-th graph configuration under intervention $I_k$. Finally, we construct the discrepancy score $D_k$ of $I_k$ as: 
\vspace{-1mm}
$$
D_k \leftarrow \frac{\texttt{VBG}_k}{\texttt{VWG}_k}.
$$
In contrast to the original definition of the F-Score, we can ignore the normalization constants due to equal group size and degree-of-freedoms. 
An outline of the method is provided in Algorithm \ref{algo:AIT}.

\setlength{\textfloatsep}{5pt}
            \begin{algorithm}
                \small
                \caption{Active Intervention Targeting (AIT)}\label{algorithm}
                \textbf{Input:}  Functional Parameters $\theta$,  Structural Parameters $\gamma$, \\ \hspace*{9mm} Interventional Target Space $I$ \\
                \textbf{Output:} Intervention Target $I_{k^*}$
                \begin{algorithmic}
                    \State{$\mathcal{G}$ $\leftarrow$ Sample a set of hypothesis graphs from $\gamma$}
                    \For{each intervention target $I_k$ in $I$}
                        \State $\theta_k$ $\leftarrow$ Perform intervention $I_k$ on $\theta$
                        \For{\textbf{each} graph $\mathcal{G}_i$ \textbf{in} $\mathcal{G}$}
                            \State{$\mathcal{G}_{i,k}$ $\leftarrow$ Apply intervention $I_k$ to $\mathcal{G}_{i,k}$ }
                            \State{$S_{i,k}$ $\leftarrow$ Draw samples from $\mathcal{G}_{i,k}$ using $\theta_k$ }
                            \State{$S_{i,k}$ $\leftarrow$ Set variables in $I_k$ to 0}
                        \EndFor
                        \State{$D_k$ $\leftarrow$ $\frac{
                        \sum_i < \big(\mu_{i,k} - \bar{\mu}_{k}\big), \big(\mu_{i,k} - \bar{\mu}_{k}\big) > }{  \sum_i \sum_j <\big(\big[S_{i,k}\big]_{j} - \mu_{i,k}\big), \big(\big[S_{i,k}\big]_{j} - \mu_{i,k}\big)>}$
                        }
                    \EndFor
                    \State{Target Intervention $I_{k^*}$ $\leftarrow$ $\argmax_k(D_k)$}
                \end{algorithmic}
                \label{algo:AIT}
            \end{algorithm}

\begin{figure*}[ht!]
    \vspace{-0.5\baselineskip}
    \begin{center}
        \includegraphics[width=0.95\textwidth]{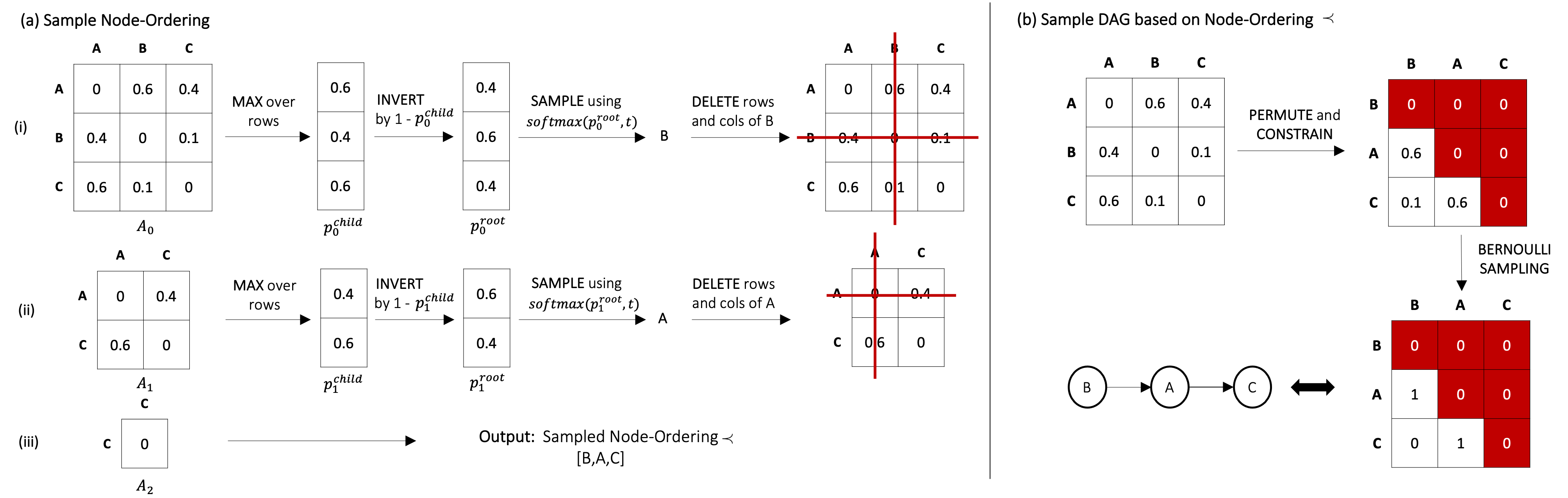}
    \end{center}
    \vspace{-0.8\baselineskip}
    \caption{Two-Stage DAG Sampling: Based on a soft-adjacency $\sigma(\gamma)$, we sample a topological node ordering from an iterative refined score which is repeatedly computed until we have processed all nodes of the graph. We proceed by permuting $\sigma(\gamma)$ according to the drawn node ordering and constrain the upper triangular part to ensure DAGness. Finally, we take independent Bernoulli draws of the unconstrained edge beliefs and arrive at a sampled DAG.}
    \label{fig:DAGSampling}
    \vspace{-1.2\baselineskip}
\end{figure*}

\subsection{Two-Phase DAG sampling}
\label{sec:method_DAGsampling}
Embedding AIT into recent differentiable causal discovery frameworks requires a graph sampler that generates a set of likely graph configurations under the current graph belief state. However, drawing samples from unconstrained graphs (e.g. partially undirected graphs or cyclic directed graphs) is an expensive multi-pass process. Here, we thus constrain our graph sampling space to DAGs. Since most differentiable causal structure learning algorithms learn edge beliefs in the form of a soft-adjacency matrix, we present a scalable, two-stage DAG sampling procedure which exploits structural information of the soft-adjacency matrix beyond independent edge confidences (see Figure \ref{fig:DAGSampling} for a visual illustration). More precisely, we start by sampling topological node orderings from an iterative refined score and construct DAGs in the constrained space by independent Bernoulli draws over possible edges. We can thus guarantee DAGness by construction and do not have to rely on expensive, non-scalable techniques such as rejection sampling or Gibbs sampling. The overall method is inspired by topological sorting algorithms of DAGs where we iteratively identify nodes with no incoming edges, remove them from the graph and repeat until all nodes are processed.

\vspace{-1mm}
\textbf{Soft-Adjacency.} Given a learnable graph structure $\gamma \in \mathbb{R}^{N \times N}$ of a graph over $N$ variables, the soft-adjacency matrix is given as $\sigma(\gamma) \in [0,1]^{N \times N}$ such that $\sigma{(\gamma_{ij})} \in [0,1]$ encodes the probabilistic belief in random variable $X_j$ being a direct cause of $X_i$, where $\sigma(x) = (1+\exp(-x))^{-1}$ denotes the sigmoid function. For the ease of notation, we define $A = \sigma(\gamma)$ and $A_l$ denotes the considered soft-adjacency $\sigma(\gamma)$ at iteration $l$. Note that the shape of $A_l$ changes through the iterations.

\vspace{-1mm}
\textbf{Sample node orderings.} For the iterative root sampling procedure, we start at iteration $l=0$ with an initial soft-adjacency $A_l = A$ and apply the following routine for $N$ iterations. We take the maximum over rows of $A_l$, resulting in a vector of independent probabilities $p_l^{child}$, where $p_l^{child}(i)$ denotes the maximal probability of variable $X_i$ being a child of any other variable at the current belief state. After taking the complement $p_l^{root} = 1-p_l^{child}$, we arrive at $p_l^{root}$ where $p_l^{root}(i)$ denotes the approximated probability of variable $X_i$ being a root node in the current round. In order to arrive at a normalized distribution to sample a root node, we apply a temperature-scaled softmax:
\vspace{-1mm}
\[
\small
    p_l(i) = \mathrm{softmax}(p_l^{root}/t)_i = \frac{\exp\big[p_l^{root}(i)/t\big]}{\sum_{j}^{ }\exp\big[p_l^{root}(j)/t\big]}
\]
where $t$ denotes the temperature. The introduction of temperature-scaling allows to control the distribution over nodes and account for the entropy of the structural belief. We proceed by sampling a (root) node as $r_l \sim Categorical(p_l)$ and delete all corresponding rows and columns from $A_l$ and arrive at a shrinked soft-adjacency $A_{l+1} \in [0,1]^{(N-l-1) \times (N-l-1)}$ over the remaining variables. We repeat the procedure until we have processed all nodes and have a resulting topological node ordering $\prec$ of $[r_0, ..., r_{N-1}]$.

\textbf{Sample DAGs based on node orderings.} Given a node ordering $\prec$, we permute the soft-adjacency $A$ accordingly and constrain the upper triangular part by setting values to $0$ to ensure DAGness by construction (as shown in Figure \ref{fig:DAGSampling}). Finally, we sample a DAG by independent Bernoulli draws of the edge beliefs, as proposed in \citet{ke2019learning}.

\subsection{Applicability to SDI} 
\label{sec:applicability_SDI}
Before integrating our method into the SDI framework, we must choose/design a graph sampler based on SDI's graph belief characterization and define a sampling routine to generate interventional samples under a given state of the structural and functional parameters. SDI offers a learnable graph structure over $N$ variables with $\gamma \in \mathbb{R}^{N \times N}$ such that $\sigma{(\gamma)} \in [0,1]^{N \times N}$ encodes the soft-adjacency matrix. This formulation naturally suggests the application of the introduced \emph{two-phase DAG sampling} to generate hypothetical DAGs under current beliefs. Under these acyclic graph configurations, one may then apply an intervention to SDI functional parameters $\gamma$ and sample data using ancestral sampling. 
SDI's architectural choices allow a seamless integration of AIT into the \emph{structural fitting} stage of SDI, where graphs are evaluated using interventional data to update the structural belief. See \S\ref{sec:SDI} for a compact description of the base framework.


\begin{figure*}[!t]
\minipage{0.30\textwidth}
    \centering
    \includegraphics[trim=0 0 0 0, clip,width=0.9\textwidth]{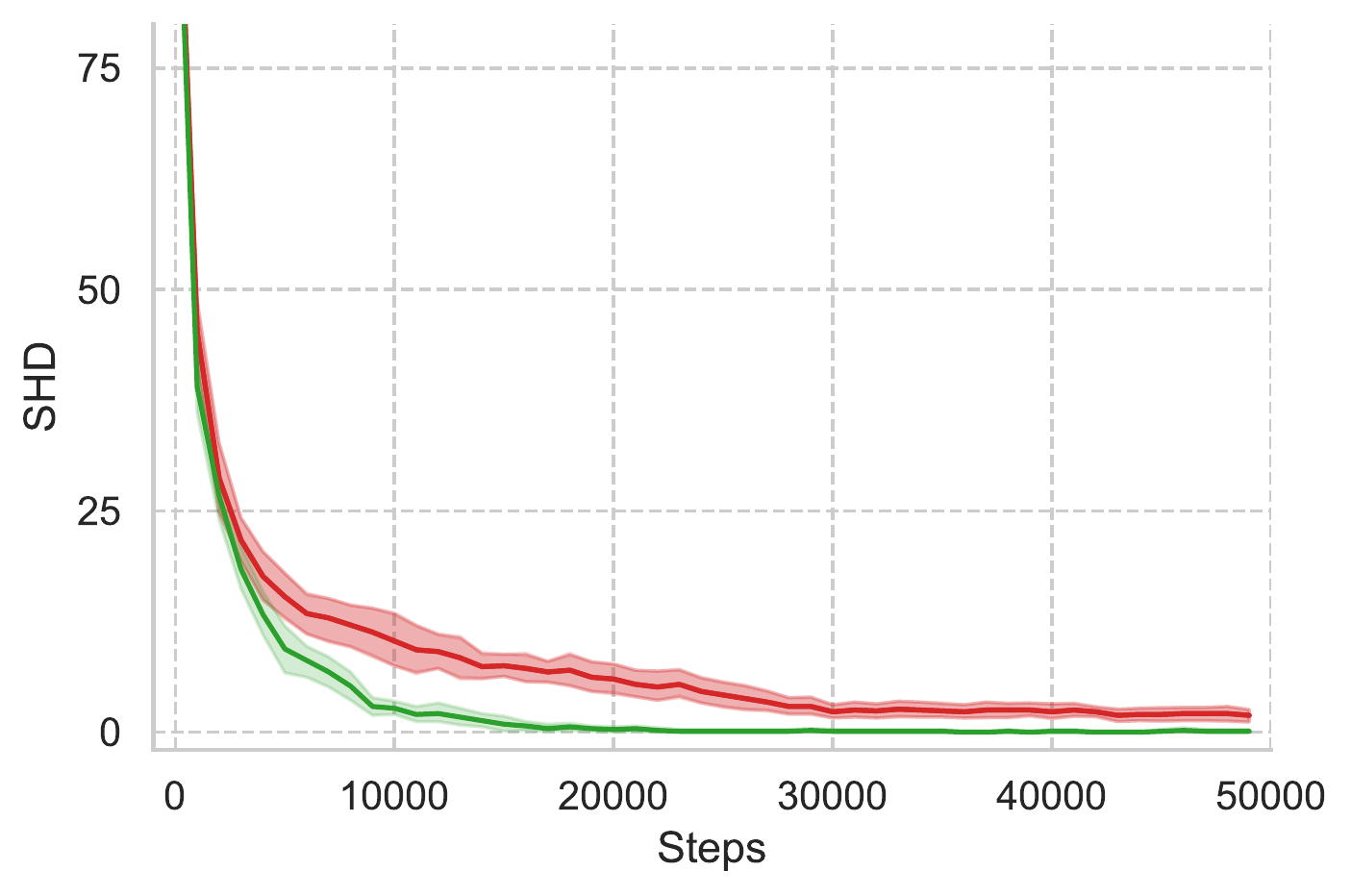}\\
    (a) \texttt{ER-1:}
\endminipage\hfill
\minipage{0.30\textwidth}
    \centering
    \includegraphics[trim=0 0 0 0, clip,width=0.9\textwidth]{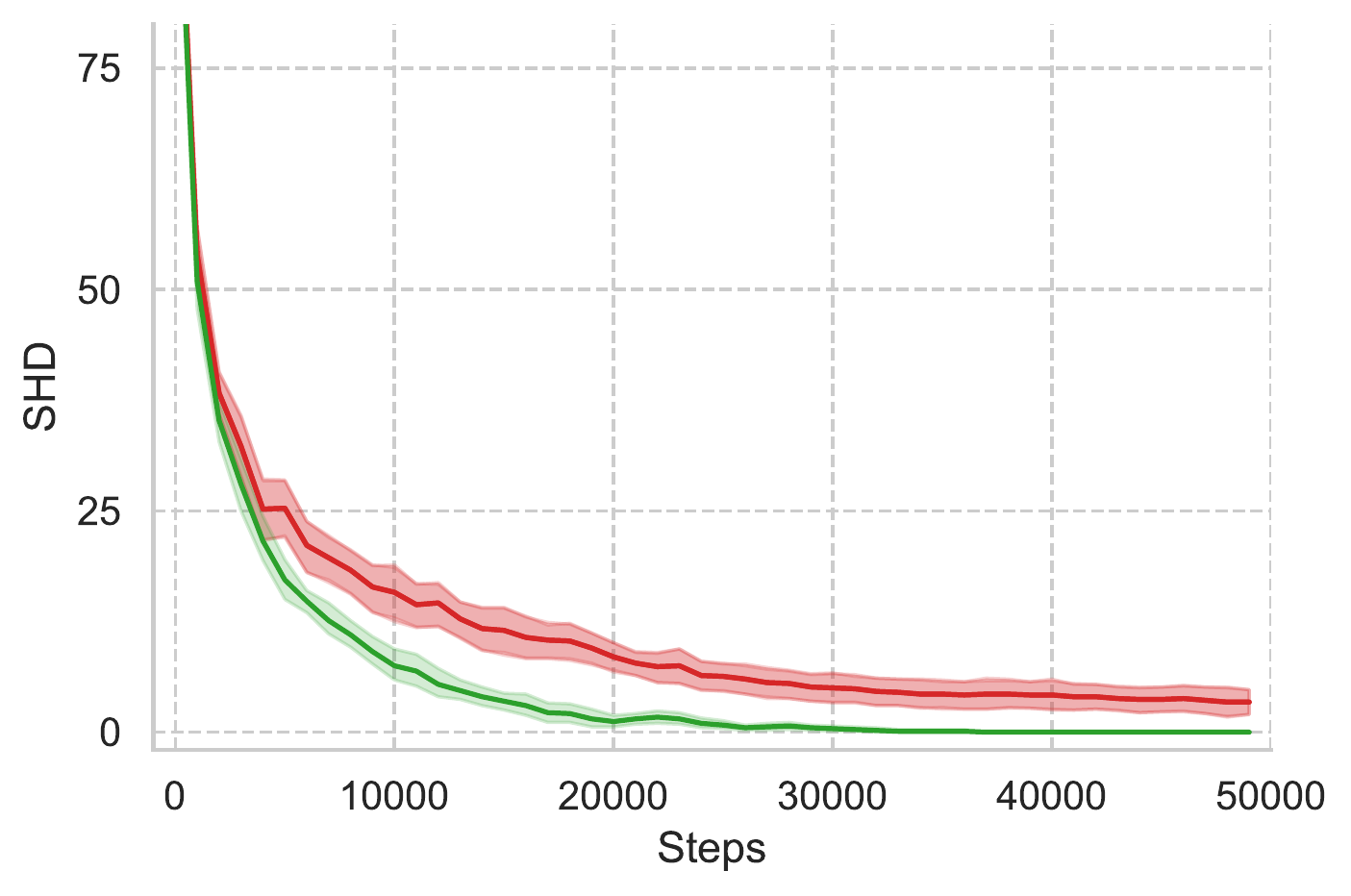}\\
    (b) \texttt{ER-2:}
\endminipage\hfill
\minipage{0.30\textwidth}%
    \centering
    \includegraphics[trim=0 0 0 0, clip,width=0.9\textwidth]{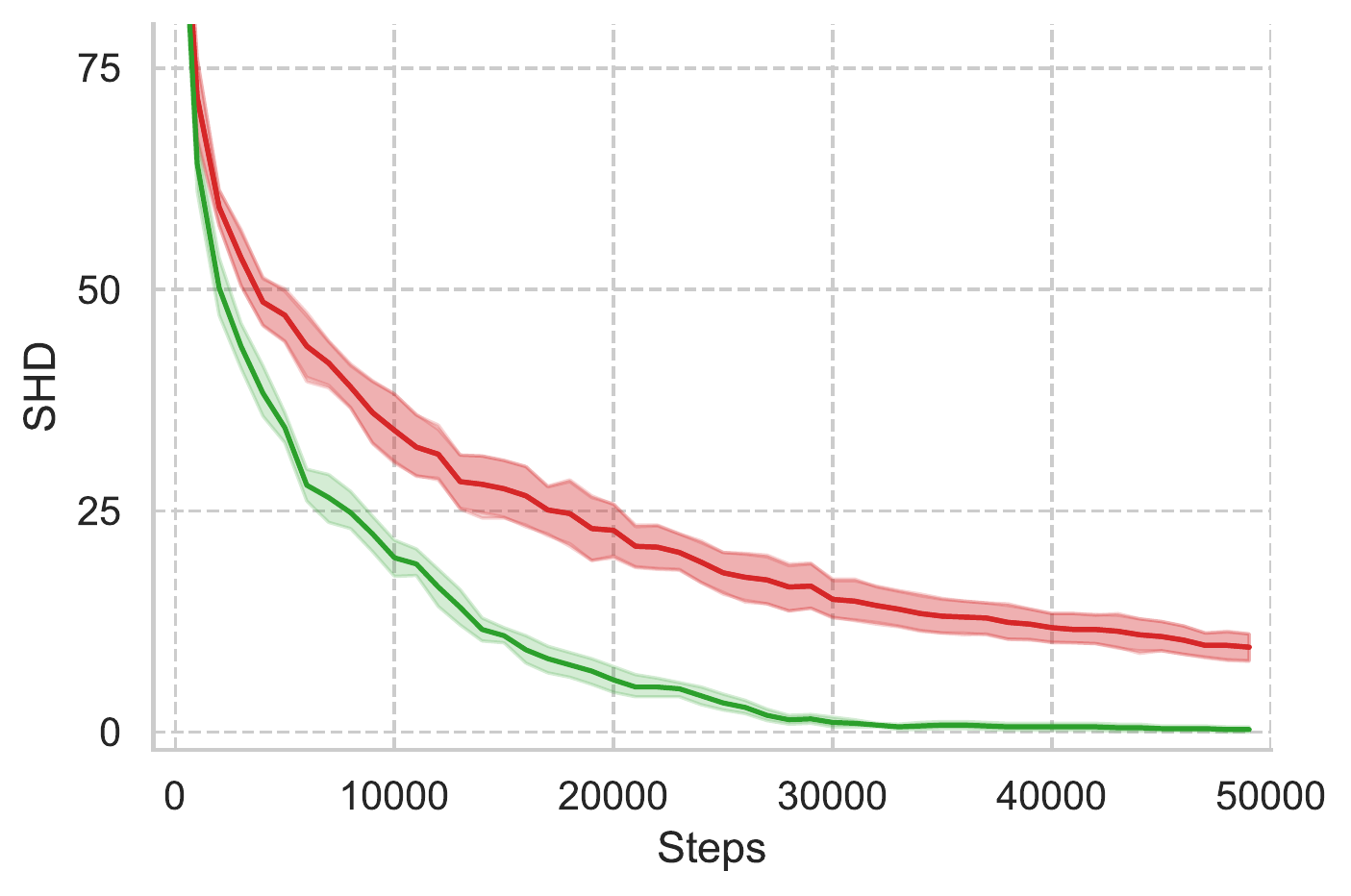}\\
    (c) \texttt{ER-4:}
\endminipage
\caption{SDI with active intervention targeting (green) leads to superior performance over random intervention targeting (red) on random graphs of size 15. The performance gap becomes more significant with increasing edges density. The plot shows average performance in terms of SHD. Error bands were estimated using 10 random ER graphs per setting.}
\label{fig:performance_ER-Graphs}
\vspace{-0.7\baselineskip}
\end{figure*}

\vspace{-2mm}
\section{Experiments}
\vspace{-1mm}
We evaluate AIT on single-target interventions under two different settings: SDI \citep{ke2019learning} and DCDI \citep{brouillard2020differentiable}. We investigate the impact of AIT under both settings with respect to identifiability, sample complexity, and convergence behaviour compared to random targeting  where the next intervention target is chosen independent of the current  evidence. In a further line of experiments, we analyze the targeting dynamics with respect to convergence behaviour and the distribution of target node selections. 
This section will highlight our results on SDI while also pointing to key findings with respect to DCDI (structural discovery and identifiability). However, the results and analysis of DCDI results have been shifted to the appendix.

\vspace{-2mm}
\paragraph{Evaluation Setup.} A huge variety of SCMs and their induced DAGs exist, each of which can stress causal structure discovery algorithms in different ways. We perform a systematic evaluation over a selected set of synthetic and non-synthetic SCMs (and datasets). We distinguish between synthetic \textit{structured} graphs and \textit{random} graphs, the latter generated from the Erdős–Rényi (ER) model with varying edge densities (see \S\ref{app:synthDatasets} for a detailed description of the setup). For conciseness, we only report results on 15-node graphs in this section for the noise-free synthetic setting for AIT on SDI and on 10-node graphs for the noisy setting for AIT on SDI (discrete data). In addition, we point to key results on 10-node graphs for AIT on DCDI (continuous data) in the main text and provide further results and ablation studies in Appendix. We complete the setup with the Sachs flow cytometry dataset \citep{sachs2005causal} and the Asia network  \citep{lauritzen1988local} to evaluate the proposed method on well-known real-world datasets for causal structure discovery. 

\vspace{-1mm}
\paragraph{Key Findings.} (a) We report strong results for active-targeted structure discovery on both discrete and continuous-valued datasets, outperforming random targeting in all experiments. (b) The proposed intervention targeting mechanism significantly reduces sample complexity with strong benefits for graphs of increasing size and density. (c) The distribution of target selections during graph exploration is strongly connected to the topology of the underlying graph. (d) Our method is capable of identifying informative targets. (e) Undesirable interventions are drastically reduced. (f) When monitoring structured Hamming distance (SHD) throughout the procedure, an ``elbow'' point appears approximately when the Markov equivalence class (MEC) has been isolated. (g) AIT introduces desirable properties such as improved recovery of erroneously converging edges. (h) AIT significantly improves robustness in noise-perturbed environments.

\textbf{Structure discovery: Synthetic datasets.} 
We  evaluate accuracy in terms of Structural Hamming Distance (SHD) \citep{acid2003searching} on a diverse set of synthetic non-linear datasets under both SDI and DCDI, adopting their respective evaluation setups. SDI with AIT outperforms all baselines and SDI with random intervention targeting over all presented datasets (see results in \Cref{tab:shd_synthetic_15}). It enables almost perfect identifiability on all structured graphs of size 15 except for the \texttt{full15} graph, and significantly improves structure discovery of random graphs with varying densities. As the size or density of the underlying causal graphs increases, the benefit of the selection policy becomes more apparent (see Figure \ref{fig:performance_ER-Graphs}). \label{sec:dcdi_experiments}
We also examine the effectiveness of our proposed method for DCDI \citep{brouillard2020differentiable} on non-linear data from random graphs of size 10. Active Intervention Targeting improves the identification in terms of sample complexity and structural identifiability compared with random exploration (see \S\ref{sec:appendix_dcdi} for results and analysis). We observe a clear impact of the targeting mechanisms on the order and frequency of selected interventional targets by the learner. 

\begin{table*}
    \centering
    \makebox[0pt][c]{\parbox{1.0\textwidth}{ 
        \centering
        \begin{minipage}[b]{0.97\hsize}
            \centering
            \caption{SHD (lower is better) on various datasets of synthetic or real-world origin. Structured graphs are sorted in ascending order according to their edge density. Notation: {$\bm^{(*)}$} denotes average SHD over 10 different random graphs / \\ {\bfseries O}: Uses observational data / {\bfseries I}: Uses interventional data / {\bfseries A}: Active approach / {$\bm \partial$}: Differentiable approach \vspace{-0.0\baselineskip}}
            \label{tab:shd_synthetic_15}
            \resizebox{0.95\textwidth}{!}{
                \begin{tabular}{lccccrrrrrrrrrrr}
                    \toprule
                    & \multicolumn{4}{c}{\bfseries Method}
                    & \multicolumn{9}{c}{\bfseries SYNTHETIC (N=15)}
                    & \multicolumn{2}{c}{\bfseries REAL}\\ 
                    \cmidrule(l){6-14} \cmidrule(l){15-16}
                    & \multicolumn{4}{c}{\bfseries Classification}
                    & \multicolumn{6}{c}{\bfseries Structured Graphs} &
                    \multicolumn{3}{c}{\bfseries Random Graphs} & \multicolumn{2}{c}{\bfseries BnLearn} \\
                    \cmidrule(l){2-5} \cmidrule(l){6-11} \cmidrule(l){12-14}  \cmidrule(l){15-16}  
                    & O & I & A & $\partial$ & Chain & Collider & Tree & Bidiag & Jungle & Full  & ER-1$^{(*)}$ & ER-2$^{(*)}$ & ER-4$^{(*)}$  & Sachs & Asia   \\
                    \midrule
                    GES \citep{chickering2002optimal}        & \checkmark & & &  & 13 &  1 & 12  & 14  & 14 & 69     & 8.3 ($\pm 1.9$) & 17.6 ($\pm 4.6$) & 39.4 ($\pm 6.7$)  & 19 &  4 \\
                    NOTEARS \citep{zheng2018dags}            & \checkmark & & & \checkmark & 22 & 21 & 26 & 33  & 35 & 93     & 23.7 ($\pm 4.0$)  & 35.8 ($\pm 5.2$) & 59.5 ($\pm 3.7$) & 22 & 14  \\
                    DAG-GNN \citep{yu2019dag}                & \checkmark & & & \checkmark & 11 & 14 & 15  & 27  & 25 & 97    & 16.0 ($\pm 3.7$) & 30.6 ($\pm 3.4$) & 59.7 ($\pm 4.1$) & 19 & 10 \\
                    \midrule
                    GIES \citep{hauser2012characterization}  & \checkmark & \checkmark & & & 13 &  6 & 10  & 17  & 23 & 60    & 10.9 ($\pm 4.2$) & 18.1 ($\pm 4.3$) & 39.3 ($\pm 5.6$) & 16 & 11  \\
                    ICP \citep{peters2016causal}             & & \checkmark & & & 14 & 14 & 14  & 27  & 26 & 105   & 16.2 ($\pm 3.6$) & 31.1 ($\pm 3.4$) & 60.1 ($\pm 3.9$) & 17 & 8  \\
                    \midrule 
                    A-ICP \citep{gamella2020active}          & & \checkmark & \checkmark & & 14 & 14 & 14  & 27  & 26 & 105   & 16.2 ($\pm 3.6$) & 31.1 ($\pm 3.4$) & 60.1 ($\pm 3.9$)  & 17 & 8   \\
                    \midrule
                    SDI (Random) \citep{ke2019learning}     & \checkmark & \checkmark & & \checkmark & \highlight{0} & \highlight{0} & 2  & 3 & 7 & 24    & 1.4 ($\pm 1.6$)   & 2.1 ($\pm 2.3$)  & 7.2 ($\pm 2.7$)  & \highlight{6} & \highlight{0}  \\
                    SDI  (AIT)                              & \checkmark & \checkmark & \checkmark & \checkmark & \highlight{0}  & \highlight{0} & \highlight{0}     & \highlight{0} & \highlight{0} & \highlight{7}    & \highlight{0.0 ($\pm 0.0$)}   & \highlight{0.0 ($\pm 0.0$)}  & \highlight{0.0 ($\pm 0.0$)} & \highlight{6} & \highlight{0}  \\
                    \bottomrule
                \end{tabular}
            }
        \end{minipage}
}}
\end{table*}

\begin{figure*}[!t]
    \centering
    \includegraphics[trim=0 0 0 20, clip,width=0.9\textwidth]{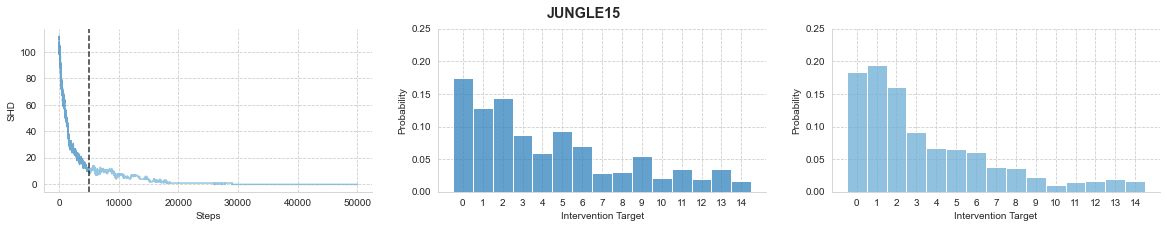}
    \caption{SDI: Dynamics and target distribution of AIT for a structured \texttt{jungle} graph of size 15. The graphs' nodes are sorted in topological order, root node first. The graph is binary-tree-like with 4 levels. For the dense \texttt{jungle15}, the multi-level structure characteristic of the tree-like graph is readily apparent even before the elbow point. Nodes without children are very rarely chosen.
    }
     \label{fig:targetSelection_distribution}
\vspace{-1.1\baselineskip}
\end{figure*}

\textbf{Structure  discovery: flow cytometry and asia dataset.} While the synthetic datasets systematically explore the strengths and weaknesses of causal structure discovery methods, we further evaluate their capabilities on the real-world flow cytometry dataset (also known as Sachs network)\citep{sachs2005causal} and the Asia network \citep{lauritzen1988local} from the BnLearn Repository. SDI with active intervention targeting outperforms all measured baselines and achieves the same result as random targeting in terms of SHD, but with reduced sample complexity.  Despite AIT deviating only by 6 undirected edges from the (concensus) ground truth structure of Sachs et al. \citep{sachs2005causal}, there is some concern about the correctness of this graph and the different assumptions associated with the dataset \citep{mooij2020jointcausal, zemplenyi2021bayesian}. Therefore, perfect identification may not be achievable by any method in practice in the Sachs setting.

\textbf{Effect of intervention targeting on sample complexity.}
\label{sec:results_effectOfAIT}Aside from the significantly improved identification of underlying causal structures, our method allows for a substantial reduction in interventional sample complexity. After reaching the ``elbow'' point in terms of structural Hamming distance, random intervention targeting requires a fairly long time to converge to a solution within the MEC. In contrast, our proposed technique continues to select informative intervention targets beyond the elbow point and more quickly converges to the correct graph within the MEC. The continued effectiveness of our method directly translates to increased sample-efficiency and convergence speed, and is apparent for all examined datasets (see Figure \ref{fig:performance_ER-Graphs}).

\begin{figure*}[!t]
    \centering
    \minipage{0.45\textwidth}
        \centering
        \includegraphics[width=\linewidth]{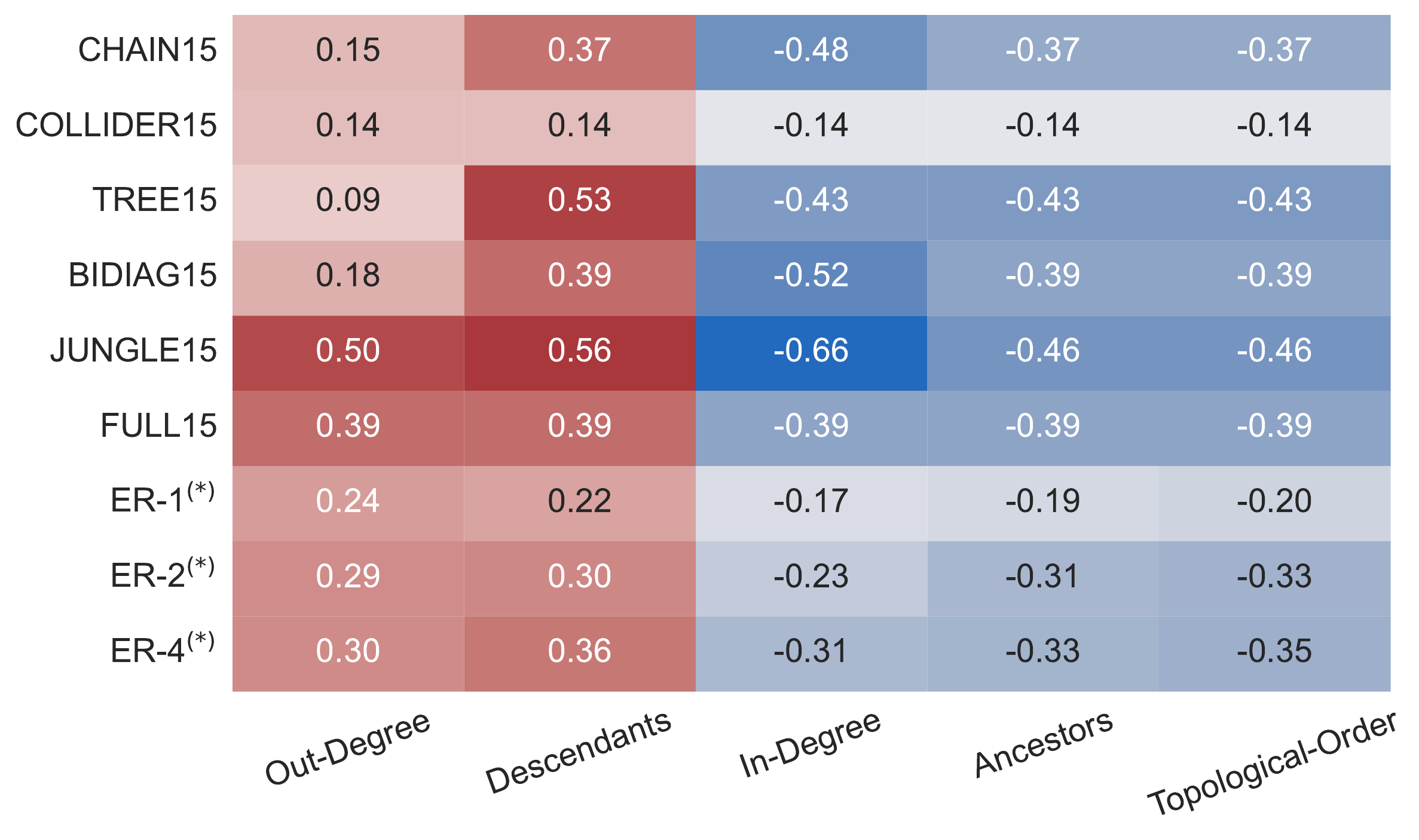}
        (a) Random Targeting
    \endminipage\hfill
    \minipage{0.45\textwidth}
        \centering
        \includegraphics[width=\linewidth]{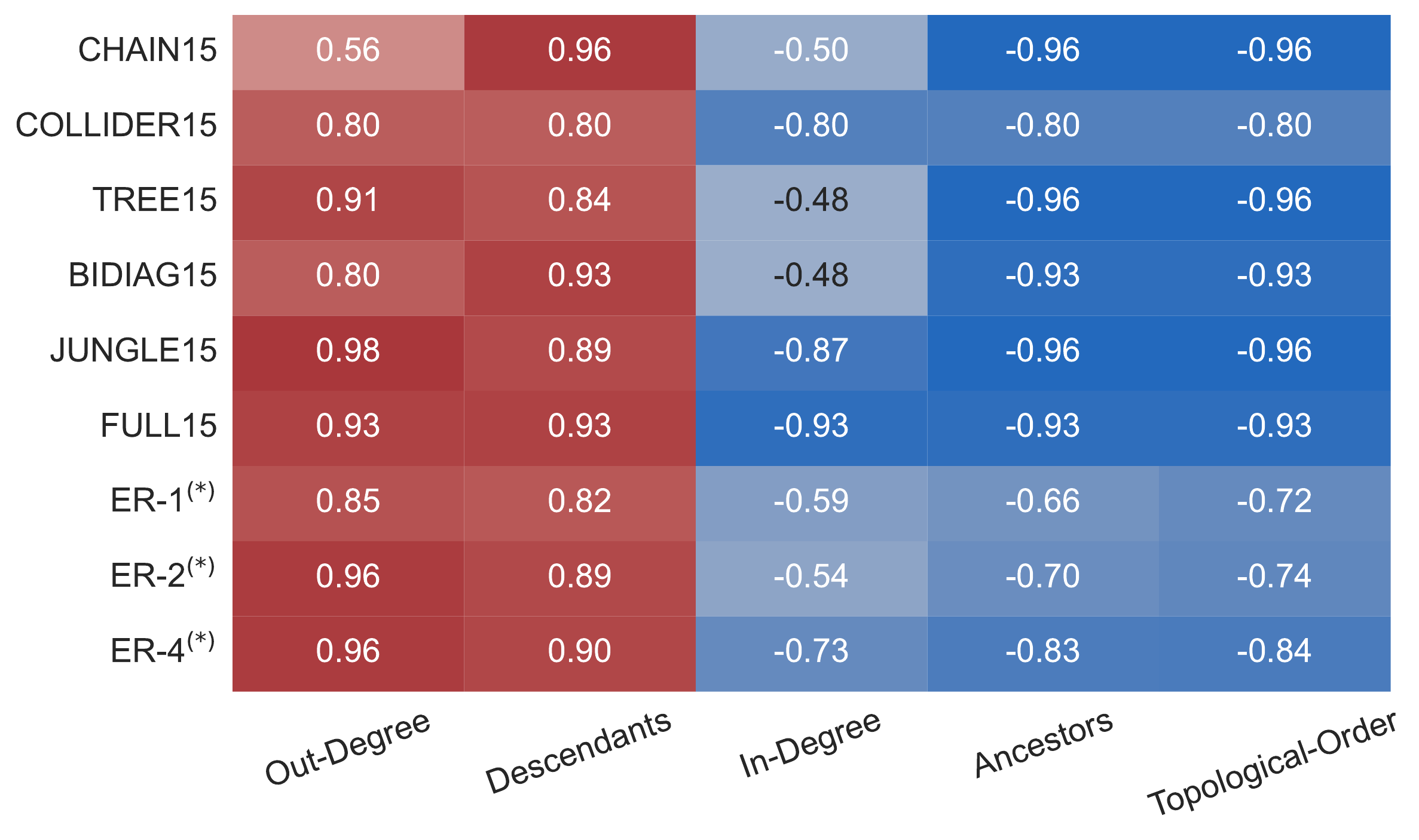}
        (b) Active Intervention Targeting
    \endminipage\hfill
    \vspace{-2mm}
    \caption{Correlation scores between the number of individual target selections and different topological properties of those targets. AIT shows strong correlations with the measured properties over all graphs, which indicates a controlled discovery of the underlying structure through preferential targeting of nodes with greater (downstream) impact on the overall system.}
    \label{fig:correlationAnalysis}
    \vspace{-0.7\baselineskip}
\end{figure*}

\textbf{Distribution of intervention targets.}
The careful study of the behaviour of the proposed method under our chosen synthetic graphs enable us to reason about the method's underlying dynamics. Analysing the dynamics of intervention targeting reveals that the distribution of target node selections is linked to the topology of the underlying graph. More specifically, the number of selections of a given target node strongly correlates with its out-degree and number of descendants in the underlying ground-truth graph structure (see Figure \ref{fig:correlationAnalysis}). That our method prefers interventions on nodes with greater (downstream) impact on the overall system can be most clearly observed in the distribution of target selection on the example of the synthetic \texttt{jungle} graph in Figure \ref{fig:targetSelection_distribution}.

\textbf{Selection of informative targets.} Apart our strong results in the discovery of the underlying causal graph, we demonstrate AIT's general ability of detecting informative intervention targets. In a careful designed empirical, we preinitalize the structural parameters to the ground truth graph but keep one or multiple edges within the skeleton undirected. Over all evaluated settings, AIT rapidly detects the informative intervention targets in order to direct the undirected edges. Detailed results are shown in \S \ref{app:ablation_informativeTargets}.

\textbf{Reduction of undesirable interventions.}
\label{sec:results_undesirableInterventions}
An intervention destroys the original causal influence of other variables on the intervened target variable $I_k$, so its samples cannot be used to determine the causal parents of $I_k$ in the undisturbed system. Therefore, if a variable without children is detected, interventions upon it should be avoided since they effectively result in redundant observational samples of the remaining variables that are of no benefit for causal structure discovery. Active intervention targeting leads to the desirable property that interventions on such variables are drastically reduced (see 
\Cref{fig:targetSelection_distribution}).

 \begin{figure*}[t!]
    \centering
    \minipage{0.22\textwidth}
        \centering
        \includegraphics[width=0.95\linewidth]{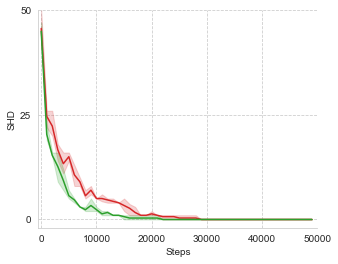}\\
        (a) $\eta = 0$
    \endminipage\hfill
    \minipage{0.22\textwidth}
        \centering
        \includegraphics[width=0.95\linewidth]{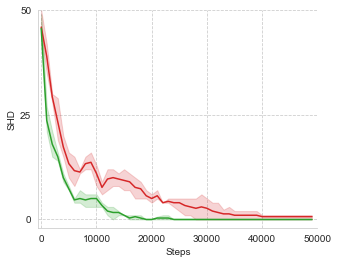}\\
        (b) $\eta = 0.01$
    \endminipage\hfill
    \minipage{0.22\textwidth}
        \centering
        \includegraphics[width=0.95\linewidth]{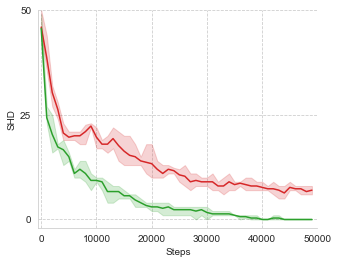}\\
        (c) $\eta = 0.02$
    \endminipage\hfill
    \minipage{0.22\textwidth}
        \centering
        \includegraphics[width=0.95\linewidth]{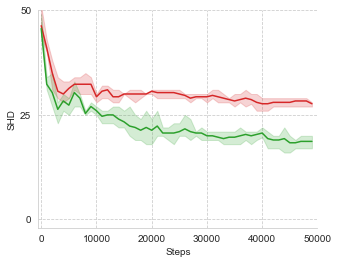}\\
        (d) $\eta = 0.05$
    \endminipage\hfill
    \caption{Convergence Behaviour in terms of SHD for random ER-4 graphs over 10 variables under different noise levels $\eta$, where AIT (green) clearly outperforms Random Targeting (red) over all noise levels. The performance gap becomes of larger magnitude as the noise level increases. Error bands were estimated using 3 random ER graphs per setting.}
   \label{fig:SDI_noise_speed}
   \vspace{-1.1\baselineskip}
\end{figure*}

\textbf{Identification of Markov equivalence class.} Investigating the evolution of the intervention target distribution over time reveals that the causal discovery seems to be divided into two phases of exploration: Phase 1 lasts until the elbow point in terms of SHD, and Phase 2 from the elbow point until convergence (see Figure \ref{fig:performance_ER-Graphs}). We observed over multiple experiments that phase 1 tends to quickly discover the underlying skeleton (removing superfluous connections while keeping some edges undirected), until a belief state $\gamma_{elbow}$ is reached representing a MEC, or a class of graphs very close to a MEC. Phase 2 is predominantly operating on the partially directed skeleton and directs the remaining edges.

\textbf{Recovery of erroneously converging edges.} Recovery of incorrectly-converging edges critically depends on adapting the order of interventions, which a random intervention policy does not. In sharp contrast, intervention targeting significantly promotes early recovery from incorrect assignment of an edge. 
In contrast, the observed edge dynamics and the corresponding graph belief states indicate that the random policy can lock itself into unfavorable belief states from which recovery is extremely difficult, while AIT provides an escape hatch throughout learning.

\textbf{Improved robustness in noise-perturbed environments.}
\label{sec:results_noise}
Considering that noise significantly impairs the performance of causal discovery, we examine the performance of active intervention targeting in noise-perturbed environments with respect to SHD and convergence speed and compare it with random intervention targeting. We conduct experiments under different noise levels in the setting of binary data generated from structured and random graphs of varying density. A noise level $\eta$ denotes the probability of flipping a random variable and applying it to all measured variables of observational and interventional samples. Through all examined settings, we observe that active intervention targeting significantly improves identifiability in contrast to random targeting (see \S \ref{appendix:noise} for detailed results). Active intervention targeting perfectly identifies all structured graphs, except for the collider and full graph, up to a noise level of $\eta=0.05$, i.e. where every 20th variable is flipped. 
The observed performance boost is even more noticeable in the convergence speed, as shown in Fig. \ref{fig:SDI_noise_speed} for ER-4 graphs spanning over 10 variables. While the convergence-gap gets more significant with an increasing noise level, random targeting does not converge to the ground-truth graphs for a noise level higher than $\eta = 0.02$. In contrast, AIT still converges to the correct graph and shows even a convergence tendency for $\eta = 0.05$. These findings support our observation from different experiments that active intervention targeting leads to a more controlled and robust graph discovery. Further experimental results in noise-perturbed environments can be found in \S\ref{appendix:noise}.

\vspace{-2mm}
\section{Conclusion}
\vspace{-2mm}
Promising results have driven the recent surge of interest in differentiable methods for causal structure learning from observational and interventional data. In this work, we augment existing neural causal discovery methods with the ability to actively intervene and propose an active learning method to choose interventions. We show in a systematic empirical study across multiple noise-free and noise-perturbed datasets that active intervention targeting not only improves sample efficiency but also the identification of the underlying causal structures compared to random intervention targeting. Our results indicate that the guided selection of intervention targets leads to a more controlled discovery with favourable properties with respect to the optimization. The increased performance boost for larger graphs is in line with our expectation as random intervention targeting scales poorly to graphs of larger size. 

While our method shows significant improvements with respect to sample efficiency and graph recovery over existing methods across multiple noise-free and noise-perturbed datasets, the number of interventions is not yet optimal \citep{atkinson1975optimal, eberhardt2012number} and can potentially be reduced in future work. Further, in this work, the interventional samples were presented to the evaluated frameworks according to a fixed learning schema (e.g. fixed number of samples for evaluated interventions in graph scoring). It would be interesting to see if the information discovered by AIT could be used for a more adaptive learning procedure to further improve sample efficiency.

\clearpage
\bibliography{main}
\bibliographystyle{styles/icml22/icml2022}

\newpage
\onecolumn
\appendix
\section{Appendix}


\subsection{Related Work}
\label{sec:relatedWork}
 Causal induction can use  either observational and (or) interventional data. With purely observational data, the causal graph is only \textit{identifiable} up to a Markov equivalence class (MEC) \citep{spirtes2000causation}, \textit{interventions} are needed in order to identify the underlying causal graph \citep{eberhardt2007interventions}. Our work focuses on causal induction from fused data (observational and interventional data). 

\textbf{Causal Structure Learning.} There exists several approaches for causal induction from interventional data: score-based, constraint-based, conditional independence test based and continuous optimization. We refer to \citep{heinze2018causal, vowels2021d} for recent overviews.  While most algorithms perform heuristic, guided searches through the discrete space of DAGs, \citet{zheng2018dags} reformulates it as a continuous optimization problem constrained to the zero level set of the adjacency matrix exponential. This important result has driven recent work in the field and showed promising results \citep{kalainathan2018sam, yu2019dag, ng2019graph, lachapelle2020gradient-based, zheng2020learning, Zhu2020Causal}. Due to the limitations of purely observational data, \citet{ke2019learning} and \citet{brouillard2020differentiable} extend the continuous optimization framework to make use of interventional data. \citet{lippe2021efficient} scales in a concurrent work with ours the work of \citep{ke2019learning} to higher dimensions by splitting structural edge parameters in separate orientation and likelihood parameters and leveraging it in an adapted gradient formulation with lower variance. In contrast to \citep{brouillard2020differentiable, ke2019learning} and our work, they require interventional data on every variable. 

\textbf{Active Causal Structure Learning.} Interventions are usually hard to perform and in some cases even impossible~\citep{peters2017elements}. Minimizing the number of interventions performed is desirable. Active causal structure learning addresses this problem, and a number of approaches have been proposed in the literature. These approaches can be divided into those that select intervention targets using graph-theoretic frameworks, and those using Bayesian methods and information gain.

Graph-theoretic frameworks usually proceed from a pre-specified MEC or CPDAG (completed partially directed acyclic graph) and either investigate special graph substructures \citep{he2008active} such as cliques \citep{eberhardt2012almost, squires2020active}, trees \citep{greenewald2019sample}, or they prune and orient edges until a satisfactory solution is reached \citep{ghassami2018budgeted,ghassami2019interventional,hyttinen2013experiment}, perhaps under a cost budget \citep{kocaoglu2017cost,lindgren2018experimental}. Their chief limitation is that an incorrect starting CPDAG can prevent reaching the correct graph structure even with an optimal choice of interventions.

The other popular set of techniques involve sampling graphs from the posterior distribution in a Bayesian framework using MCMC and then selecting the interventions which maximize the information gain on discrete \citep{murphy2001active, tong2001active} or Gaussian \citep{cho2016reconstructing} variables. The drawbacks of these techniques is the difficulty of integrating them with non-Bayesian methods, except perhaps by bootstrapping \citep{agrawal2019abcd}.

In contrast to existing work, our base frameworks do not start from a pre-specified MEC or CPDAG and existing graph-theoretical approaches are hence not directly applicable unless we pre-initalize them with a known skeleton. However, in the case we offer access to a predefined structure in the form of a MEC or CPDAG, a previously directed edge is likely to be inverted during the ongoing process which contradicts with the underlying assumptions of existing approaches. Further, we build atop non-Bayesian frameworks and are therefore limited in applying methods based on information gain which require access to a posterior distribution over graph structures. While bootstrapping would allow us to approximate the posterior distribution over graph structures in our non-Bayesian setting, it is not guaranteed to achieve full support over all graphs since the support is limited to graphs estimated in the bootstrap procedure \citep{agrawal2019abcd}. Furthermore, the computational burden of bootstrap would limit us in scaling to graphs of larger size.

\clearpage
\subsection{Two-Stage DAG Sampling}
\label{sec:appendix_DAGSampling}
\subsubsection{Algorithm Outline}
We present an outline of the proposed two-stage DAG sampling procedure which exploits structural information of the soft-adjacency beyond independent edge confidences. The routine is based on a graph belief state $\gamma$ where $\sigma(\gamma)$ denotes a soft-adjacency characterization. We start by sampling topological node orderings from an iterative refined score and construct DAGs in the constrained space by independent Bernoulli draws over possible edges. We can therefore guarantee DAGness by construction.

The temperature parameter $t>0$ of the temperature-scaled softmax can be used to account for the entropy of the graph belief state. However, in the general setting we suggest to initialize the parameter to $t=0.1$. Note that initializing $t \rightarrow 0$ results in always picking the maximizing argument and $t \rightarrow \infty $ results in an uniform distribution.

\begin{algorithm}[hp!]
    \caption{Two-Stage DAG Sampling}\label{algorithm}
    \textbf{Input:} Graph Belief State $\sigma(\gamma)$ in the form of a soft-adjacency matrix\\
    \textbf{Output:} DAG Adjacency Matrix $A_{Dag}$
    \begin{algorithmic}[1]
        \item[]
        \item[] \Comment{{\bf Phase 1:} Sample Node Ordering $\prec$}
        \State{$A_0 \leftarrow \sigma(\gamma)$}
        \State{$nodes \leftarrow [0, ..., N-1]$}
        \For{$k=0$ to $N-1$}
            \State{$p^{child}_k(i) \leftarrow \max A_k[i,:]$}
            \State{$p_k^{root}(i) \leftarrow 1 - p^{child}_k(i)$}
            \State{$p_k(i) \leftarrow \frac{\exp[p_k^{root}(i)/t]}{\sum_{j}^{ }\exp[p_k^{root}(j)/t]}$}
            \State{$r_k \leftarrow nodes[idx_k]$ where $idx_k \sim Categorical(p_k)$}
            \State{Remove $r_k$ from $nodes$}
            \State{$A_{k+1} \leftarrow A_{k}[nodes, nodes]$}
        \EndFor
        \State{$\prec = [r_0, ..., r_{N-1}]$}
        \item[] \Comment{{\bf Phase 2:} Sample DAG based on node ordering $\prec$}
        \item[]
        \State{$A_{Perm} \leftarrow$ Permute $\sigma(\gamma)$ according to $\prec$}
        \State{$A_{Perm} \leftarrow$ Constrain upper diagonal part by setting values to  0}
        \State{$A_{Ber}\leftarrow \text{Bernoulli}(A_{Perm})$}
        \State{$A_{Dag}\leftarrow$ Apply inverse permutation of $\prec$ to $A_{Ber}$}
    \end{algorithmic}
    \label{algo:DAGSampling}
\end{algorithm}

\subsubsection{Connection to Plackett-Luce distribution \citep{Luce59, plackett1975analysis}} 
Our proposed node ordering sampling routine can be regarded as an extension of the Placket-Luce distribution over node permutations. In contrast, we refine scores in an iterative fashion rather than setting them apriori as we account for previously drawn nodes to estimate the probability of a node being the root node in the current iteration.


\clearpage
\subsection{Experimental Setup}

\label{app:synthDatasets}
A huge variety of SCMs and their induced DAGs exist, each of which can stress causal structure discovery algorithms in different ways. In this work, We perform a systematic evaluation over a selected set of synthetic and non-synthetic SCMs. We distinguish between discrete (based on DSDI \citep{ke2019learning}) or continuous (based on DCDI \citep{brouillard2020differentiable}) valued random variables. Through all experiment, we limit us to 1000 samples per intervention.

\subsubsection{Synthetic Datasets}
\begin{figure}[h!]
    \centering
    \textbf{Graphs with acyclic skeletons:} \\
    \minipage{0.32\textwidth}
        \centering
        \includegraphics[width=0.99\linewidth]{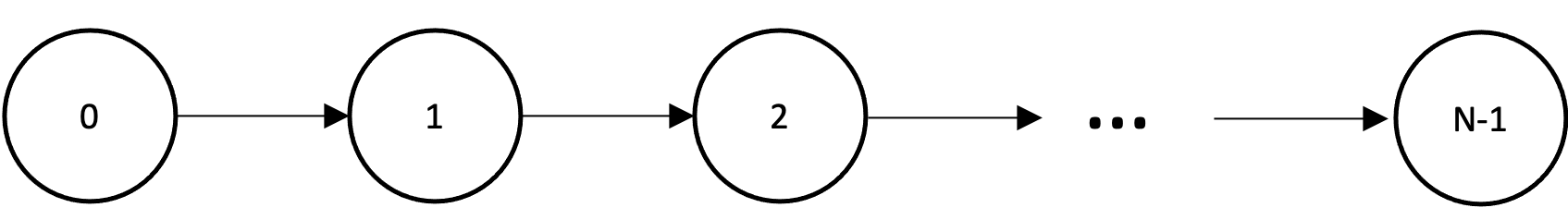}\\
    \endminipage\hfill
    \minipage{0.32\textwidth}
        \centering
        \includegraphics[width=0.99\linewidth]{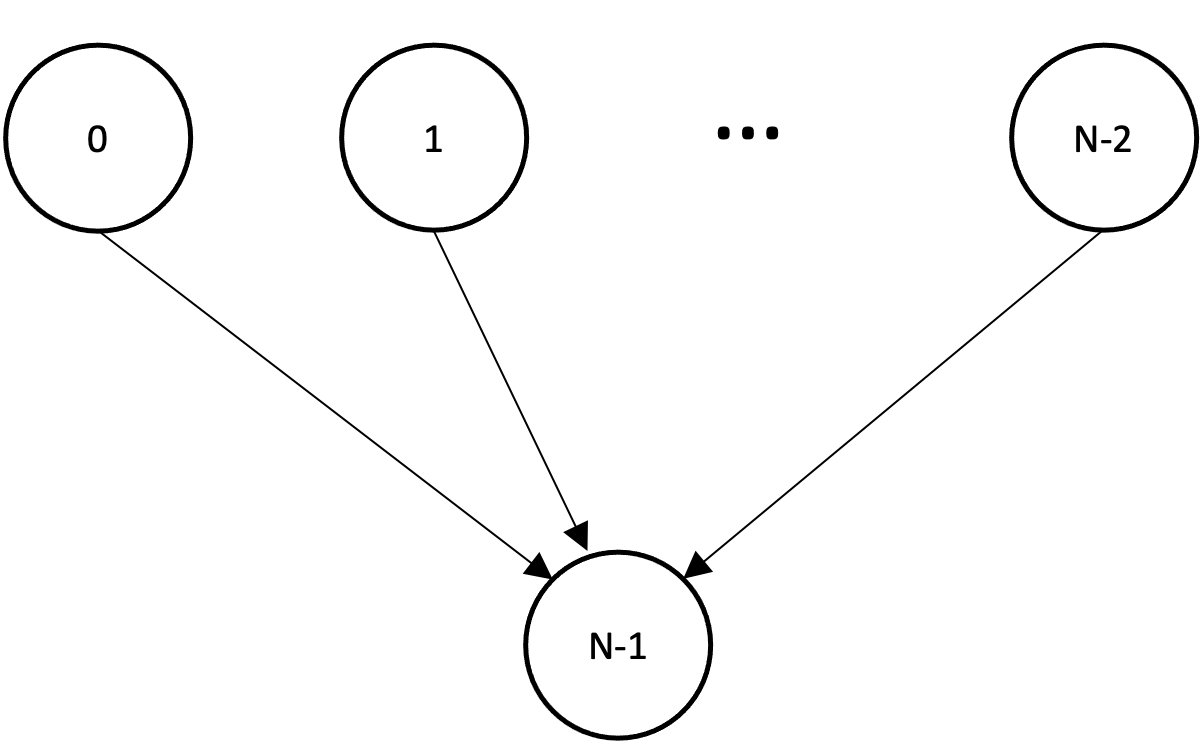}\\
    \endminipage\hfill
    \minipage{0.32\textwidth}
        \centering
       \includegraphics[width=0.99\linewidth]{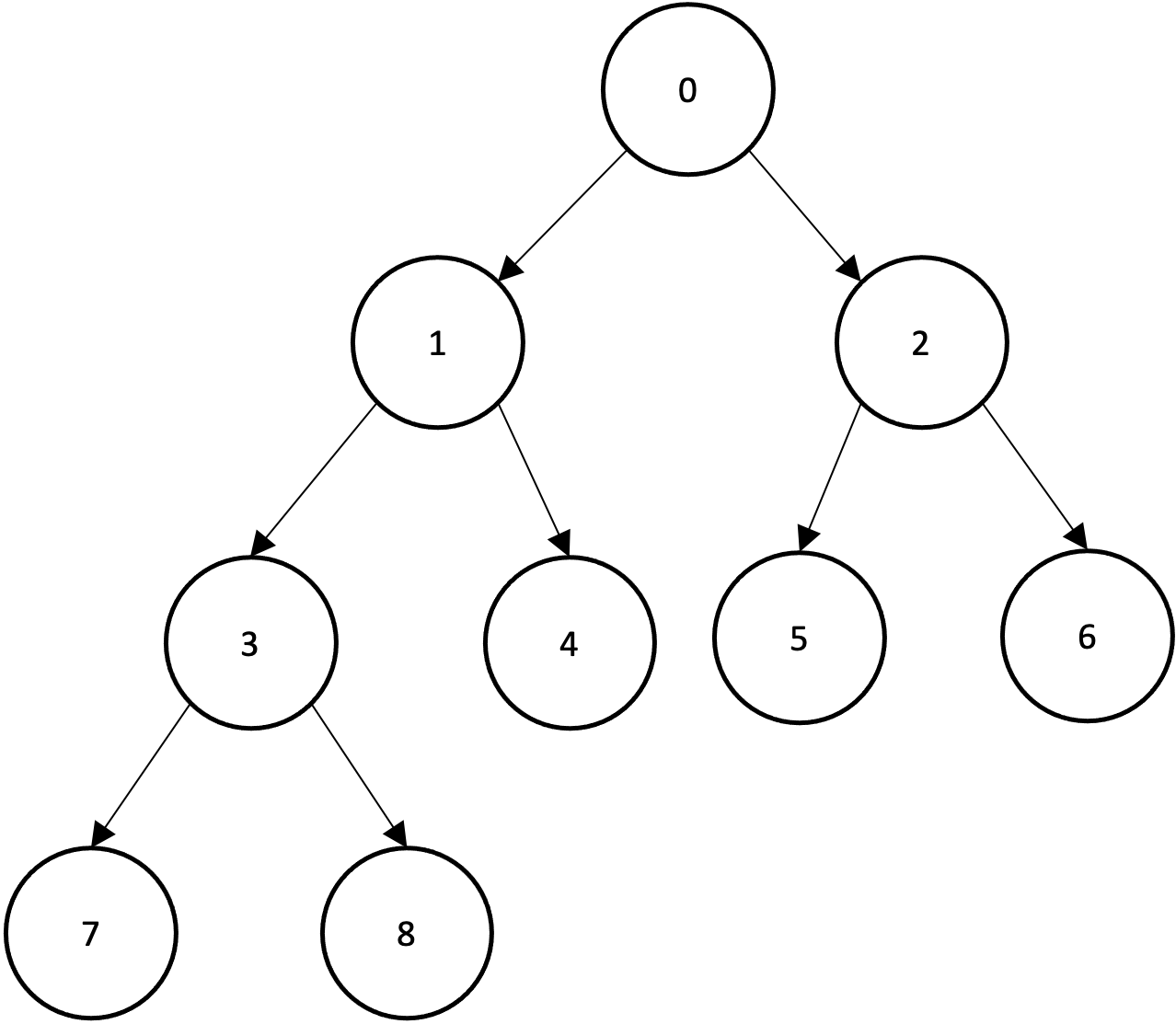}\\
    \endminipage\hfill
    \vspace{3mm}\\
    \minipage{0.32\textwidth}
        \centering
        (a) \texttt{Chain}
        \endminipage\hfill
    \minipage{0.32\textwidth}
        \centering
        (b) \texttt{Collider}
    \endminipage\hfill
    \minipage{0.32\textwidth}
        \centering
       (c) \texttt{Tree}
    \endminipage\hfill
    \vspace{6mm}\\
    \textbf{Graphs with cyclic skeletons:}\\
    \vspace{3mm}
    \minipage{0.32\textwidth}
        \centering
        \includegraphics[width=0.99\linewidth]{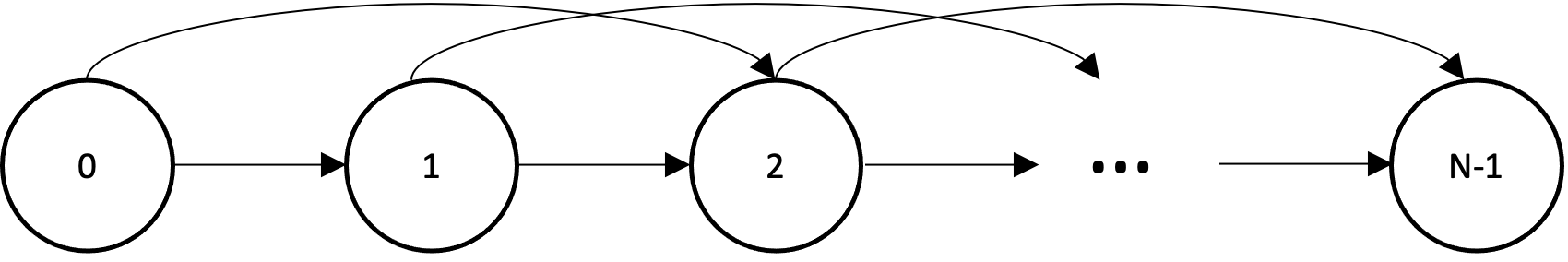}\\
    \endminipage\hfill
    \minipage{0.32\textwidth}
        \centering
        \includegraphics[width=0.99\linewidth]{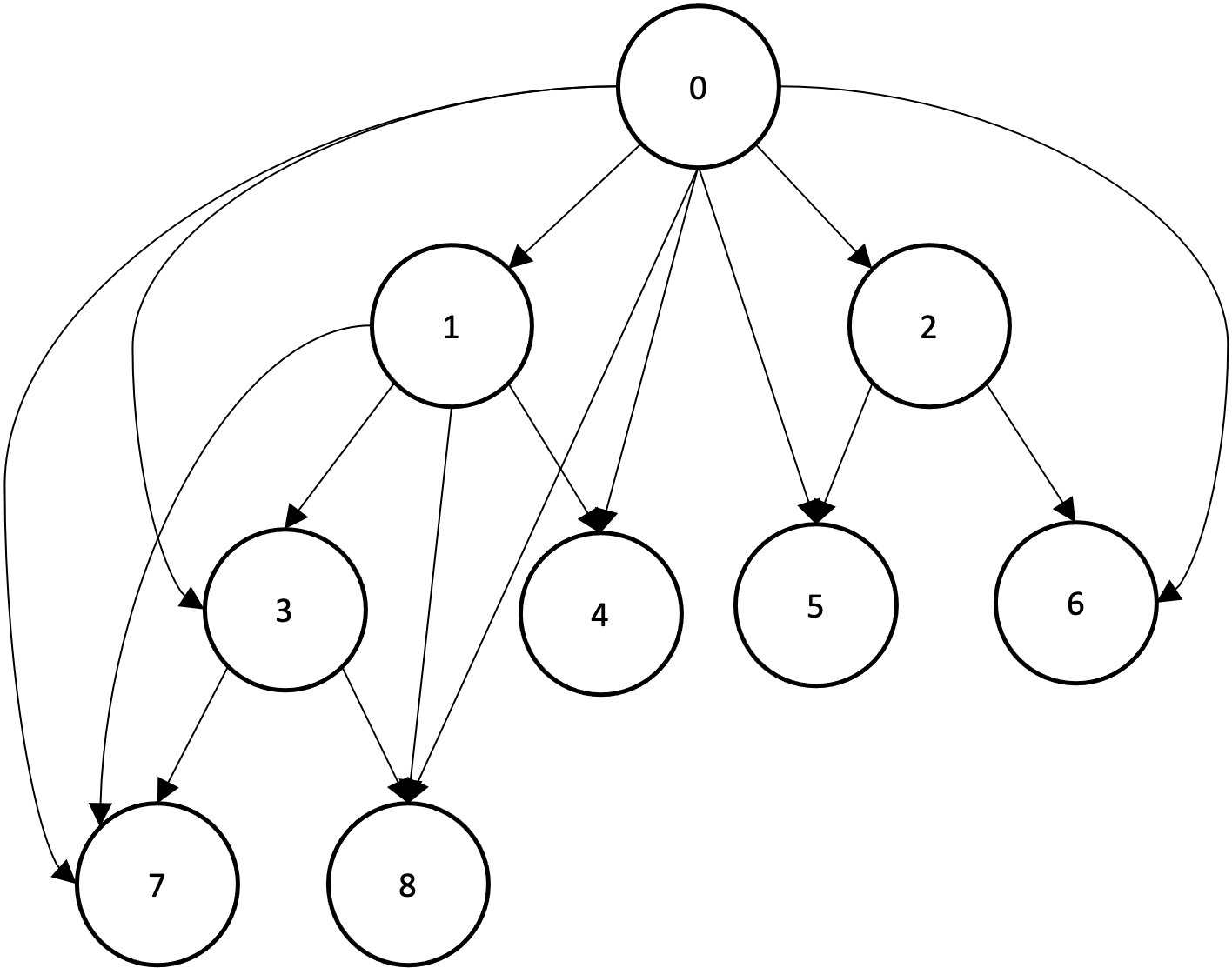}\\
    \endminipage\hfill
    \minipage{0.32\textwidth}
        \centering
       \includegraphics[width=0.99\linewidth]{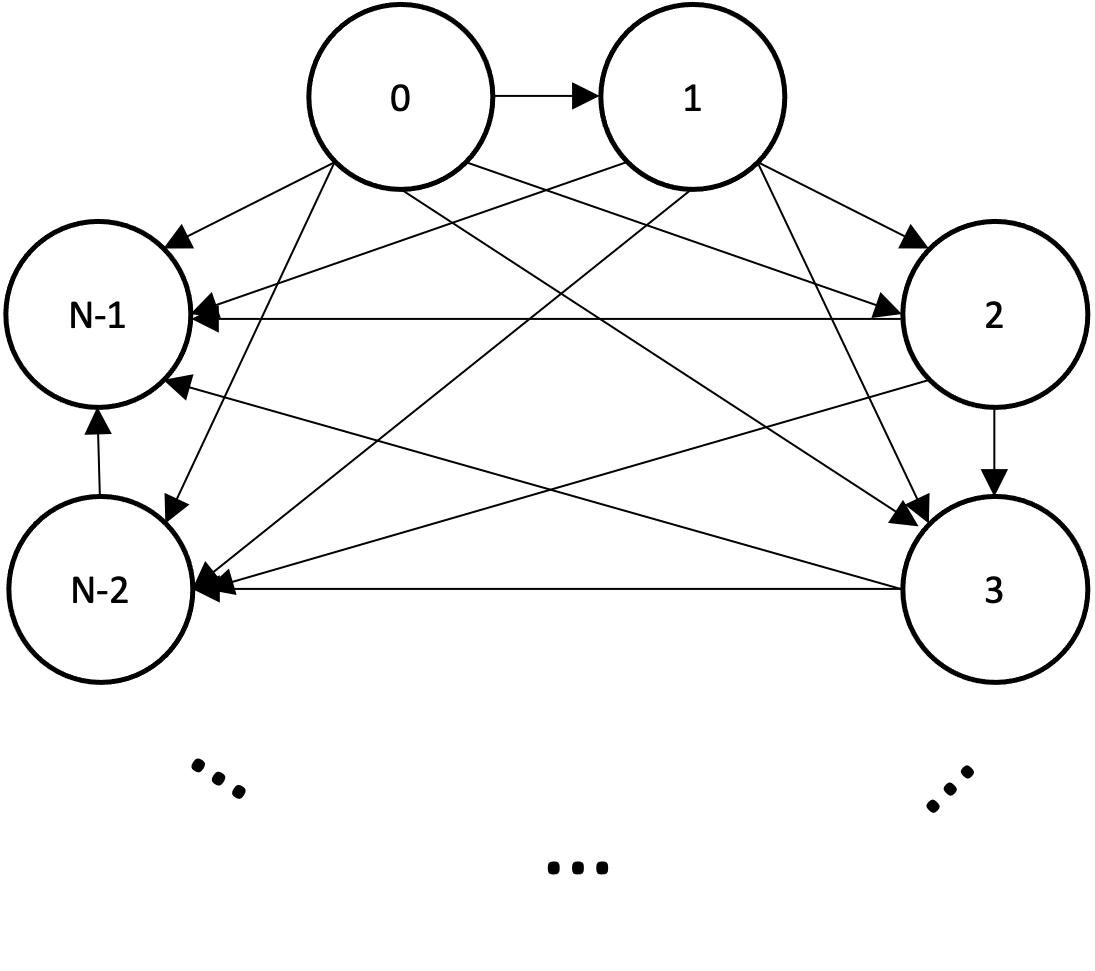}\\
    \endminipage\hfill
   \vspace{3mm}\\
    \minipage{0.32\textwidth}
        \centering
        (d) \texttt{Bidiag}
    \endminipage\hfill
    \minipage{0.32\textwidth}
        \centering
        (e) \texttt{Jungle}
    \endminipage\hfill
    \minipage{0.32\textwidth}
        \centering
        (f) \texttt{Full}
    \endminipage\hfill
    
    \caption{Visualization of Structured Graphs as proposed in \citet{ke2019learning} - adapted illustration}
    \label{fig:structuredGraphs_DSDI}
\end{figure}

\textbf{Graph Structure.} We adopt the structured graphs (see Fig. \ref{fig:structuredGraphs_DSDI}) proposed in the work of DSDI \citep{ke2019learning} as they adequately represent topological diversity of possible DAGs in a compact fashion. They can be split up in a set of graphs without cycles in the undirected skeletons, and one group with cycles. Extending the setup with random graphs with varying edge densities, generated from the Erdős–Rényi (ER) model, allows us to assess the generalized performance of the proposed method from sparse to dense DAGs.

\textbf{Discrete Data Generation.} We adopt the generative setup of DSDI \citep{ke2019learning} and model the SCMs using two-layer MLPs with Leaky ReLU activations between layers. For every variable $X_i$, a seperate MLP models the conditional relationship $P(X_i|X_{pa(i)})$. The MLP parameters are initialized orthogonally within the range of $[-2.5, 2.5]$ and biases uniformly in the range of $[-1.1, 1.1]$.

\textbf{Continuous Data Generation.} For the evaluation of the adapted DCDI framework, we adopt their generative setup as described in \citep{brouillard2020differentiable} and use the existing non-linear datasets.

\clearpage
\subsubsection{Real-World Datasets}
\begin{figure}[h!]
    \centering
    \minipage{0.49\textwidth}
        \centering
        \includegraphics[width=0.8\linewidth]{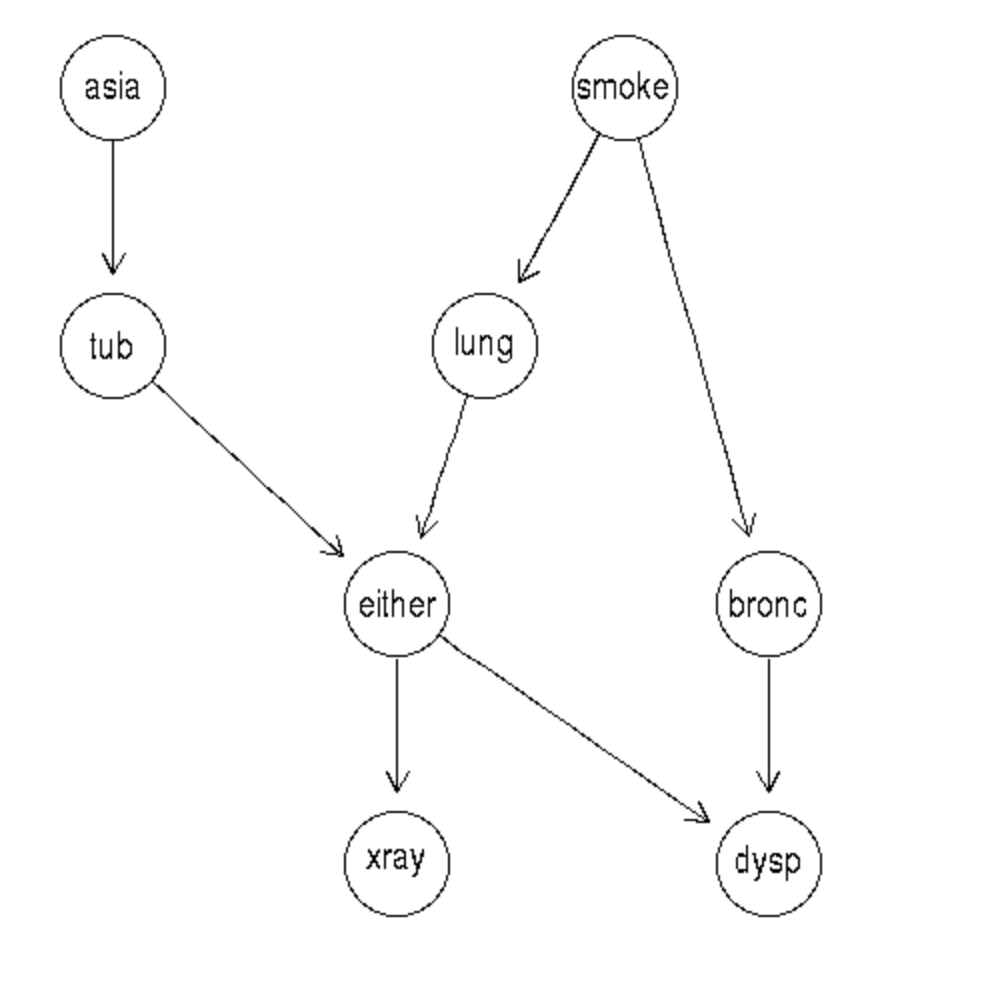}\\
        (a) Asia
    \endminipage\hfill
    \minipage{0.49\textwidth}
        \centering
        \includegraphics[width=0.8\linewidth]{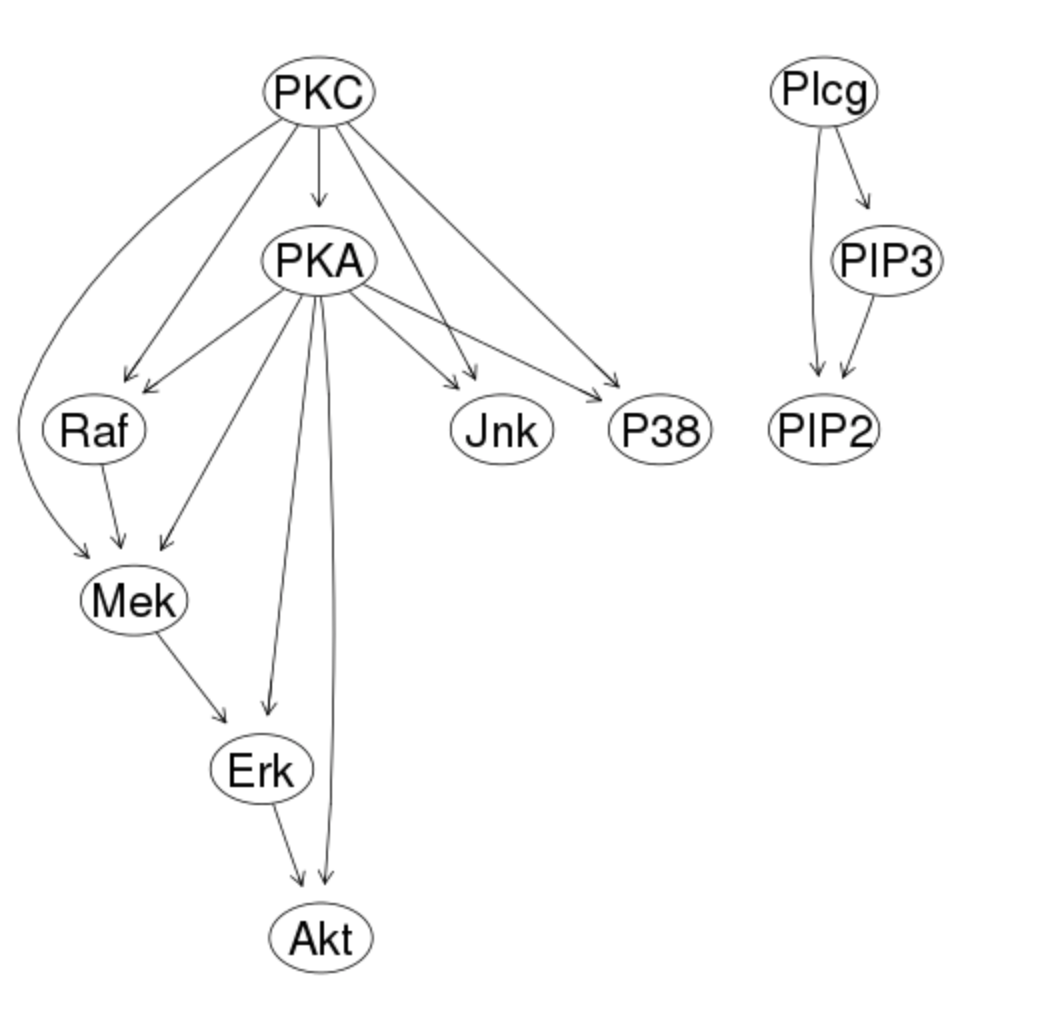}\\
        (b) Sachs
    \endminipage\hfill
    \caption{Ground-truth structure of the evaluated real-world datasets provided by the BnLearn data repository - Illustration from: \url{https://www.bnlearn.com/bnrepository/discrete-small.html}}
    \label{fig:bnlearnGraphs}
\end{figure}

Besides the many synthetic graphs, we evaluate our method on real-world datasets provided by the BnLearn data repository. Namely on the Asia \citep{lauritzen1988local} and the Sachs \citep{sachs2005causal} datasets (see Fig. \ref{fig:bnlearnGraphs} for a visualization of their underlying ground-truth structure).  Sachs \citep{sachs2005causal} represents a systems biology dataset which exhibits non-linearity, confounding and complex structure.

\vspace{5mm}
\subsection{Availability of Used (Existing) Assets}
\textbf{Base Frameworks.}
\begin{list}{\labelitemi}{\leftmargin=2em}
     \item DSDI \citep{ke2019learning}: \small\url{https://github.com/nke001/causal_learning_unknown_interventions}
    \item DCDI \citep{brouillard2020differentiable}: \url{https://github.com/slachapelle/dcdi}
\end{list}

\textbf{Baseline Methods.}
\begin{list}{\labelitemi}{\leftmargin=2em}
    \item GES  \citep{chickering2002optimal} and GIES \citep{hauser2012characterization}: \url{www.github.com/FenTechSolutions/CausalDiscoveryToolbox} \citep{kalainathan2019causal}
    \item ICP \citep{peters2016causal}: \url{https://github.com/juangamella/aicp}
    \item A-ICP \citep{gamella2020active}: \url{https://github.com/juangamella/aicp}  
    \item NOTEARS  \citep{zheng2018dags}: \url{https://github.com/xunzheng/notears}
    \item DAG-GNN \citep{yu2019dag}:  \url{https://github.com/fishmoon1234/DAG-GNN}  
\end{list}

\textbf{Datasets.}
\begin{list}{\labelitemi}{\leftmargin=2em}
    \item BnLearn Data Repository: \url{https://www.bnlearn.com/bnrepository/} 
\end{list}

\clearpage
\subsection{Hyper-Parameters}
\label{sec:app_hyperparameters}
We used a similar set of hyperparameters for our AIT + DSDI and AIT + DCDI models as those used in the original paper \citep{ke2019learning,brouillard2020differentiable}. The specific hyperparamters we used are stated as follows. 

\textbf{DSDI.} \\

\begin{table}[h!]
    \centering
    \caption{Hyperparameters for DSDI including the corresponding \\ AIT parameters}
    \begin{tabular}{lr}
    \toprule
        Number of iterations & 1000  \\
        Batch size & 256 \\
        Sparsity Regularizer & 0.1 \\
        DAG Regularizer & 0.5 \\
        Functional parameter training iterations & 10000 \\
        Number of interventions per phase 2 & 25 \\
        Number of data batches for scoring & 10 \\
        Number of graph configurations for scoring & \\
                    - Graph Size 5:   & 10   \\
                    - Graph Size 10:  & 20   \\
                    - Graph Size 15   & 40   \\ 
        \midrule
        AIT: \\
        - Number of graph configurations & 100 \\
        - Number of interventional samples per graph \& target & 256 \\
    \bottomrule
    \end{tabular}
    \label{tab:hyperparam_dsdi}
\end{table}

\textbf{DCDI.} 
\begin{table}[h!]
    \centering
    \caption{Hyperparameters for DCDI including the corresponding \\ AIT parameters}
    \begin{tabular}{lr}
    \toprule
        $\mu_0$ & $10^{-8}$  \\
        $\gamma_0$ & 0 \\
        $\eta$ & 2 \\
        $\delta$ & 0.9 \\
        Augmented Lagrangian Thresh & $10^{-8}$ \\
        Learning rate & $10^{-3}$ \\
        Nr. of hidden units & 16 \\
        Nr. of hidden layers & 2 \\
        \midrule
        AIT: \\
        - Number of graph configurations & 100 \\
        - Number of interventional samples per graph \& target & 256  \\
    \bottomrule
    \end{tabular}
    \label{tab:hyperparam_dcdi}
\end{table}
\clearpage

\subsection{Discrete Setting: Additional Experiments / Results}
In this section, we show further results and visualizations of experiments on discrete data and single-target interventions in various settings (such as graphs of varying size, noise-free vs. noise-perturbed, limited intervention targets). All experiments are based on the framework DSDI.

\subsubsection{Evaluation (SHD) on graphs of varying size and density}

\begin{table}[ht]
\vspace{-0.5\baselineskip}
    \centering
    \caption{SHD (lower is better) on various 5-variable synthetic datasets. Structured graphs are sorted in ascending order according to their edge density. \;\;$^{(*)}$ denotes average SHD over 10 random graphs., \; \textdagger ER-2 graphs on 5 results in the full5 graph and ER-4 graphs on 5 node graphs are non-existing 
    }
    \label{app_tab:shd_synthetic_5}
    \resizebox{0.88\textwidth}{!}{
    \begin{tabular}{lrrrrrrrrr}
        \toprule
        & \multicolumn{6}{c}{\bfseries Structured Graphs} &
        \multicolumn{3}{c}{\bfseries Random Graphs} \\
        \cmidrule(l){2-7} \cmidrule(l){8-10}  
        & Chain & Collider & Tree & Bidiag & Jungle & Full  & ER-1$^{(*)}$  & ER-2$^{(*)}$ & ER-4$^{(*)}$  \\
        \midrule
        GES \citep{chickering2002optimal}        & 3 & 0 & 4 & 6 & 4 &  9  & 4.3 ($\pm 1.0 $) & \textdagger &\textdagger \\
        NOTEARS \citep{zheng2018dags}            & 5 & 3 & 6 & 5 & 7 & 9    &  6.1 ($\pm 1.7 $) & \textdagger &\textdagger \\
        DAG-GNN \citep{yu2019dag}                & 4 & 4 & 3 & 4 & 6 & 9    &  5.1 ($\pm 1.4 $) & \textdagger &\textdagger \\
        \midrule 
        GIES \citep{hauser2012characterization}  & 3 & 4 & 2 & 6 & 5 & 10  & 4.7 ($\pm 1.6 $) & \textdagger &\textdagger\\
        ICP \citep{peters2016causal}             & 4 & 4 & 4 & 7 & 6 & 10  & 5.4 ($\pm 1.4 $) & \textdagger &\textdagger \\
        \midrule
        A-ICP \citep{gamella2020active}          & 4 & 4 & 4 & 7 & 6 & 10  & 5.4 ($\pm 1.4 $) & \textdagger &\textdagger \\
        \midrule
        DSDI (Random) \citep{ke2019learning}  & \highlight{0}  & \highlight{0} & \highlight{0}     & \highlight{0} & \highlight{0} & \highlight{0}    & \highlight{0.0 ($\pm 0.0$)} & \textdagger &  \textdagger   \\
        DSDI (AIT)                              & \highlight{0}  & \highlight{0} & \highlight{0}     & \highlight{0} & \highlight{0} & \highlight{0}    & \highlight{0.0 ($\pm 0.0$)} & \textdagger &  \textdagger   \\
        \bottomrule
    \end{tabular}
    }
\end{table}

\begin{table}[ht]
\vspace{-0.5\baselineskip}
    \centering
    \caption{SHD (lower is better) on various 10-variable synthetic datasets. Structured graphs are sorted in ascending order according to their edge density. \;\;$^{(*)}$ denotes average SHD over 10 random graphs. 
    }
    \label{app_tab:shd_synthetic_10}
    \resizebox{0.88\textwidth}{!}{
    \begin{tabular}{lrrrrrrrrr}
        \toprule
        & \multicolumn{6}{c}{\bfseries Structured Graphs} &
        \multicolumn{3}{c}{\bfseries Random Graphs} \\
        \cmidrule(l){2-7} \cmidrule(l){8-10}  
        & Chain & Collider & Tree & Bidiag & Jungle & Full  & ER-1$^{(*)}$ & ER-2$^{(*)}$ & ER-4$^{(*)}$    \\
        \midrule
        GES \citep{chickering2002optimal}        &  9  & 2 &  6 &  8 & 10 &  35   &   7.0 ($\pm 1.6 $) &  10.7 ($\pm 3.8 $) & 26.7 ($\pm 2.9$) \\
        NOTEARS \citep{zheng2018dags}            & 13 & 16 & 12 & 21 & 21 & 42    &  16.4 ($\pm 3.4$)  & 22.9 ($\pm 2.9 $) & 36.6 ($\pm 2.6 $) \\
        DAG-GNN \citep{yu2019dag}                & 8 & 7 & 6 & 15 & 13 & 38    &  10.3 ($\pm 2.8 $)  & 20.1 ($\pm 3.5 $) & 38.4 ($\pm 1.9$) \\
        \midrule 
        GIES \citep{hauser2012characterization}  & 12  & 6 & 13 & 16 &  9 &  20   &  12.2 ($\pm 5.1 $) &  14.1 ($\pm 4.7 $) & 26.1 ($\pm 4.4$)  \\
        ICP \citep{peters2016causal}             & 9 & 9 & 9 & 17 & 16 & 45    &  10.6 ($\pm 2.5$) & 20.7 ($\pm 3.3$) & 39.8 ($\pm 1.9$) \\
        \midrule
        A-ICP \citep{gamella2020active}          & 9 & 9 & 9 & 17 & 16 & 45    &  10.6 ($\pm 2.5$) & 20.7 ($\pm 3.3$) & 39.8 ($\pm 1.9$) \\
        \midrule
        DSDI (Random) \citep{ke2019learning}  & \highlight{0}  & \highlight{0} & \highlight{0}     & \highlight{0} & \highlight{0} & \highlight{0}    & \highlight{0.0 ($\pm 0.0$)}   & \highlight{0.0 ($\pm 0.0$)}  & \highlight{0.0 ($\pm 0.0$)}   \\
        DSDI (AIT)                              & \highlight{0}  & \highlight{0} & \highlight{0}     & \highlight{0} & \highlight{0} & \highlight{0}    & \highlight{0.0 ($\pm 0.0$)}   & \highlight{0.0 ($\pm 0.0$)}  & \highlight{0.0 ($\pm 0.0$)}   \\
        \bottomrule
    \end{tabular}
    }
\end{table}

\begin{table}[ht]
\vspace{-0.5\baselineskip}
    \centering
    \caption{SHD (lower is better) on various 15-variable synthetic datasets. Structured graphs are sorted in ascending order according to their edge density. \;\;$^{(*)}$ denotes average SHD over 10 random graphs. 
    }
    \label{app_tab:shd_synthetic_15}
    \resizebox{0.88\textwidth}{!}{
    \begin{tabular}{lrrrrrrrrr}
        \toprule
        & \multicolumn{6}{c}{\bfseries Structured Graphs} &
        \multicolumn{3}{c}{\bfseries Random Graphs} \\
        \cmidrule(l){2-7} \cmidrule(l){8-10}  
        & Chain & Collider & Tree & Bidiag & Jungle & Full  & ER-1$^{(*)}$ & ER-2$^{(*)}$ & ER-4$^{(*)}$    \\
        \midrule
        GES \citep{chickering2002optimal}        & 13 &  1 & 12  & 14  & 14 & 69     & 8.3 ($\pm 1.9$) & 17.6 ($\pm 4.6$) & 39.4 ($\pm 6.7$)  \\
        NOTEARS \citep{zheng2018dags}            & 22 & 21 & 26 & 33  & 35 & 93     & 23.7 ($\pm 4.0$)  & 35.8 ($\pm 5.2$) & 59.5 ($\pm 3.7$)  \\
        DAG-GNN \citep{yu2019dag}                & 11 & 14 & 15  & 27  & 25 & 97    & 16.0 ($\pm 3.7$) & 30.6 ($\pm 3.4$) & 59.7 ($\pm 4.1$)  \\
        \midrule 
        GIES \citep{hauser2012characterization}  & 13 &  6 & 10  & 17  & 23 & 60    & 10.9 ($\pm 4.2$) & 18.1 ($\pm 4.3$) & 39.3 ($\pm 5.6$)  \\
        ICP \citep{peters2016causal}             & 14 & 14 & 14  & 27  & 26 & 105   & 16.2 ($\pm 3.6$) & 31.1 ($\pm 3.4$) & 60.1 ($\pm 3.9$)  \\
        \midrule
        A-ICP \citep{gamella2020active}          & 14 & 14 & 14  & 27  & 26 & 105   & 16.2 ($\pm 3.6$) & 31.1 ($\pm 3.4$) & 60.1 ($\pm 3.9$)  \\
        \midrule
        DSDI (Random) \citep{ke2019learning}  & \highlight{0} & \highlight{0} & 2  & 3 & 7 & 24    & 1.4 ($\pm 1.6$)   & 2.1 ($\pm 2.3$)  & 7.2 ($\pm 2.7$)   \\
        DSDI (AIT)                              & \highlight{0}  & \highlight{0} & \highlight{0}     & \highlight{0} & \highlight{0} & \highlight{7}    & \highlight{0.0 ($\pm 0.0$)}   & \highlight{0.0 ($\pm 0.0$)}  & \highlight{0.0 ($\pm 0.0$)}   \\
        \bottomrule
    \end{tabular}
    }
\end{table}

\clearpage
\subsubsection{Evaluation of convergence speed on graphs of varying size and density}
\begin{figure}[!h]
\minipage{0.30\textwidth}
    \centering
    \includegraphics[width=\linewidth]{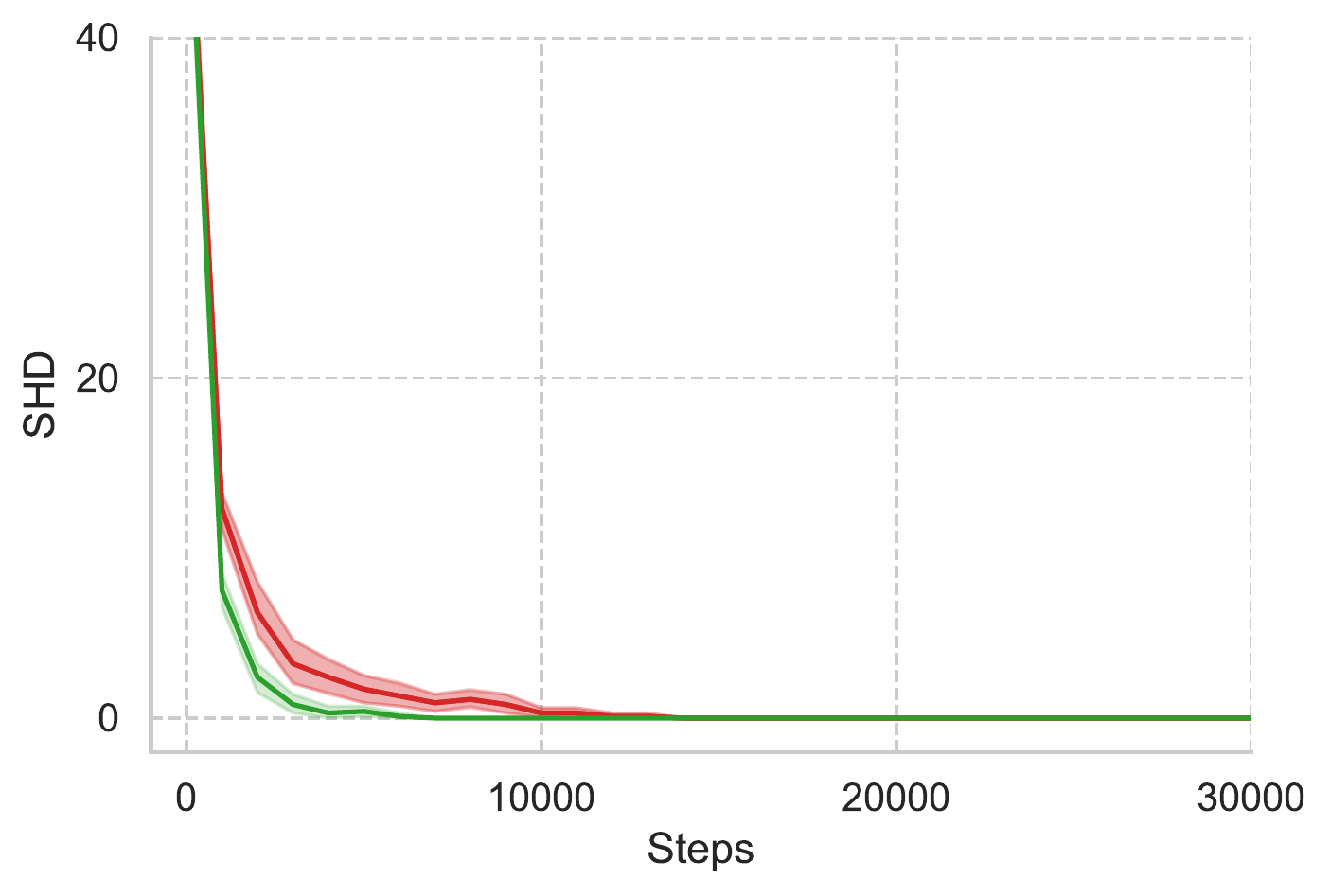}
    (a) \texttt{ER-1:}
\endminipage\hfill
\minipage{0.30\textwidth}
    \centering
    \includegraphics[width=\linewidth]{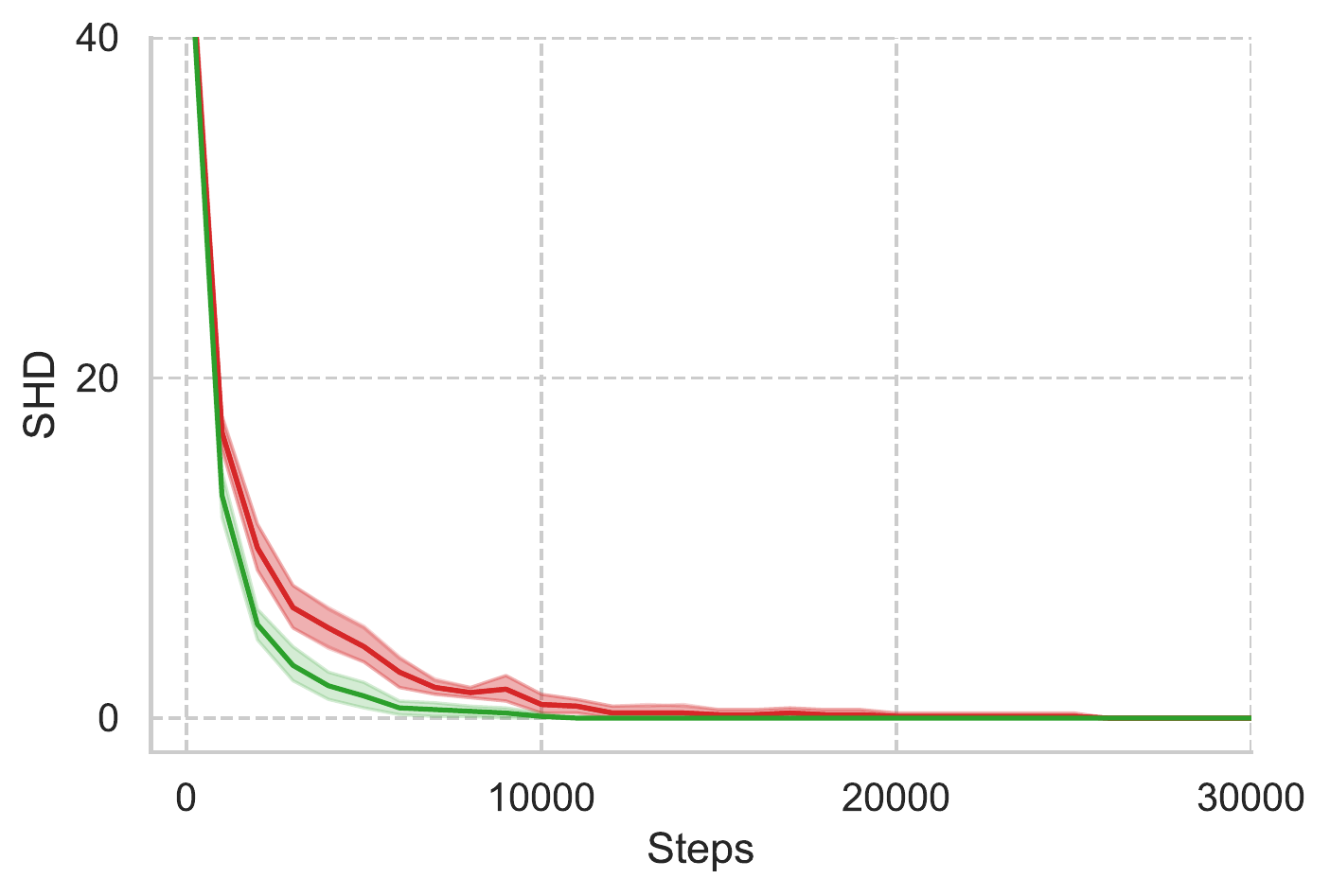}
    (b) \texttt{ER-2:}
\endminipage\hfill
\minipage{0.30\textwidth}%
    \centering
    \includegraphics[width=\linewidth]{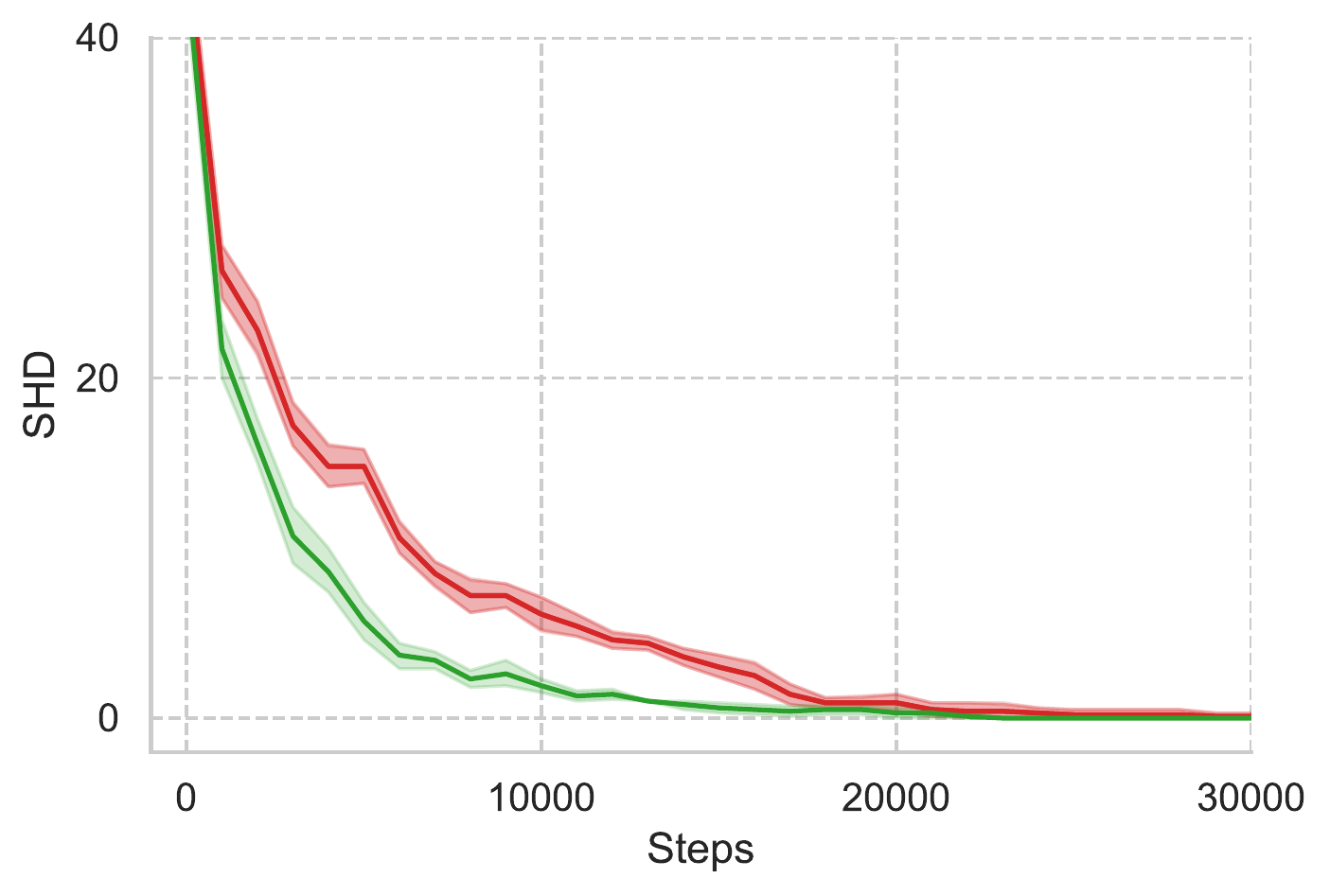}
    (c) \texttt{ER-4:}
\endminipage
\caption{DSDI with AIT (green) leads to superior performance over random intervention targeting (red) on \textbf{random graphs of size 10} of varying edge densities.  Error bands were estimated using 10 random ER graphs per setting.}
\label{fig:performance_ER-Graphs_N=10}
\end{figure}

While we have shown the effectiveness of AIT on random ER graphs of size $15$ in \S\ref{sec:results_effectOfAIT}, we observe similar effects on ER graphs of size $10$ (see Figure \ref{fig:performance_ER-Graphs_N=10}). Overall, the results indicate a greater impact of our proposed targeting mechanisms on graphs of bigger size compared to random intervention targeting which poorly scales to graphs of larger size.\\

\begin{figure}[!h]
\minipage{0.30\textwidth}
    \centering
    \includegraphics[width=\linewidth]{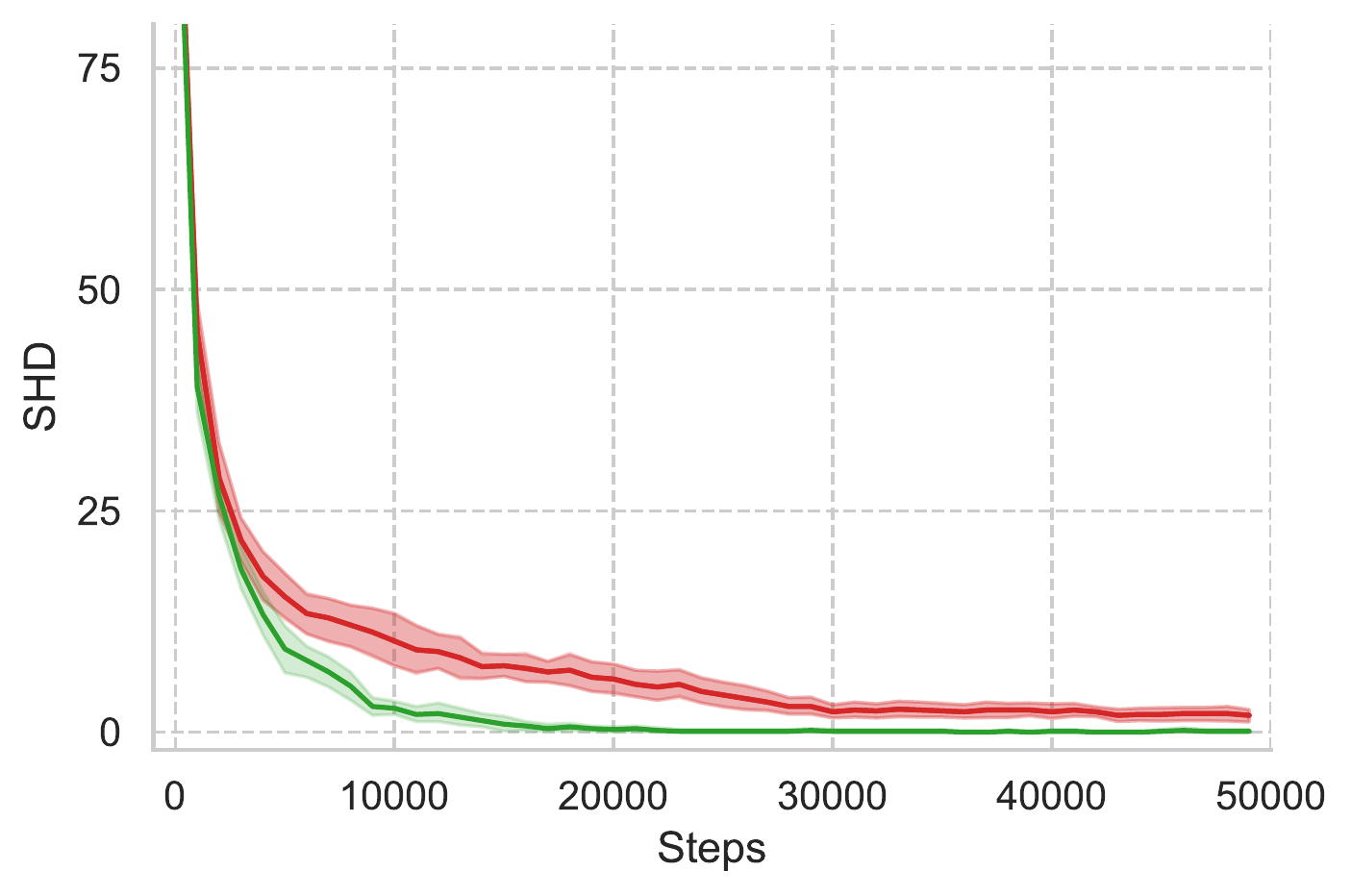}
    (a) \texttt{ER-1:}
\endminipage\hfill
\minipage{0.30\textwidth}
    \centering
    \includegraphics[width=\linewidth]{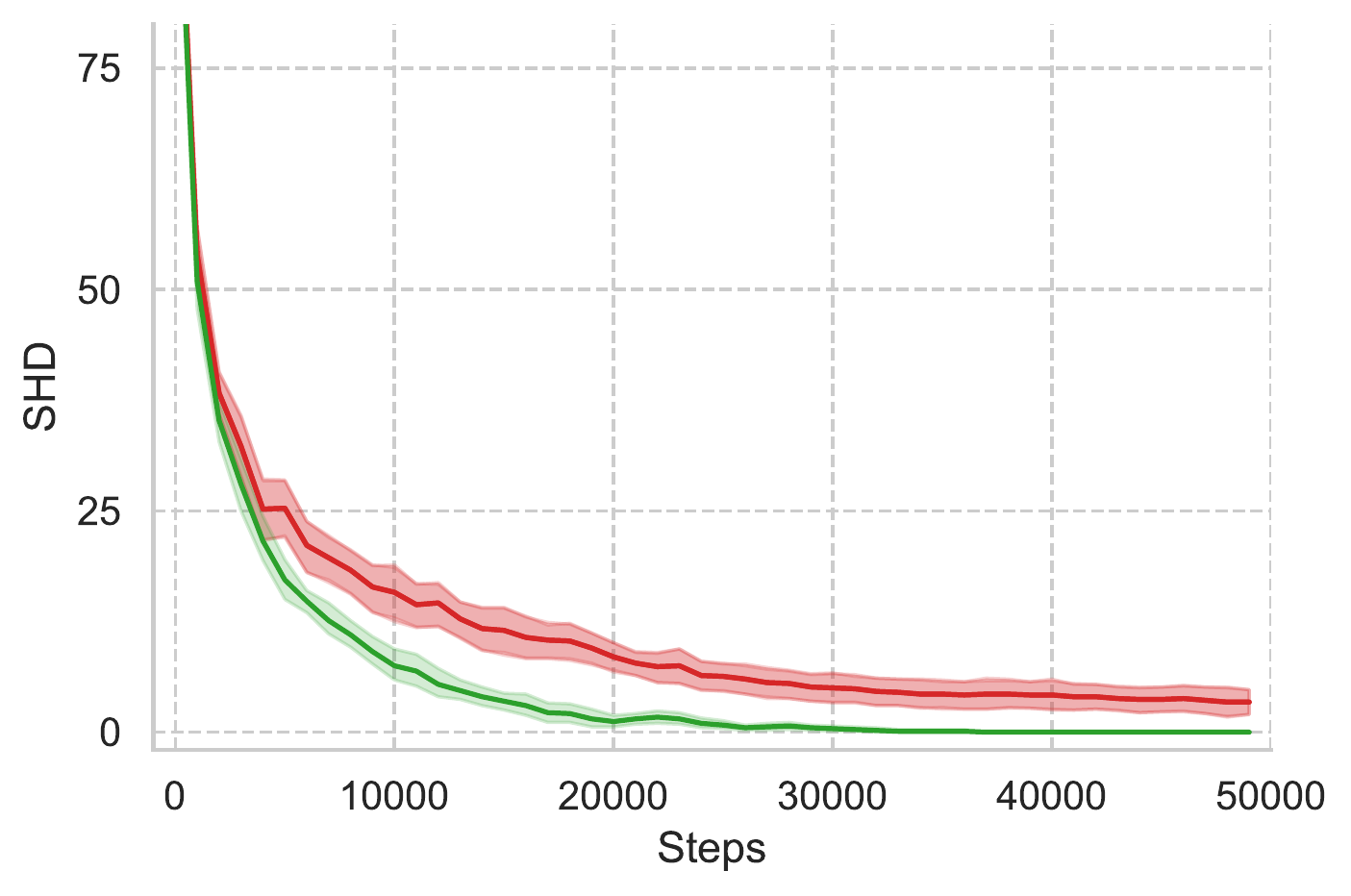}
    (b) \texttt{ER-2:}
\endminipage\hfill
\minipage{0.30\textwidth}%
    \centering
    \includegraphics[width=\linewidth]{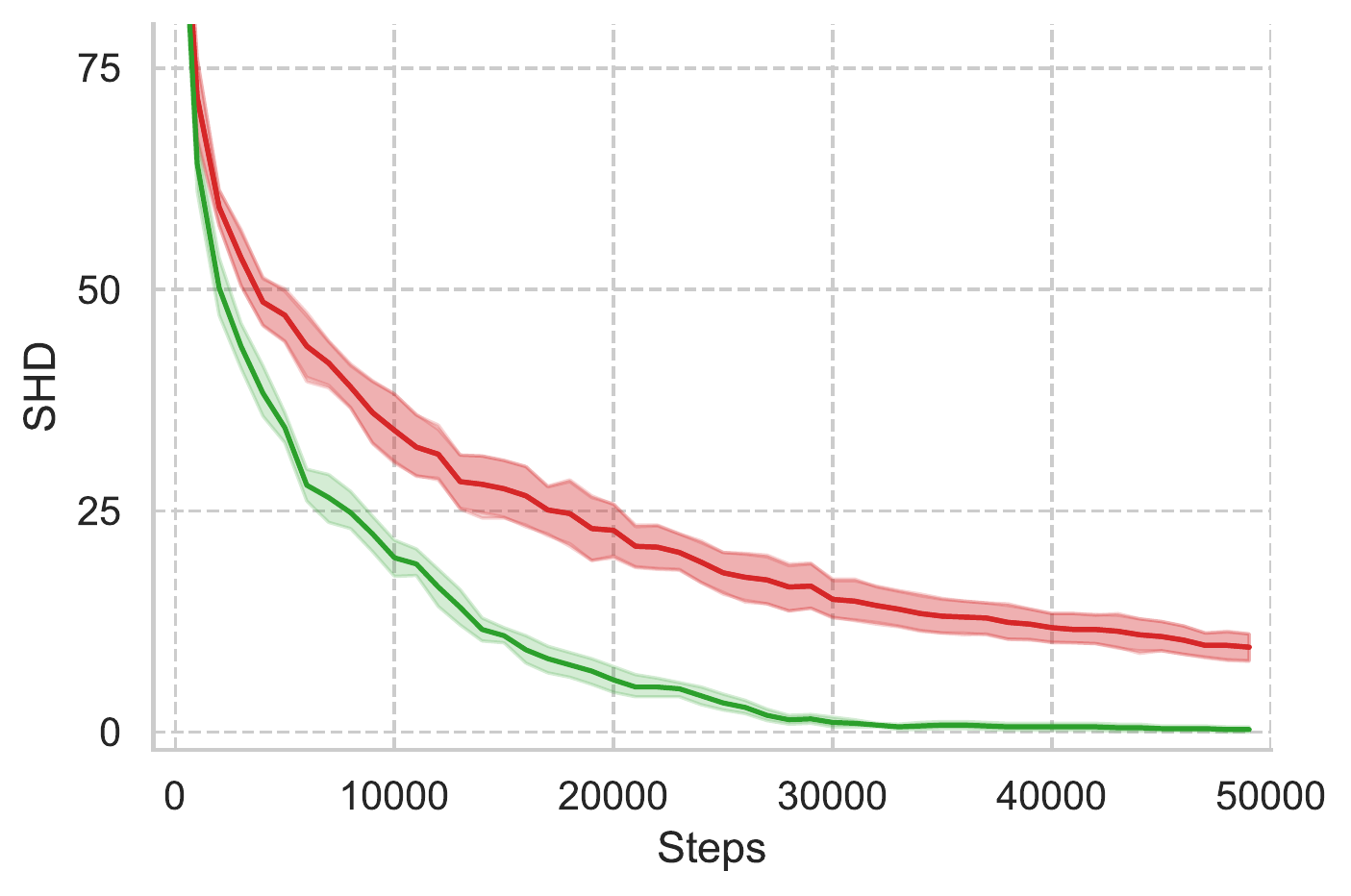}
    (c) \texttt{ER-4:}
\endminipage
\caption{DSDI with AIT (green) leads to superior performance over random intervention targeting (red) on \textbf{random graphs of size 15} of varying edge densities. Error bands were estimated using 10 random ER graphs per setting.}
\label{fig:performance_ER-Graphs_N=15}
\end{figure}

\clearpage
\subsubsection{Target selection analysis for graphs of varying size and density}
We evaluate the distribution of target node selections over multiple DAGs of varying size to investigate the behaviour of our proposed method. Over all performed experiments, our method prefers interventions on nodes with greater (downstream) impact on the overall system, i.e. nodes of higher topological rank in the underlying DAG. \vspace{5mm} \\

\begin{figure}[!h]
    \centering
    \minipage{0.49\textwidth}
        \centering
        \includegraphics[width=\linewidth]{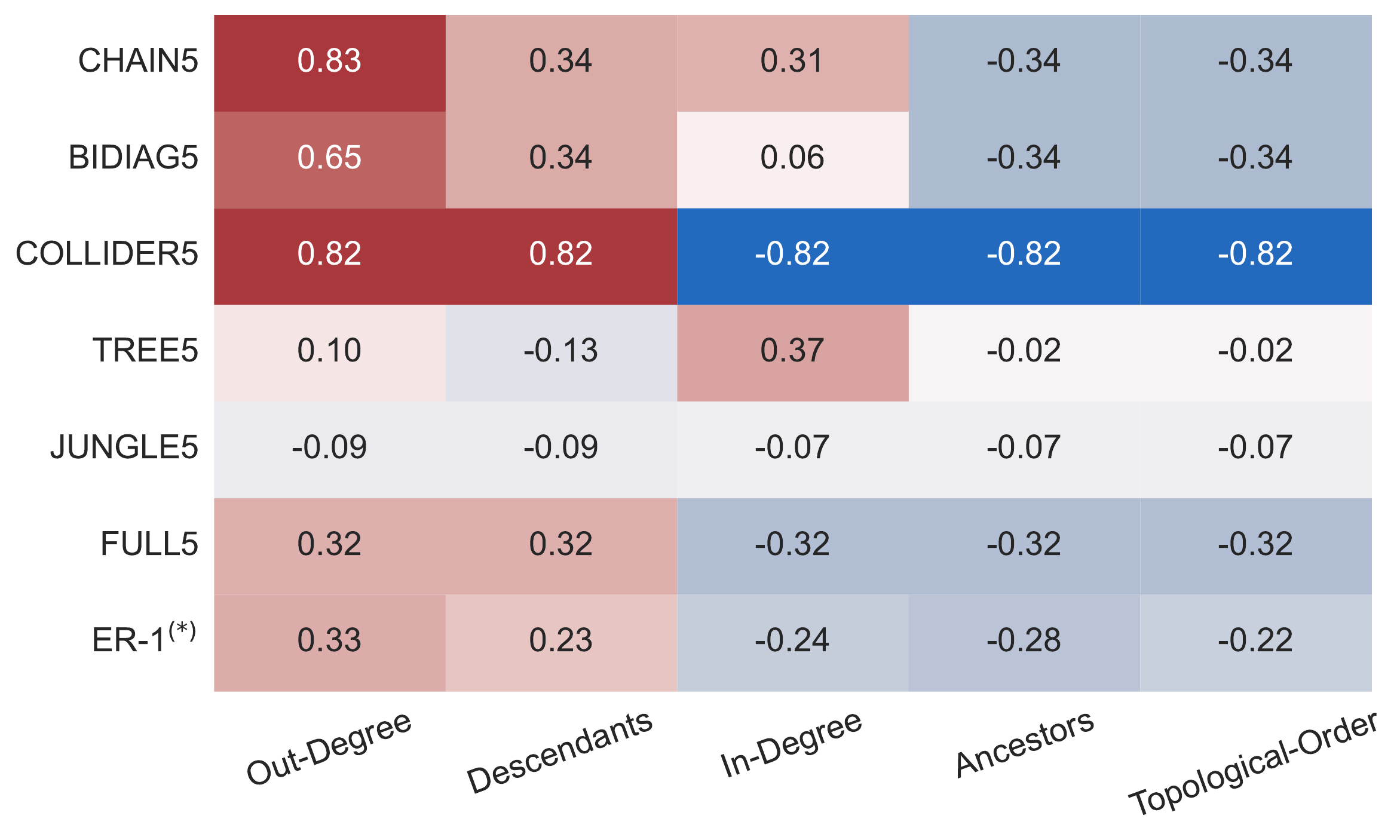}
        \includegraphics[width=\linewidth]{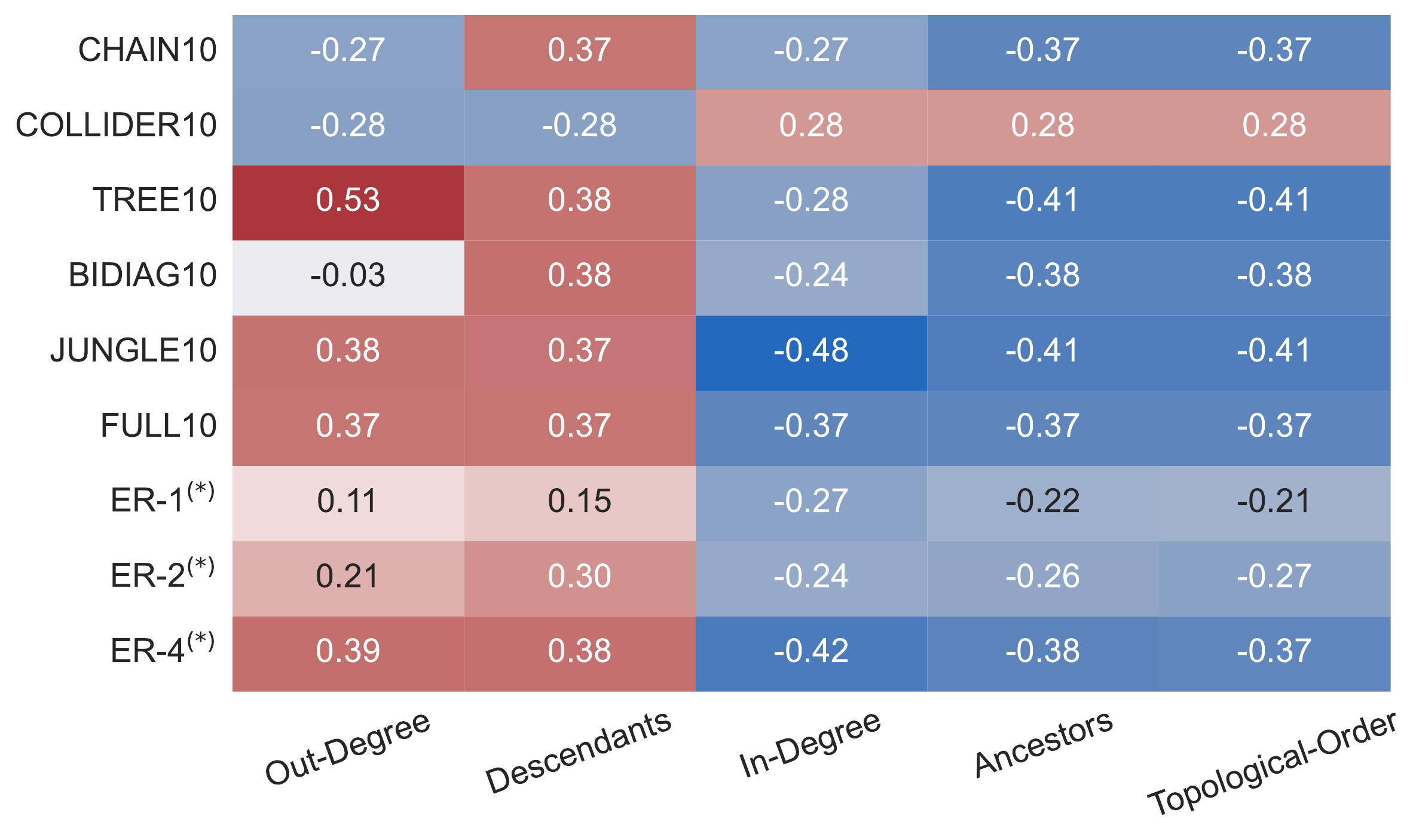}
        \includegraphics[width=\linewidth]{figures/correlationAnalysis/CorrelationAnalysis_Graphs15_random.pdf}
        (a) Random Targeting
    \endminipage\hfill
    \minipage{0.49\textwidth}
        \centering
        \includegraphics[width=\linewidth]{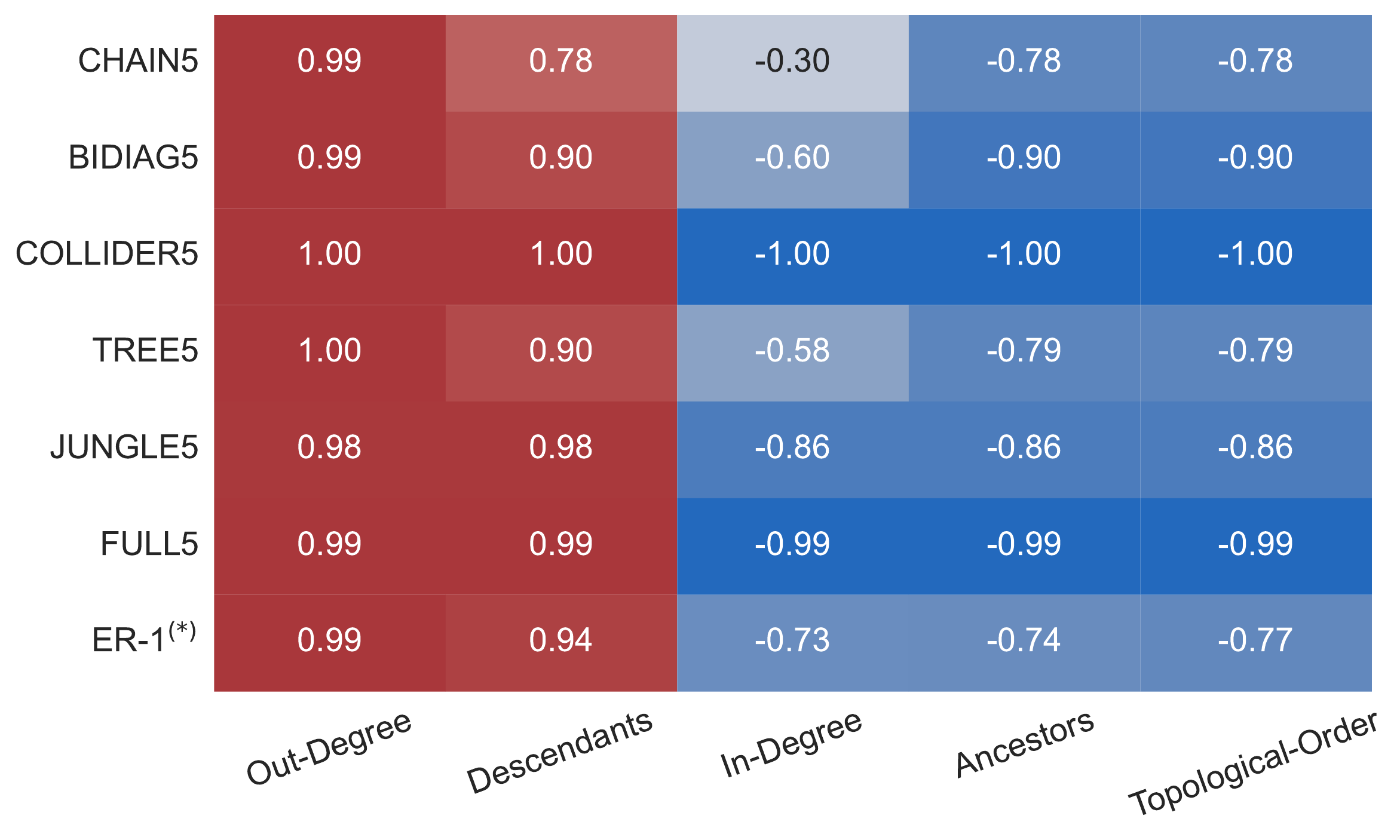}
        \includegraphics[width=\linewidth]{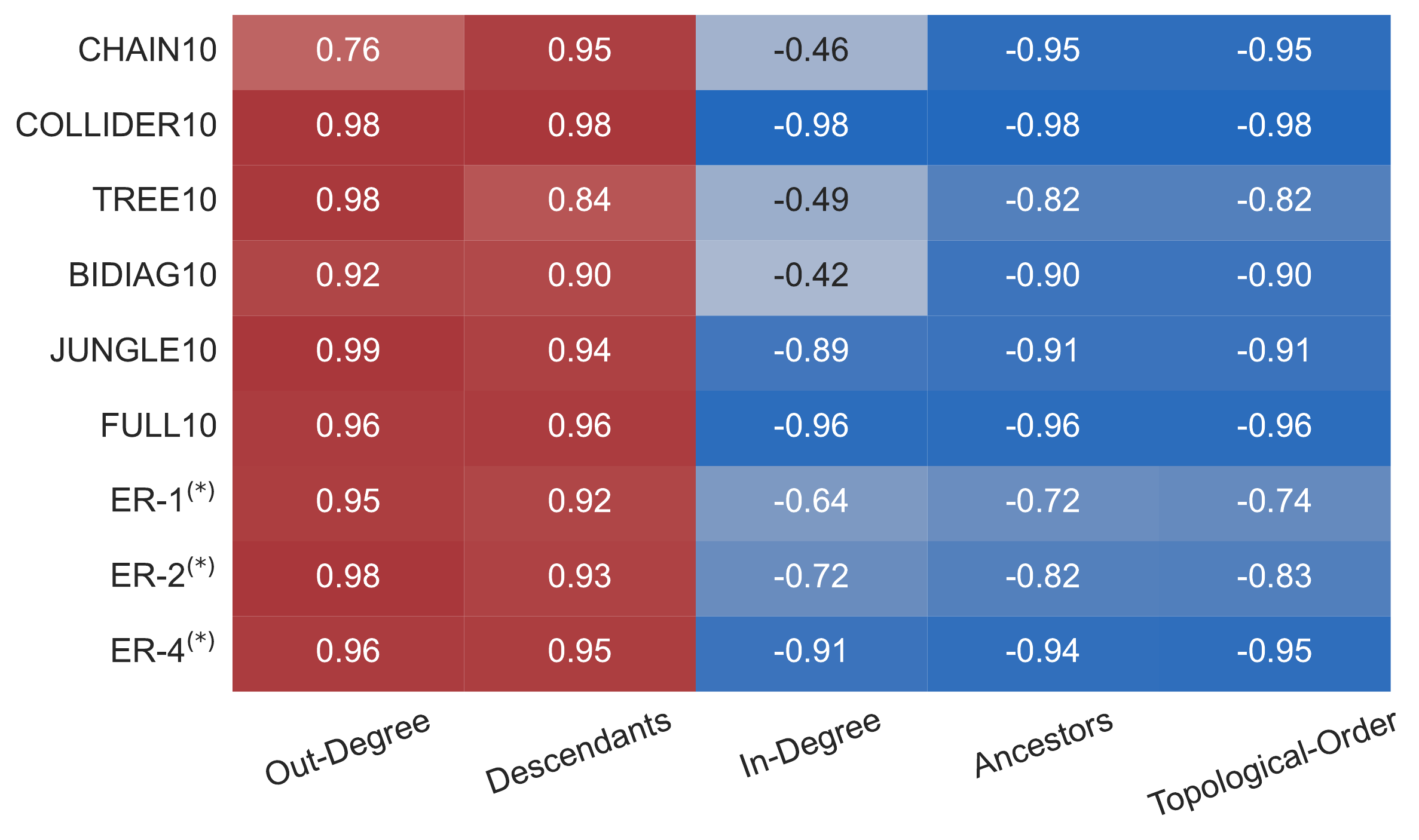}
        \includegraphics[width=\linewidth]{figures/correlationAnalysis/CorrelationAnalysis_Graphs15_AIT.pdf}
        (a) Active Intervention Targeting
    \endminipage\hfill
    \caption{Correlation scores over graphs of varying size and density between the number of individual target selections and different topological properties of those targets. AIT shows strong correlations with the measured properties over all graphs, which indicates a controlled discovery of the underlying structure through preferential targeting of nodes with greater (downstream) impact on the overall system.}
    \label{fig:correlationAnalysis_extended}
\end{figure}

\clearpage
\subsubsection{Visualization of target distribution on structured graphs of size 5}

\begin{figure}[!h]
    \centering
    \includegraphics[width=0.7\linewidth]{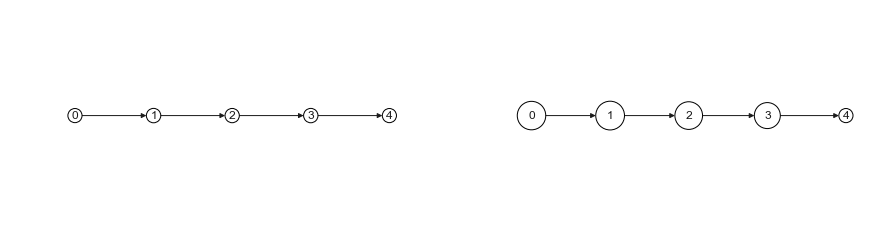}
    \includegraphics[width=0.7\linewidth]{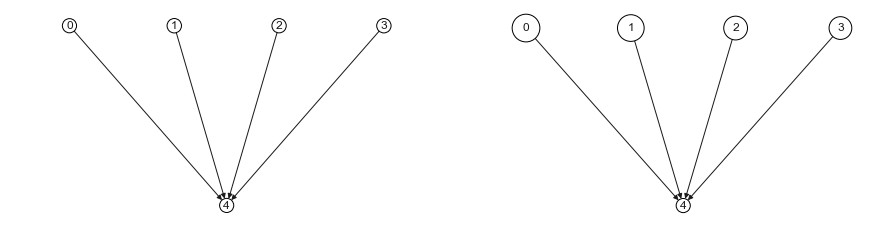}
    \includegraphics[width=0.7\linewidth]{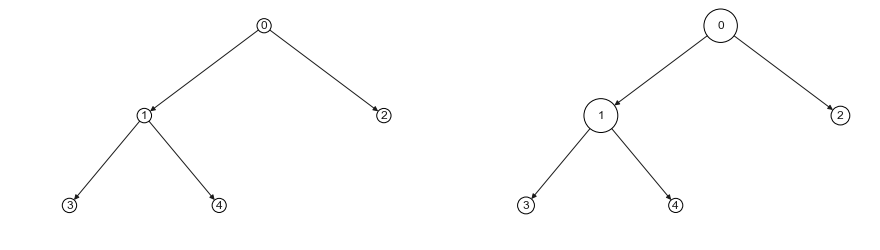}
    \includegraphics[width=0.7\linewidth]{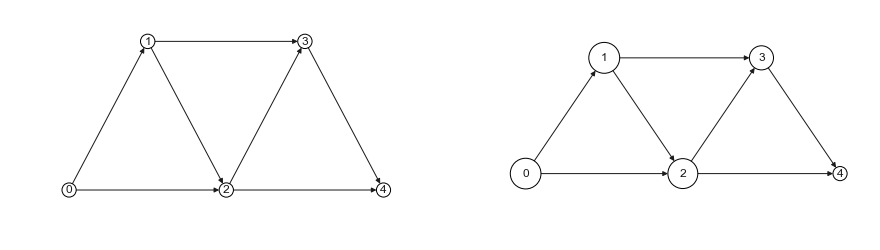}
    \includegraphics[width=0.\linewidth]{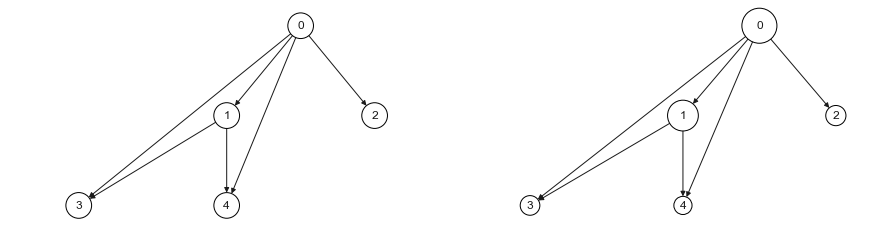}
    \includegraphics[width=0.7\linewidth]{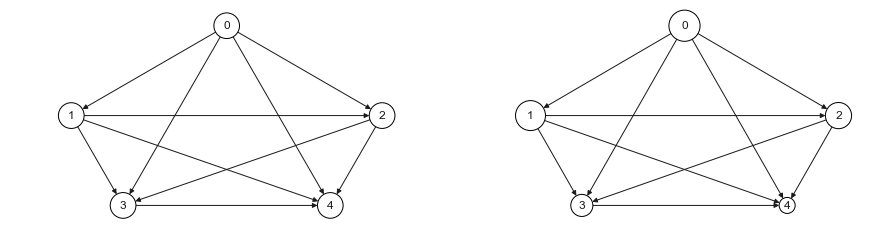}
    \\
    \hspace{8mm} (a) Random Targeting \hspace{18mm} (b) Active Intervention Targeting
    \caption{Visualization of target selection on structured graphs of size 5 - bigger node size denotes more selection of the node. While Random Targeting acts as we expect and selects every node an uniform amount, AIT prefers targeting of nodes with greater (downstream) impact on the overall system, i.e. nodes of higher topological order.}
    \label{fig:targetDistribution_visualization}
\end{figure}

\clearpage
\subsubsection{Extended Analysis of Edge Dynamics}

We show all edge dynamics of all structured graphs over 15 variables and compare the dynamics of random targeting to active intervention targeting in a noise-free setting where we have access to all possible single-target interventions.

\begin{figure}[h!]
    \vspace{-1\baselineskip}
    \centering
    \includegraphics[width=0.88\textwidth]{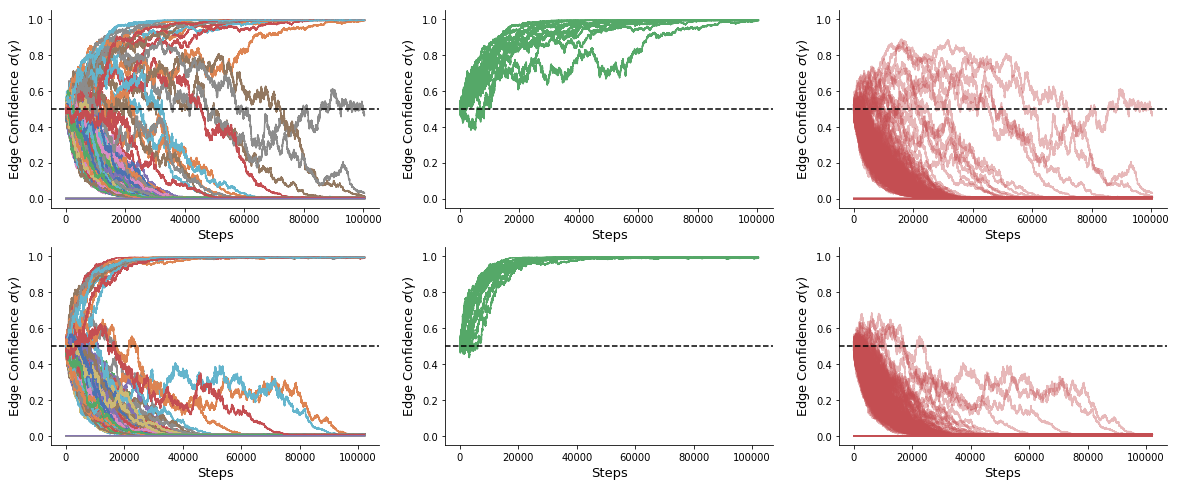}\\
    (a) Graph: Chain
    \includegraphics[width=0.88\textwidth]{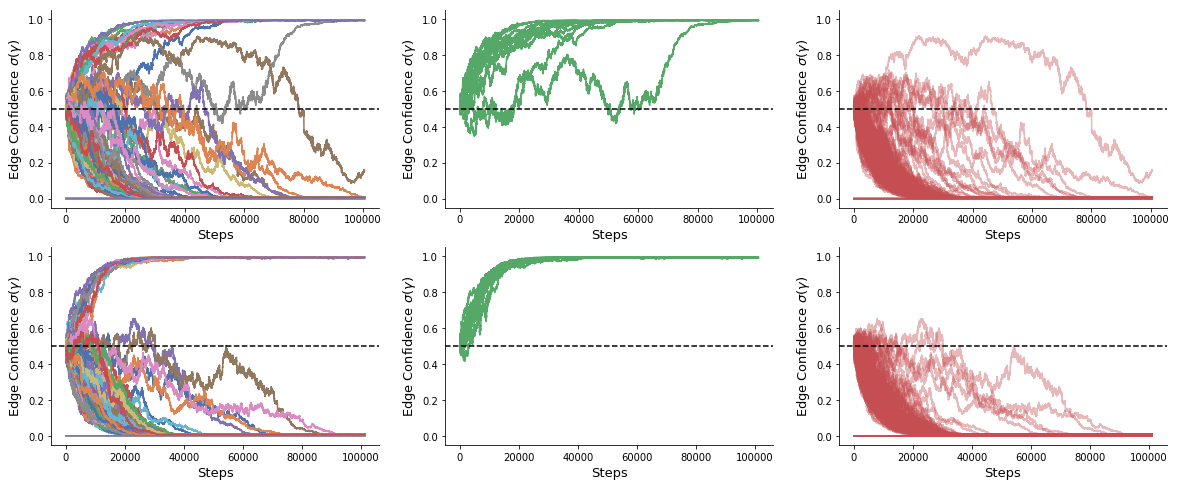}\\
    (b) Graph: Tree
    \includegraphics[width=0.88\textwidth]{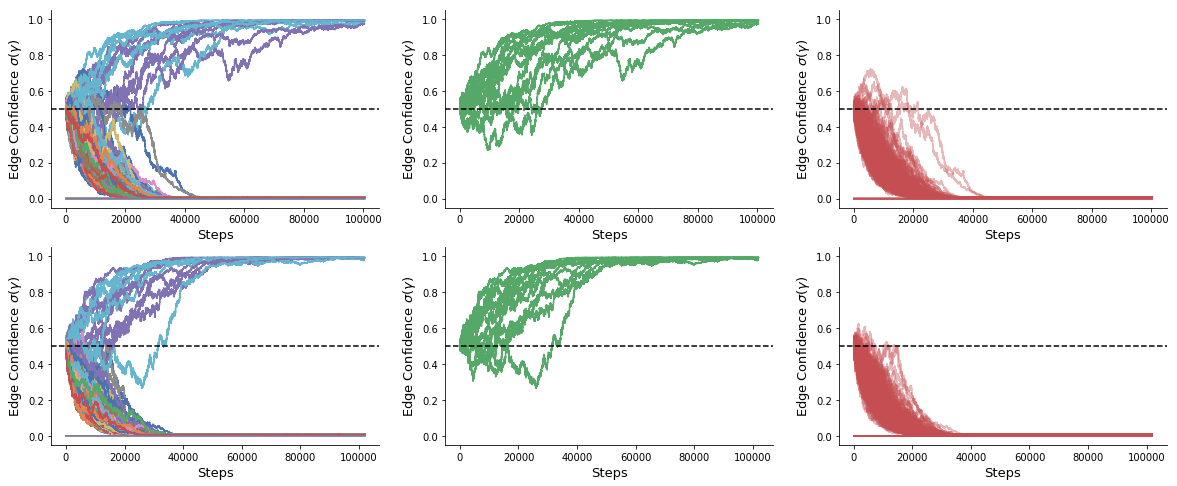}\\
    (c) Graph: Collider
    \caption{Edge Dynamics of the examined structured graphs spanning over 15 variables - Part 1: The upper part shows the dynamics of random targeting and the lower of active intervention targeting. }
    \label{fig:extAnalysis_edgeDynamics}
    \vspace{-1\baselineskip}
\end{figure}

\begin{figure}[h!]
    \centering
    \includegraphics[width=0.9\textwidth]{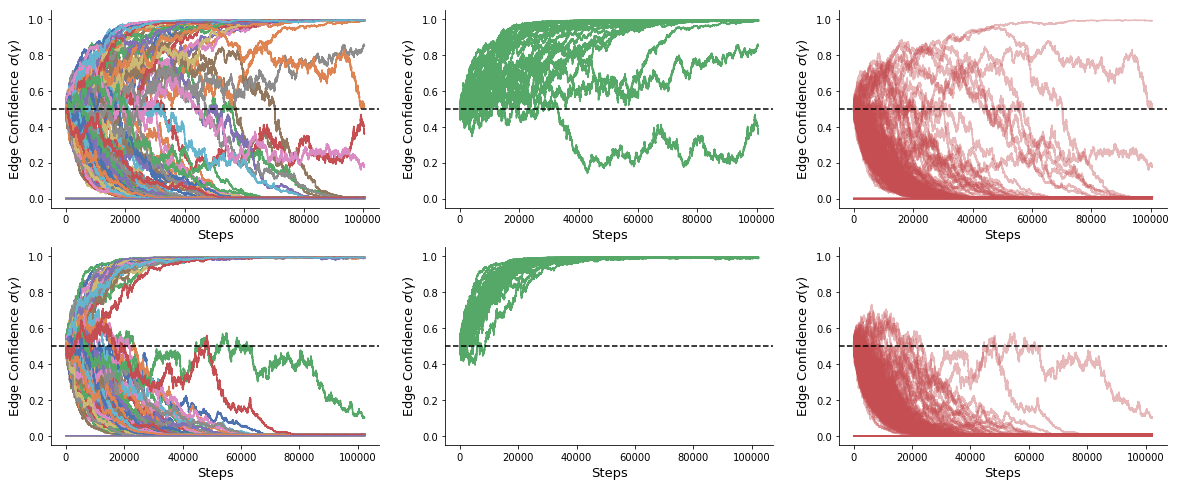}\\
    (d) Graph: Bidiag
    \includegraphics[width=0.9\textwidth]{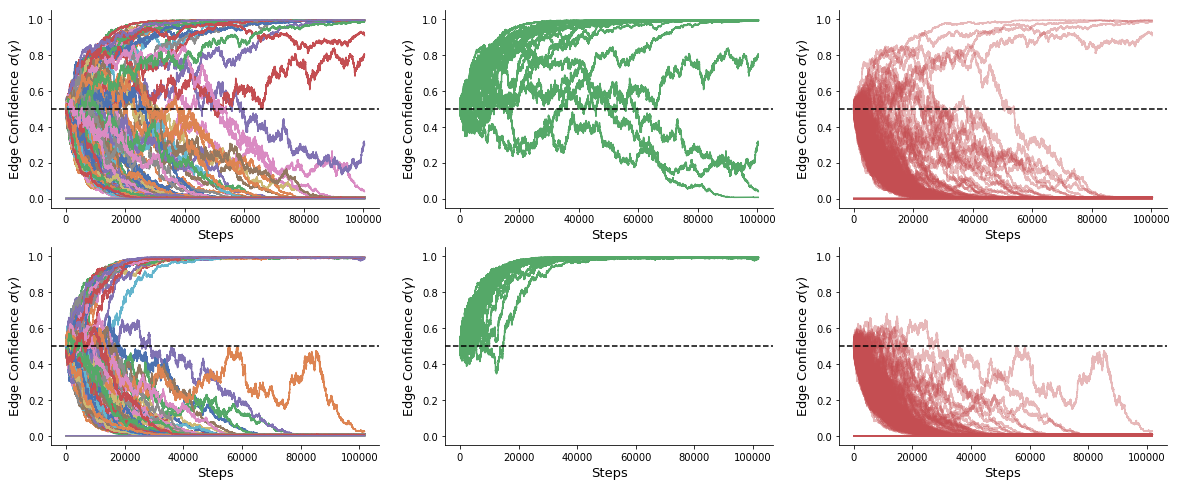}\\
    (e) Graph: Jungle
    \includegraphics[width=0.9\textwidth]{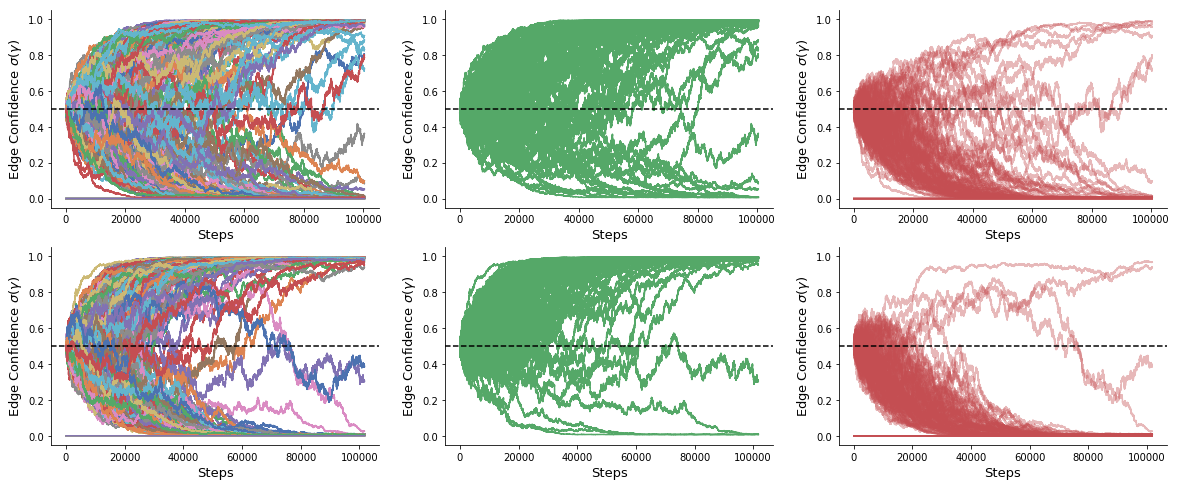}\\
    (f) Graph: Full
    \caption{Edge Dynamics of the examined structured graphs spanning over 15 variables - Part 2: The upper part shows the dynamics of random targeting and the lower of active intervention targeting.}
    \label{fig:extAnalysis_edgeDynamics}
\end{figure}

\clearpage
\subsubsection{Improved robustness with DSDI+AIT in noise perturbed environments}
\label{appendix:noise}
While section \S\ref{sec:results_noise} highlights our key findings in noise-perturbed systems, we examine the impact of AIT in noise perturbed environments more thoroughly in this section. Therefore, we systematically analyze experiments under different noise levels in the setting of binary data generated from random graphs of varying densities. A noise level $\eta$ denotes the probability of flipping a random variable and apply it to all measured variables of observational and interventional samples. 
 
Evaluating convergence on various ER graphs of varying densities over 10 variables under different noise levels reveals that the impact of AIT becomes of larger magnitude as the density of the graph and the noise level increases.

\begin{table}[h!]
    \centering
    \caption{Performance evaluation (SHD) under different noise level $\eta$ for structured and random graphs \;\;$^{(*)}$ denotes average SHD over 3 random graphs.}
    \label{app_tab:dsdi_noiseSparse}
    \resizebox{0.9\textwidth}{!}{%
    \begin{tabular}{llcccccc|rrr}
    \toprule
    \multicolumn{1}{c}{}    &        & \textbf{Chain10} & \textbf{Collider10} & \textbf{Tree10} & \textbf{Bidiag10} & \textbf{Jungle10} & \textbf{Full10} & \textbf{ER-1$^{(*)}$} & \textbf{ER-2$^{(*)}$} & \textbf{ER-4$^{(*)}$} \\
    \toprule
    \multirow{2}{*}{$\eta=0.0$}  & Random & 0 & 0 & 0 & 0 & 0 & 0 &  0.0 ($\pm 0.0$)  & 0.0 ($\pm 0.0$) &  0.0 ($\pm 0.0$)  \\
                            & AIT    & 0 & 0 & 0 & 0 & 0 & 0 &  0.0 ($\pm 0.0$)  & 0.0 ($\pm 0.0$) &  0.0 ($\pm 0.0$)  \\
    \midrule
    \multirow{2}{*}{$\eta=0.01$} & Random & 0 & 0 & 0 & 0 & 0 & 3 &  0.0 ($\pm 0.0$)  & 0.0 ($\pm 0.0$) &  0.6 ($\pm 0.5$)  \\
                            & AIT    & 0 & 0 & 0 & 0 & 0 & 0 &  0.0 ($\pm 0.0$)  & 0.0 ($\pm 0.0$) &  0.0 ($\pm 0.0$)  \\
    \midrule
    \multirow{2}{*}{$\eta=0.02$} & Random & 0 & 4 & 0 & 0 & 0 & 12 &  0.0 ($\pm 0.0$)  & 0.0 ($\pm 0.0$) &  6.0 ($\pm 1.6$)  \\
                            & AIT    & 0 & 0 & 0 & 0 & 0 & 3  &  0.0 ($\pm 0.0$)  & 0.0 ($\pm 0.0$) &  0.0 ($\pm 0.0$)  \\
    \midrule
    \multirow{2}{*}{$\eta=0.05$} & Random & 1 & 9 & 0 & 2 & 1 & 33 &  1.3 ($\pm 0.5$)  & 8.0 ($\pm 2.2$) &  27.0 ($\pm 0.5$)  \\
                            & AIT    & 0 & 7 & 0 & 0 & 0 & 23 &  0.0 ($\pm 0.0$)  & 1.3 ($\pm 0.5$) &  18.7 ($\pm 1.2$)  \\
    \midrule
    \multirow{2}{*}{$\eta=0.1$} & Random & 9 & 9 & 9 & 16 & 16 & 45  & 11.0 ($\pm 0.8$) & 20.7 ($\pm 0.5$) &  40.0 ($\pm 0.8$)  \\
                            & AIT   & 7 & 9 & 6 & 16 & 15 & 44  & 10.3 ($\pm 0.5$) & 20.0 ($\pm 0.8$) &  39.3 ($\pm 1.2$)  \\
    \bottomrule
    \end{tabular}%
    }
\end{table}

 \begin{figure}[h!]
 \centering
  
    (i) \texttt{ER-1:} 
    \minipage{0.21\textwidth}
        \centering
        \includegraphics[width=0.98\linewidth]{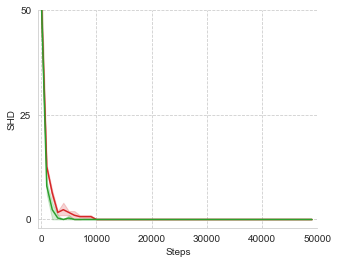}\\
    \endminipage\hfill
    \minipage{0.21\textwidth}
        \centering
        \includegraphics[width=0.98\linewidth]{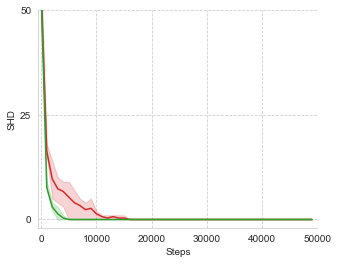}\\
    \endminipage\hfill
    \minipage{0.21\textwidth}
        \centering
        \includegraphics[width=0.98\linewidth]{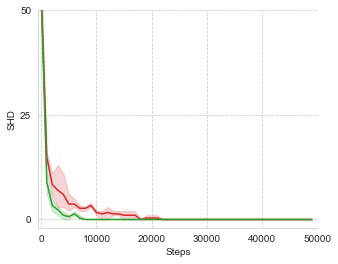}\\
    \endminipage\hfill
    \minipage{0.21\textwidth}
        \centering
        \includegraphics[width=0.98\linewidth]{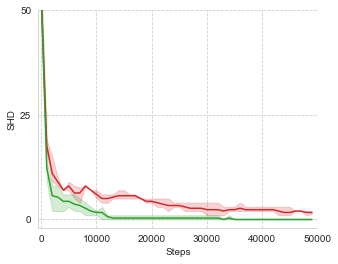}\\
    \endminipage\hfill
    
   (ii) \texttt{ER-2:}
    \minipage{0.21\textwidth}
        \centering
        \includegraphics[width=0.98\linewidth]{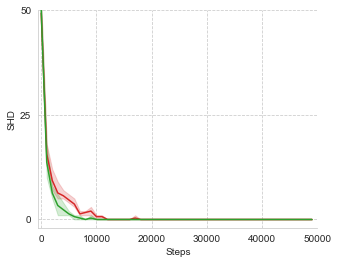}\\
    \endminipage\hfill
    \minipage{0.21\textwidth}
        \centering
        \includegraphics[width=0.98\linewidth]{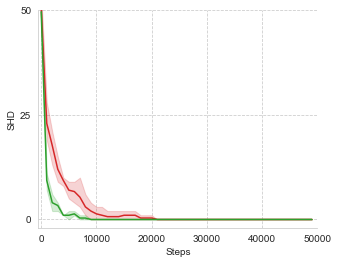}\\
    \endminipage\hfill
    \minipage{0.21\textwidth}
        \centering
        \includegraphics[width=0.98\linewidth]{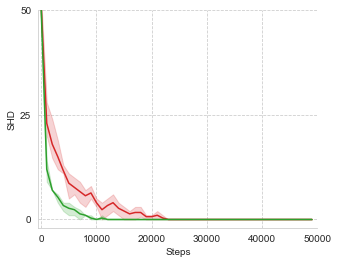}\\
    \endminipage\hfill
    \minipage{0.21\textwidth}
        \centering
        \includegraphics[width=0.98\linewidth]{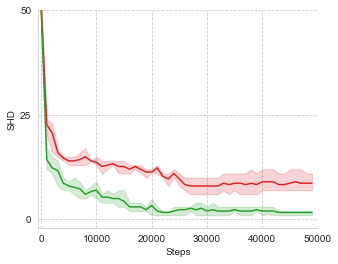}\\
    \endminipage\hfill
    
    (iii) \texttt{ER-4:}
    \minipage{0.21\textwidth}
        \centering
        \includegraphics[width=0.98\linewidth]{figures/noise/ExtendedPerformanceComparions_N10_ER4_NOISE0.0_SHDonly.png}\\
    \endminipage\hfill
    \minipage{0.21\textwidth}
        \centering
        \includegraphics[width=0.98\linewidth]{figures/noise/ExtendedPerformanceComparions_N10_ER4_NOISE0.01_SHDonly.png}\\
    \endminipage\hfill
    \minipage{0.21\textwidth}
        \centering
        \includegraphics[width=0.98\linewidth]{figures/noise/ExtendedPerformanceComparions_N10_ER4_NOISE0.02_SHDonly.png}\\
    \endminipage\hfill
    \minipage{0.21\textwidth}
        \centering
        \includegraphics[width=0.98\linewidth]{figures/noise/ExtendedPerformanceComparions_N10_ER4_NOISE0.05_SHDonly.png}\\
    \endminipage\hfill
    
    \minipage{0.1\textwidth}
        \centering
        \qquad
    \endminipage\hfill
    \minipage{0.21\textwidth}
        \centering
        (a) $\eta = 0$
    \endminipage\hfill
    \minipage{0.21\textwidth}
        \centering
        (b) $\eta = 0.01$
    \endminipage\hfill
    \minipage{0.21\textwidth}
        \centering
        (c) $\eta = 0.02$
    \endminipage\hfill
    \minipage{0.21\textwidth}
        \centering
        (d) $\eta = 0.05$
    \endminipage\hfill
    \caption{Convergence behaviour in terms of SHD for random ER graphs of various densities over 10 variables under different noise levels $\eta$. Overall, Active Intervention Targeting (orange) clearly outperforms Random Targeting (blue) over all densities under all noise levels. The performance gap becomes of larger magnitude as density of the graph and the noise level increases. Error bands were estimated using 3 random ER graphs per setting.}
   \label{fig:dsdi_noise_speed_ER_various}
\end{figure}

\clearpage
\subsubsection{Identification of informative intervention targets}
\label{app:ablation_informativeTargets}
Our proposed method aims to select most \textit{informative} intervention target(s) $I_{k^*} \in I$ with respect to identifiability of the underlying structure. We conjecture that such targets yield relatively high discrepancy between samples drawn under different hypothesis graphs, indicating larger uncertainty about the target node's relation to its parents and/or children. 

In order to evaluate our methods capability of detecting informative intervention targets, we perform multiple experiments on structured graph structures (\texttt{chain5}, \texttt{tree5} and \texttt{full5}) where we preinitalize the structural belief to the ground-truth structure structure but keeping one edge between a pair of nodes ($i$,$j$) undirected, i.e. $\sigma(\gamma_{i,j}) = \sigma(\gamma_{j,i}) = 0.5$. Throughout the experiments, we vary the position of the undirected edge and analyze which nodes are targeted by our method.

\begin{figure}[h!]
    \centering
    
    \minipage{0.60\textwidth}
        \centering
        \includegraphics[width=\linewidth]{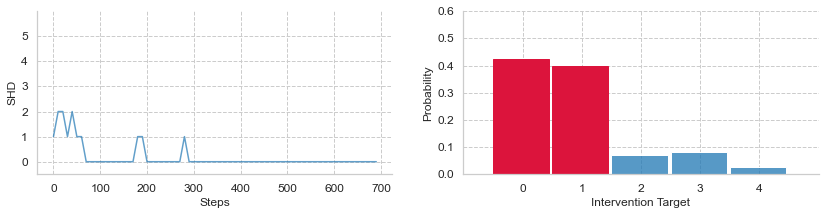}
    \endminipage\hfill
    \minipage{0.34\textwidth}%
        \centering
        \includegraphics[width=\linewidth]{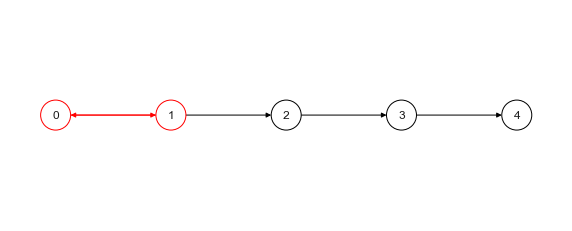}
    \endminipage
    
    \minipage{0.60\textwidth}
        \centering
        \includegraphics[width=\linewidth]{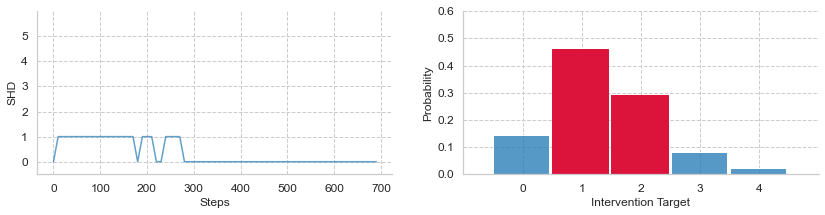}
    \endminipage\hfill
    \minipage{0.34\textwidth}%
        \centering
        \includegraphics[width=\linewidth]{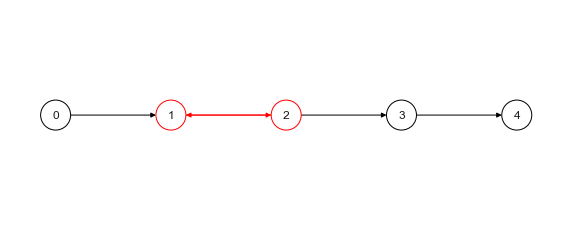}
    \endminipage
    
    \minipage{0.60\textwidth}
        \centering
        \includegraphics[width=\linewidth]{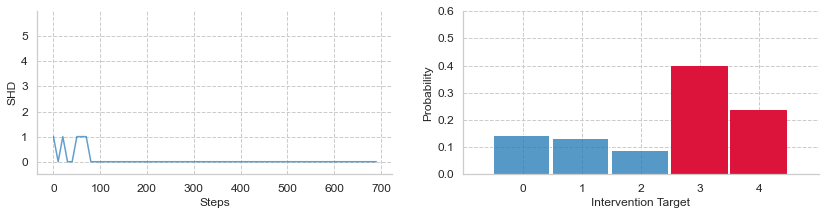}
    \endminipage\hfill
    \minipage{0.34\textwidth}%
        \centering
        \includegraphics[width=\linewidth]{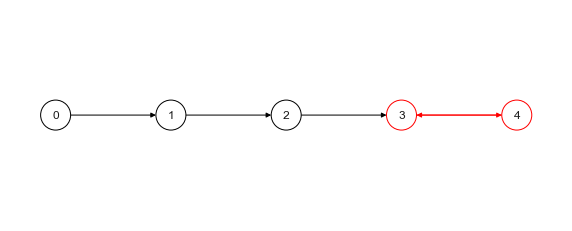}
    \endminipage
    
    \minipage{0.60\textwidth}
        \centering
        \includegraphics[width=\linewidth]{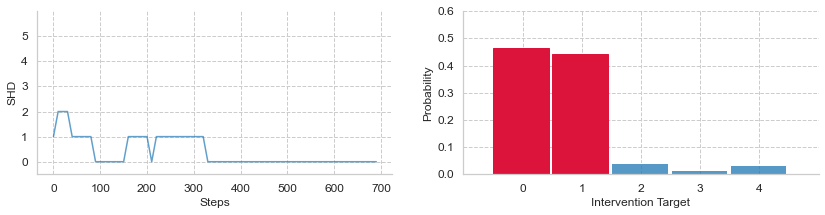}
    \endminipage\hfill
    \minipage{0.34\textwidth}%
        \centering
        \includegraphics[width=\linewidth]{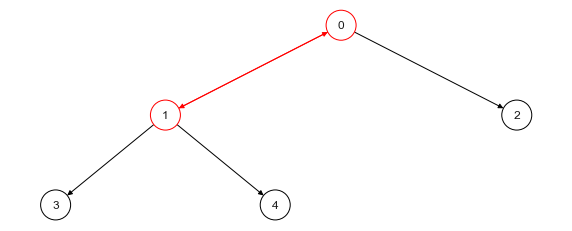}
    \endminipage
    
    \minipage{0.60\textwidth}
        \centering
        \includegraphics[width=\linewidth]{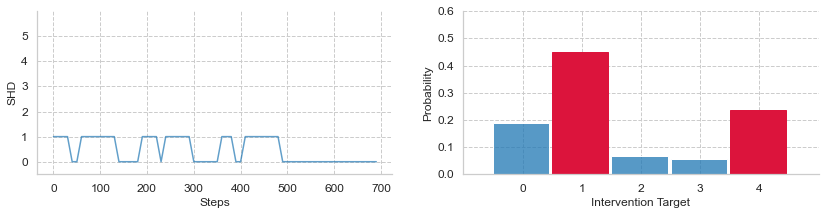}
    \endminipage\hfill
    \minipage{0.34\textwidth}%
        \centering
        \includegraphics[width=\linewidth]{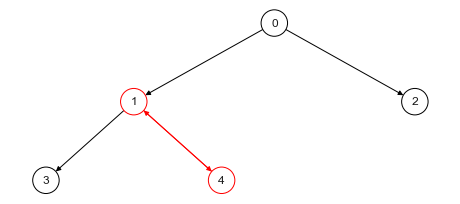}
    \endminipage
    
    \caption{AIT chooses informative intervention targets by preferentially identifying and targeting the pair of nodes corresponding to the undirected edge (nodes are marked red in the distribution of selected target nodes and edges is visualized red in the graph on the right). - First set of experiments based on the structured graphs \texttt{chain5} and \texttt{tree5}.}
    \label{fig:appendix_informativeTargets_Part1}
\end{figure}

\begin{figure}[t!]
    \minipage{0.60\textwidth}
        \centering
        \includegraphics[width=\linewidth]{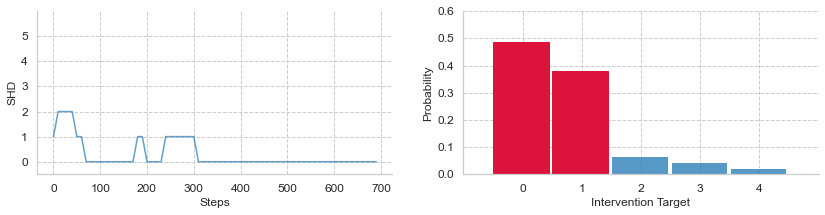}
    \endminipage\hfill
    \minipage{0.32\textwidth}%
        \centering
        \includegraphics[width=\linewidth]{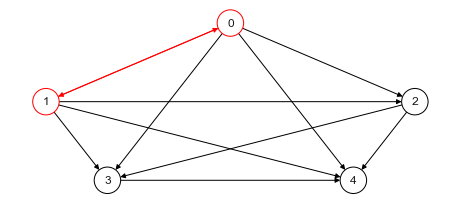}
    \endminipage
    
    \minipage{0.60\textwidth}
        \centering
        \includegraphics[width=\linewidth]{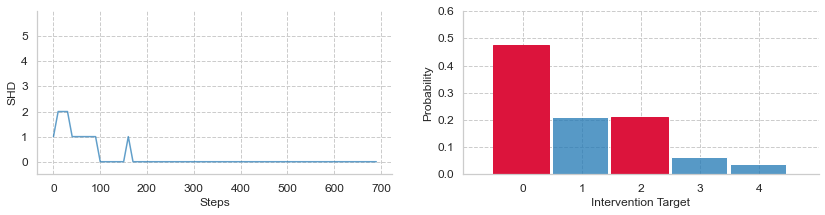}
    \endminipage\hfill
    \minipage{0.32\textwidth}%
        \centering
        \includegraphics[width=\linewidth]{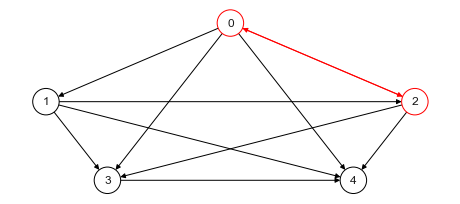}
    \endminipage
    
    \minipage{0.60\textwidth}
        \centering
        \includegraphics[width=\linewidth]{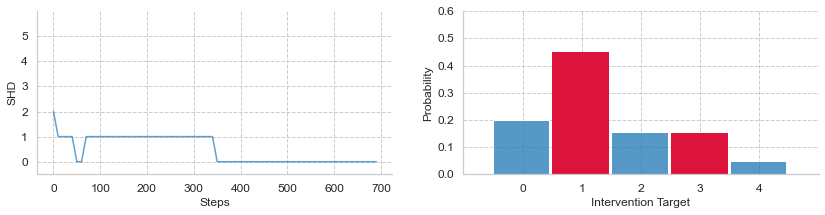}
    \endminipage\hfill
    \minipage{0.32\textwidth}%
        \centering
        \includegraphics[width=\linewidth]{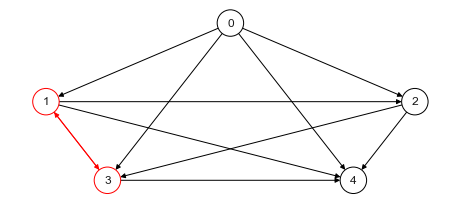}
    \endminipage
    
    \minipage{0.60\textwidth}
        \centering
        \includegraphics[width=\linewidth]{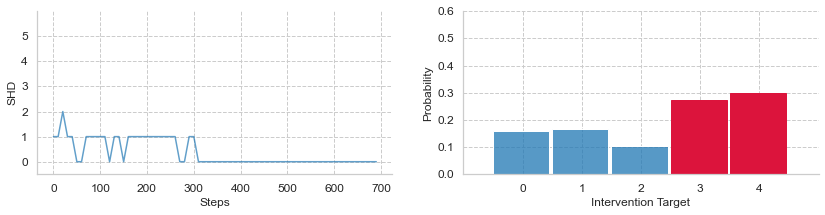}
    \endminipage\hfill
    \minipage{0.32\textwidth}%
        \centering
        \includegraphics[width=\linewidth]{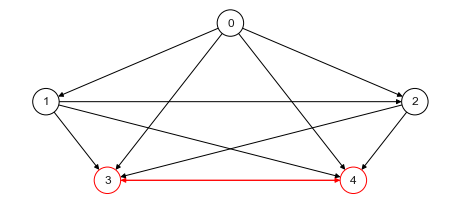}
    \endminipage
    
    \minipage{0.60\textwidth}
        \centering
        \includegraphics[width=\linewidth]{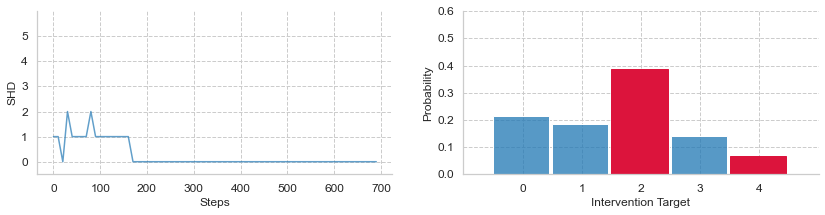}
    \endminipage\hfill
    \minipage{0.32\textwidth}%
        \centering
        \includegraphics[width=\linewidth]{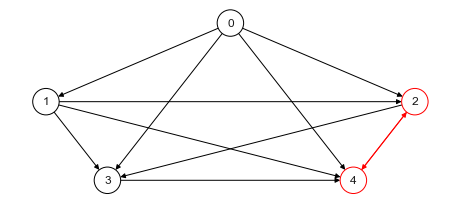}
    \endminipage
    
    \caption{AIT chooses informative intervention targets by preferentially identifying and targeting the pair of nodes corresponding to the undirected edge (nodes are marked red in the distribution of selected target nodes and edges is visualized red in the graph on the right). - Second set of experiments based on the structured graph \texttt{full5}.}
    \label{fig:appendix_informativeTargets_Part2}
    \vspace{1.5\baselineskip}
\end{figure}

Over all evaluated settings, we can observe how AIT preferentially targets the pair of nodes corresponding to the undirected edge, with small preferences towards the source nodes of the correct directed edge (see in Figure \ref{fig:appendix_informativeTargets_Part1} and Figure \ref{fig:appendix_informativeTargets_Part2}). This observation is in line with our conjecture that AIT preferentially targets nodes with larger uncertainty about the target node's relation to its parents and/or children. 

\newpage
\subsubsection{Limited Intervention Targets}
While we allow access to all possible single-target interventions in all other experiments, real world settings are usually more restrictive. Specific interventions might be either technically impossible or even unethical, or the experiments might want to prevent interventions upon specific target nodes due to increased experiment costs. In order to test the capability of AIT, we limit the set of possible intervention targets in the following experiments and analyze the resulting behaviour based on DSDI. We examine speed of convergence and the effect on the target distribution under different scenarios on structured graphs using DSDI with AIT based on single-target interventions. 

\textbf{Scenario 1:} We perform experiments on a \texttt{Chain5} graph where we restrict us on intervening upon a different node in five experiment and once allow access to all targets as a comparison.

\begin{figure}[h!]

    \minipage{0.65\textwidth}
        \centering
        \includegraphics[width=\linewidth]{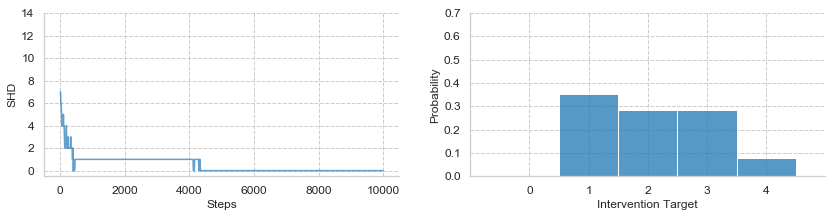}
    \endminipage\hfill
    \minipage{0.32\textwidth}%
        \centering
        \includegraphics[width=\linewidth]{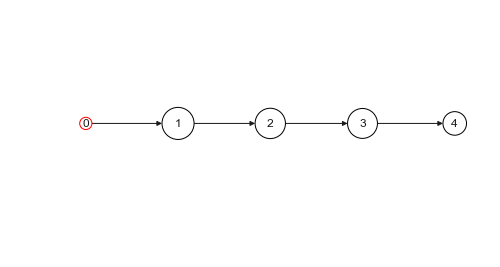}
    \endminipage
    
    \minipage{0.65\textwidth}
        \centering
        \includegraphics[width=\linewidth]{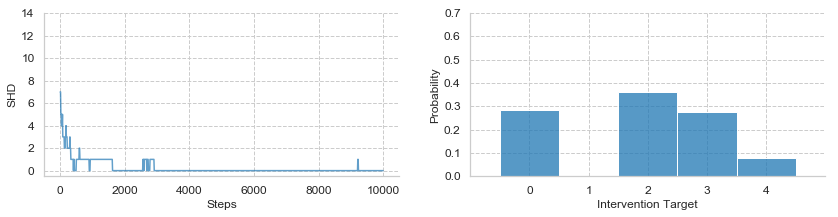}
    \endminipage\hfill
    \minipage{0.32\textwidth}%
        \centering
        \includegraphics[width=\linewidth]{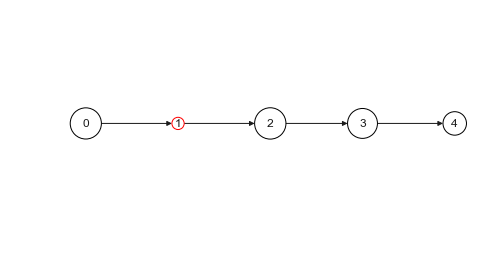}
    \endminipage
    
    \minipage{0.65\textwidth}
        \centering
        \includegraphics[width=\linewidth]{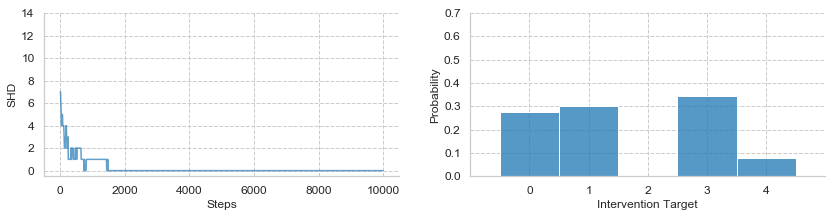}
    \endminipage\hfill
    \minipage{0.32\textwidth}%
        \centering
        \includegraphics[width=\linewidth]{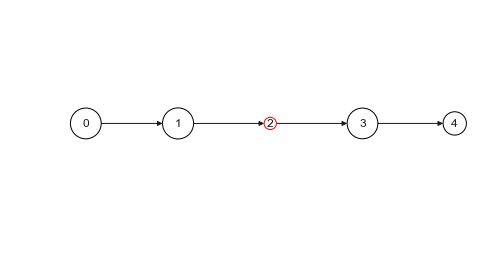}
    \endminipage
    
    \minipage{0.65\textwidth}
        \centering
        \includegraphics[width=\linewidth]{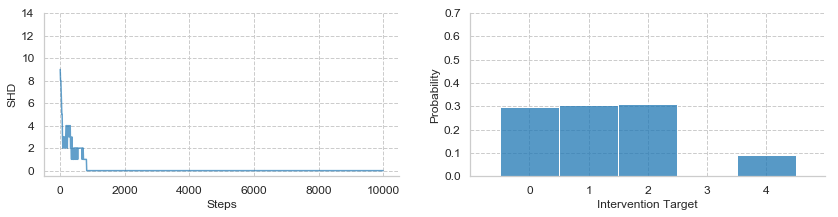}
    \endminipage\hfill
    \minipage{0.32\textwidth}%
        \centering
        \includegraphics[width=\linewidth]{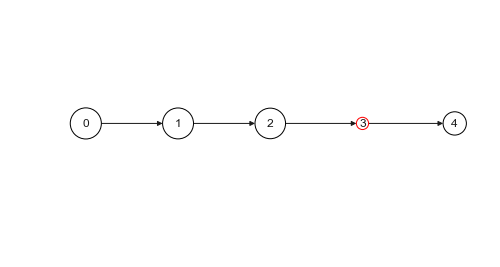}
    \endminipage
    
    \minipage{0.65\textwidth}
        \centering
        \includegraphics[width=\linewidth]{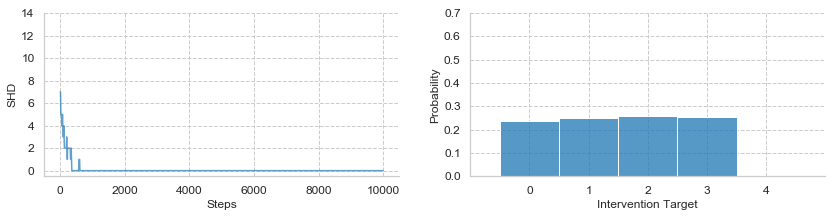}
    \endminipage\hfill
    \minipage{0.32\textwidth}%
        \centering
        \includegraphics[width=\linewidth]{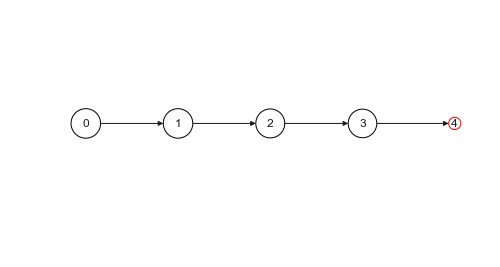}
    \endminipage
    
    \minipage{0.65\textwidth}
        \centering
        \includegraphics[width=\linewidth]{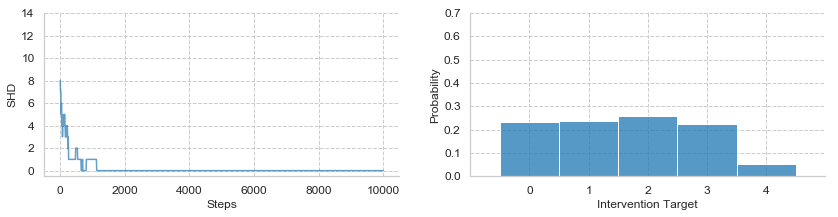}
    \endminipage\hfill
    \minipage{0.32\textwidth}%
        \centering
        \includegraphics[width=\linewidth]{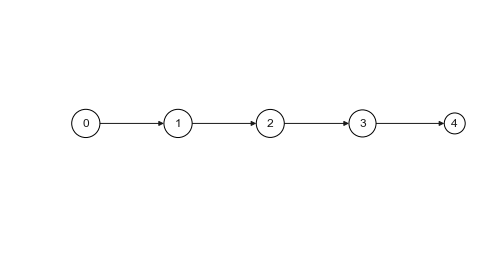}
    \endminipage

    \caption{Limited intervention targets on \texttt{Chain5}: The impact of the restricted target node (red circled node) is clearly observable in the convergence speed (left) and distribution of target selections (middle). The speed of convergence indicates a dependence on the topological characteristic of the restricted intervention target.}
    \label{fig:ablation_chain5_blockedTargets}
\end{figure}

Throughout the experiments, we observe that blocking interventions on nodes of a higher topological level results in greater degradation of the convergence speed compared to blocked intervention on lower levels (see Figure \ref{fig:ablation_chain5_blockedTargets}). Furthermore, the distribution of selected targets indicates that our method preferentially chooses neighboring nodes of a blocked target node in the restricted setting. 

\clearpage
\textbf{Scenario 2:} We perform multiple experiments on a \texttt{Tree5} graph where we restrict access to different subsets of nodes (e.g. root node, set of all sink nodes) for single-target interventions.

\begin{figure}[h!]
    
    \minipage{0.65\textwidth}
        \centering
        \includegraphics[width=\linewidth]{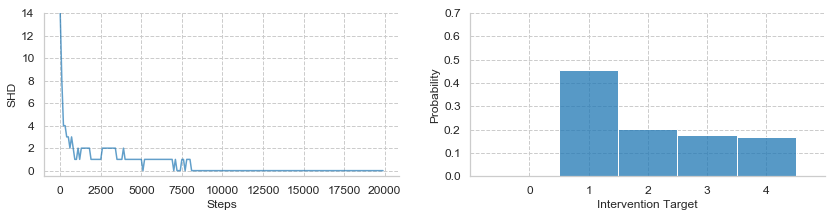}
    \endminipage\hfill
    \minipage{0.32\textwidth}%
        \centering
        \includegraphics[width=\linewidth]{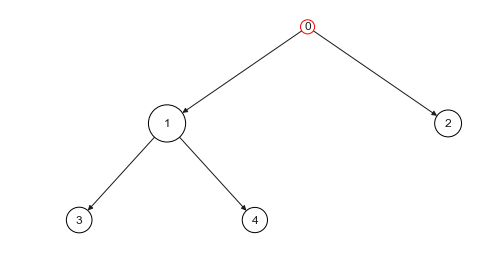}
    \endminipage
    
    \minipage{0.65\textwidth}
        \centering
        \includegraphics[width=\linewidth]{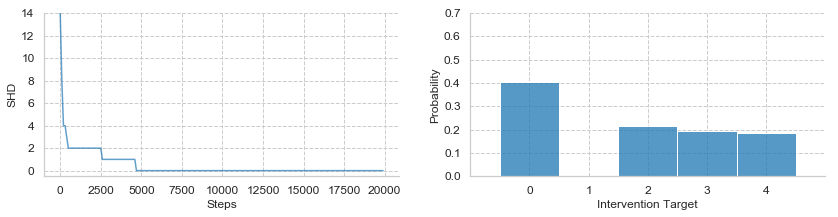}
    \endminipage\hfill
    \minipage{0.32\textwidth}%
        \centering
        \includegraphics[width=\linewidth]{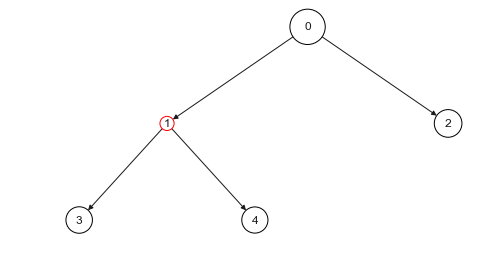}
    \endminipage
    
    \minipage{0.65\textwidth}
        \centering
        \includegraphics[width=\linewidth]{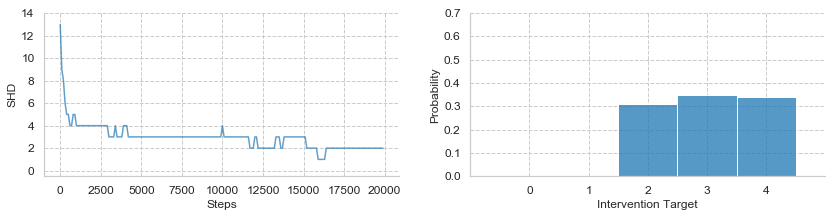}
    \endminipage\hfill
    \minipage{0.32\textwidth}%
        \centering
        \includegraphics[width=\linewidth]{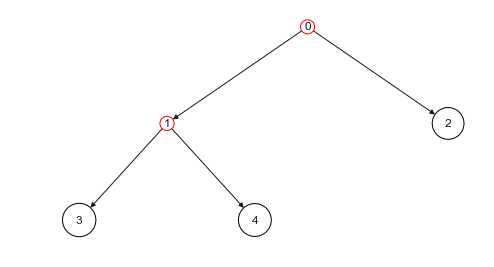}
    \endminipage
    
    \minipage{0.65\textwidth}
        \centering
        \includegraphics[width=\linewidth]{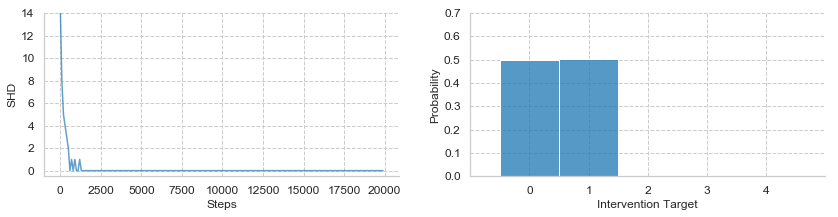}
    \endminipage\hfill
    \minipage{0.32\textwidth}%
        \centering
        \includegraphics[width=\linewidth]{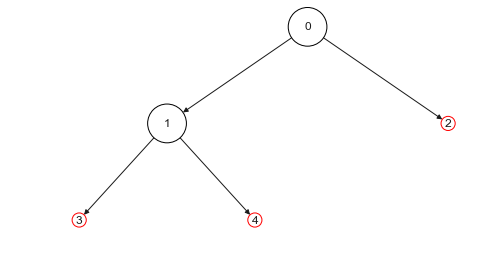}
    \endminipage
    
    \minipage{0.65\textwidth}
        \centering
        \includegraphics[width=\linewidth]{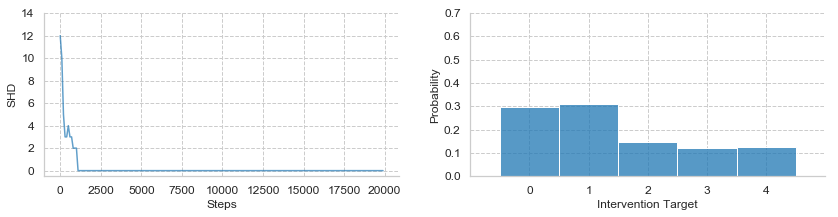}
    \endminipage\hfill
    \minipage{0.32\textwidth}%
        \centering
        \includegraphics[width=\linewidth]{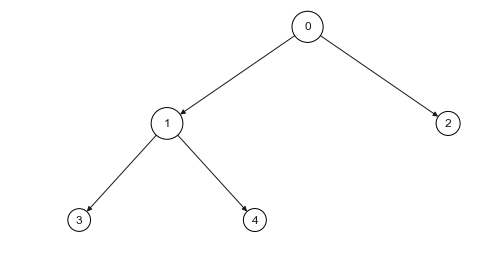}
    \endminipage
    
    \caption{Limited intervention targets on \texttt{Tree5}: The impact of the restricted target nodes (red circled nodes) is clearly observable in the convergence speed (left) and distribution of target selections (middle).}
    \label{fig:ablation_tree5_blockedTargets}
\end{figure}

Similar to the experiments on \texttt{Chain5}, we observe a clear impact of the available intervention targets on the convergence speed and identifiability of the underlying structure (see Figure \ref{fig:ablation_tree5_blockedTargets}). While preventing interventions on all sink nodes (node 2, 3 and 4) results in improved convergence towards the underlying structure, restricted access to the set of nodes which act as causes of other nodes (node 0 and 1) prevents us from identifying the correct underlying structure.


\clearpage
\subsection{Continuous Setting: Technical Details and Results}
\label{sec:appendix_dcdi}

While the original framework of DCDI \citep{brouillard2020differentiable} proposes a joint-optimization over the observational and interventional sample space by selecting samples at random, we adapt their framework to the setting of active causal discovery where we acquire interventional sample in an adaptive manner. We hypothesize that a controlled selection of informative intervention targets allows a more rapid and controlled discovery of the underlying causal structure.

\subsubsection{Differentiable Causal Discovery from Interventions (DCDI)}
The work of DCDI \citep{brouillard2020differentiable} addresses causal discovery from \emph{continuous} data as a continuous-constrained optimization problem using neural networks to model parameters of Gaussian distributions or normalizing flows \citep{rezende2015variational} which represent conditional distributions. Unlike SDI's iterative training of the structural and functional parameters, DCDI optimizes the causal adjacency matrix and functional parameters jointly over the fused data space. But like SDI, DCDI  uses random and independent interventions. 

\subsubsection{Integration of AIT into DCDI}
\label{sec:dcdi_ait_integration}
Instead of demanding the full interventional target space during the complete optimization as in the original approach, we split the optimization procedure into different episodes, where AIT is used to estimate a target space $I$ of size $K$ for each episode. This is done by computing the discrepancy scores over all possible intervention targets and selecting the $K$ highest scoring targets. During an episode, we continue by performing $L$ gradient steps using the fixed target space $I$ and reevaluate it afterwards for the next episode. We visualize the adaption in the following high-level outline of the individual methodologies. 

\begin{minipage}{0.4\textwidth}
    \begin{algorithm}[H]\small
        \caption{DCDI}\label{algorithm}
        \begin{algorithmic}[1]
            \item[]
            \State{$I \leftarrow$ Full target space of size $K=N$}
            \State{Run DCDI on $I$ until convergence} 
        \end{algorithmic}
        \label{algo:DCDI}
    \end{algorithm}
\end{minipage}
\hfill
\begin{minipage}{0.05\textwidth}
\quad\vspace{6mm}\\
$\mathbf{\Rightarrow}$
\end{minipage}
\hfill
\begin{minipage}{0.53\textwidth}
    \begin{algorithm}[H]\small
        \caption{DCDI + AIT}\label{algorithm}
        \begin{algorithmic}[1]
            \item[]
            \For{episode $e=0$ \textbf{until} convergence}
                    \State{$I \leftarrow$ Estimate target space of size $K$ using AIT}
                    \State{Run $L$ gradient steps of DCDI on  $I$} 
            \EndFor
        \end{algorithmic}
        \label{algo:DCDI_AIT}
    \end{algorithm}
\end{minipage}

\subsubsection{Evaluation}
 We evaluate the effectiveness of AIT in the base framework of DCDI in the setting of non-linear, continuous data generated from random graphs over $N=10$ variables and show the potential of our proposed method.
 
\begin{figure}[h!]
    \centering
    \includegraphics[width=\linewidth]{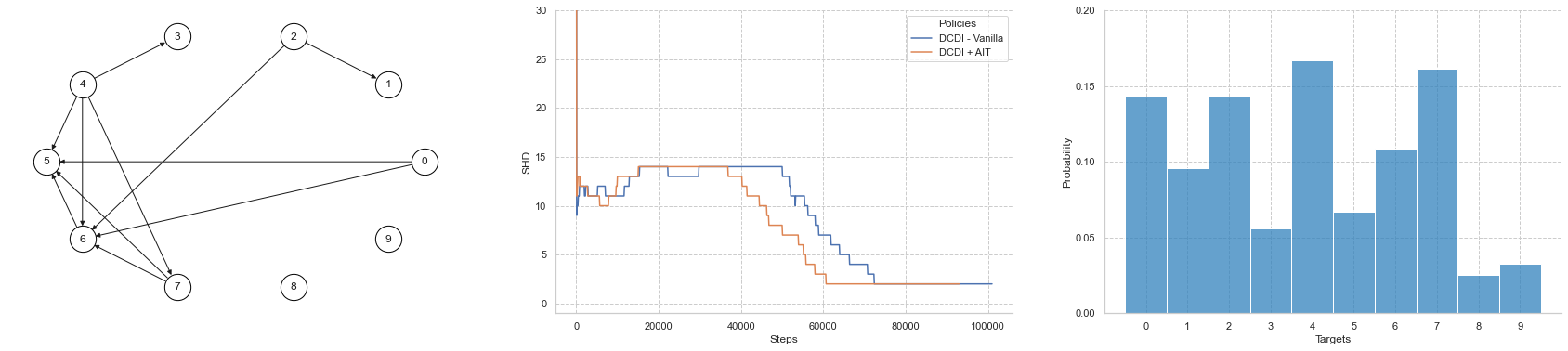}
    \includegraphics[width=\linewidth]{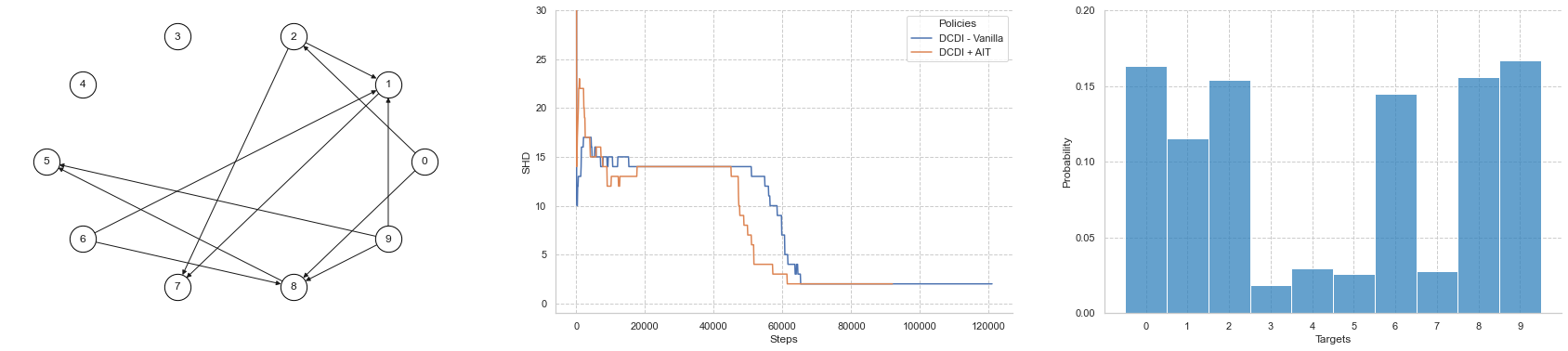}
    \includegraphics[width=\linewidth]{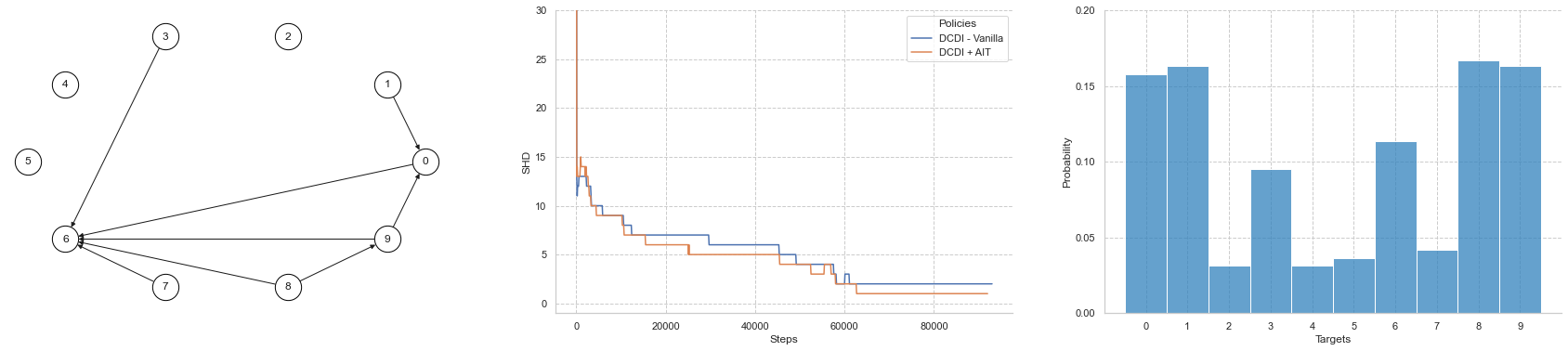}
    \includegraphics[width=\linewidth]{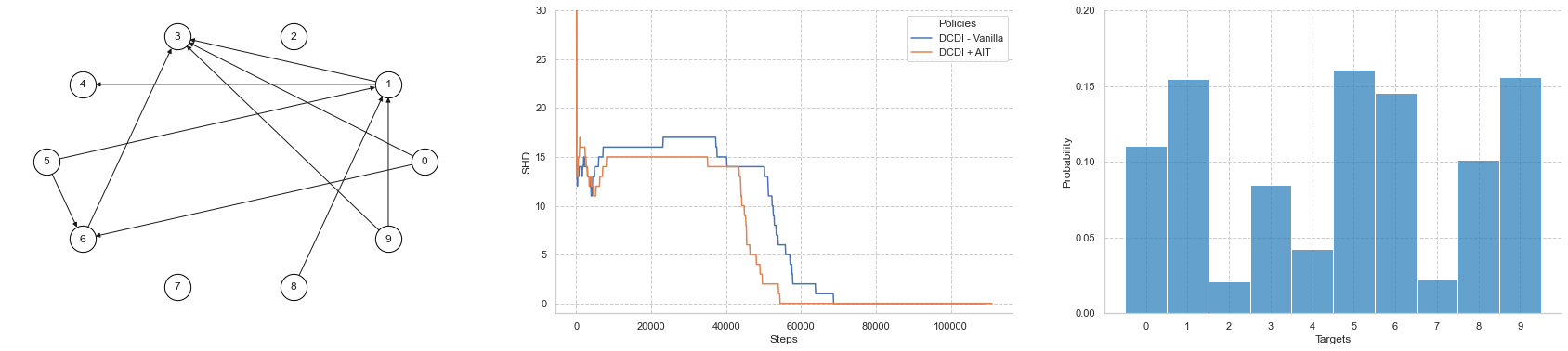}
    \caption{While DCDI Vanilla assumes access to the full interventional target space through the complete optimization, the AIT guided DCDI approach reevaluates its interventional target space of size $K=6$ every $L=1000$ gradient steps. Among the above evaluated graphs (ground-truth on the left), DCDI+AIT demonstrates a more rapid identification of the underlying causal structure while achieving same or better performance in terms of SHD. The distribution of selected single-node intervention targets reveals again its connection to the topological properties of the corresponding nodes. }
    \label{fig:dcdi_experiments}
\end{figure}

 \textbf{Structural Identification / Convergence:} Despite their joint optimization formulation is not apriori designed for the setting of experimental design, an AIT guided version shows superior/competitive performance in terms of structural identification and sample complexity over the original formulation (see Figure \ref{fig:dcdi_experiments}). 
 
 \textbf{Distribution of Intervention Targets:} As in DSDI, we observe strong correlation of the number of target selections with the measured topological properties of the specific nodes. This indicates a controlled discovery of the underlying causal structure through preferential targeting of nodes with greater (downstream) impact on the overall system. In addition, interventions on variables without children are drastically reduced (see also \S\ref{sec:results_undesirableInterventions} for equivalent observations in DSDI).
 
 \textbf{Effect of Target Space Size $K$:} While the original formulation assumes $K=N$ for the complete optimization procedure (i.e. $L=1$) and relies on random samples out of the full target space, our adapted AIT-guided version of DCDI constrains the target space to a subset of targets for each episode. An ablation study on the size of the target space shows that for all choices of $K \in \lbrace 2,4,6,8 \rbrace$, our approach outperforms the original formulation in terms of sample complexity while achieving same or better performance in terms of SHD.

\begin{figure}[h!]
    \centering
    \minipage{0.24\textwidth}
        \centering
        \includegraphics[width=\linewidth]{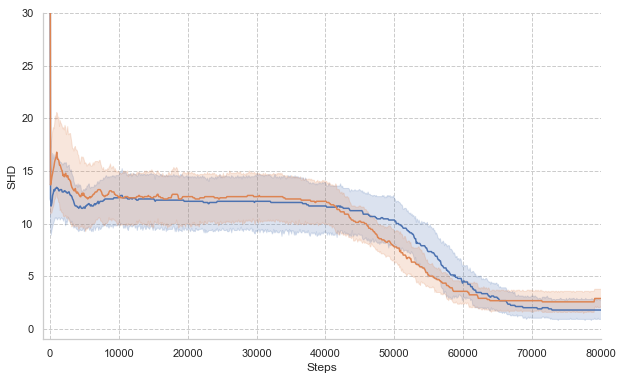}
        (a) $K=2$
    \endminipage\hfill
    \minipage{0.24\textwidth}
        \centering
        \includegraphics[width=\linewidth]{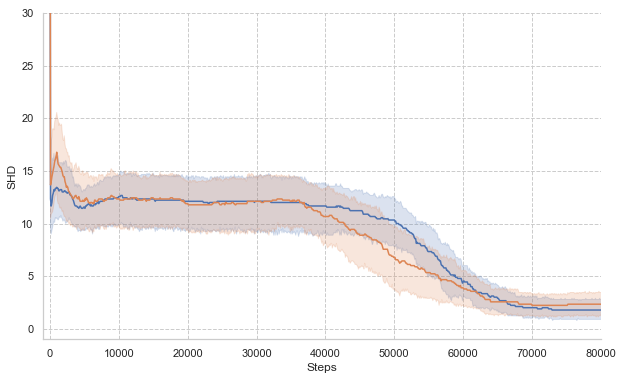}
        (b) $K=4$
    \endminipage\hfill
    \minipage{0.24\textwidth}
        \centering
        \includegraphics[width=\linewidth]{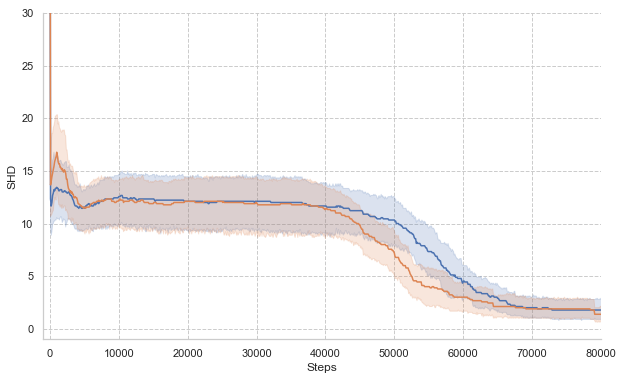}
        (d) $K=6$
    \endminipage\hfill
    \minipage{0.24\textwidth}
        \centering
        \includegraphics[width=\linewidth]{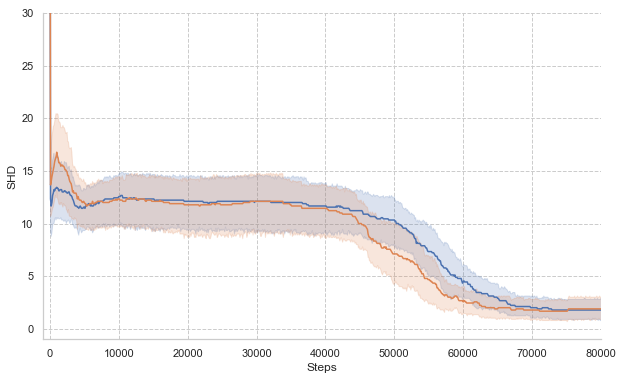}
        (d) $K=8$
    \endminipage\hfill
    \caption{All evaluated target space sizes $K \in \lbrace 2,4,6,8 \rbrace$ show that DCDI+AIT (orange) outperforms DCDI (blue) in terms of sample complexity while achieving similar performance. Error bands were estimated using 10 random ER graphs per setting.}
    \label{fig:dcdi_ablation_TargetSize}
\end{figure}

\quad \vspace{2mm} \\

\end{document}